\documentclass[10pt,twocolumn,letterpaper]{article}

\usepackage{cvpr}
\usepackage{times}
\usepackage{epsfig}
\usepackage{graphicx}
\usepackage{amsmath}
\usepackage{amssymb}
\usepackage{algorithm}
\usepackage{algorithmic}
\usepackage{microtype}
\usepackage{booktabs} 
\usepackage{wrapfig}
\usepackage{multirow}
\usepackage{subfig}
\usepackage{arydshln}
\usepackage{amsmath}
\usepackage{nicefrac}
\usepackage[flushleft]{threeparttable}

\newcommand{\norm}[1]{\left\lVert#1\right\rVert}

\usepackage[pagebackref=true,breaklinks=true,letterpaper=true,colorlinks,bookmarks=false]{hyperref}

\cvprfinalcopy 

\def\cvprPaperID{7659} 

\ifcvprfinal\fi
\begin{document}

\title{Learning to Forget for Meta-Learning}
\author{Sungyong Baik \and Seokil Hong \and Kyoung Mu Lee\\
\and ASRI, Department of ECE, Seoul National University\\
{\tt\small \{dsybaik, hongceo96, kyoungmu\}@snu.ac.kr}
}

\maketitle

\begin{abstract}
Few-shot learning is a challenging problem where the goal is to achieve generalization from only few examples. 
Model-agnostic meta-learning (MAML) tackles the problem by formulating prior knowledge as a common initialization across tasks, which is then used to quickly adapt to unseen tasks. 
However, forcibly sharing an initialization can lead to conflicts among tasks and the compromised (undesired by tasks) location on optimization landscape, thereby hindering the task adaptation. 
Further, we observe that the degree of conflict differs among not only tasks but also layers of a neural network.
Thus, we propose task-and-layer-wise attenuation on the compromised initialization to reduce its influence. 
As the attenuation dynamically controls (or selectively forgets) the influence of prior knowledge for a given task and each layer, we name our method as L2F (Learn to Forget)\footnote{The code is available at \url{https://github.com/baiksung/L2F}}.
The experimental results demonstrate that the proposed method provides faster adaptation and greatly improves the performance. 
Furthermore, L2F can be easily applied and improve other state-of-the-art MAML-based frameworks, illustrating its simplicity and generalizability.
\end{abstract}

\section{Introduction}
\begin{figure}[t]
\begin{center}
\includegraphics[width=0.7\linewidth,angle =270]{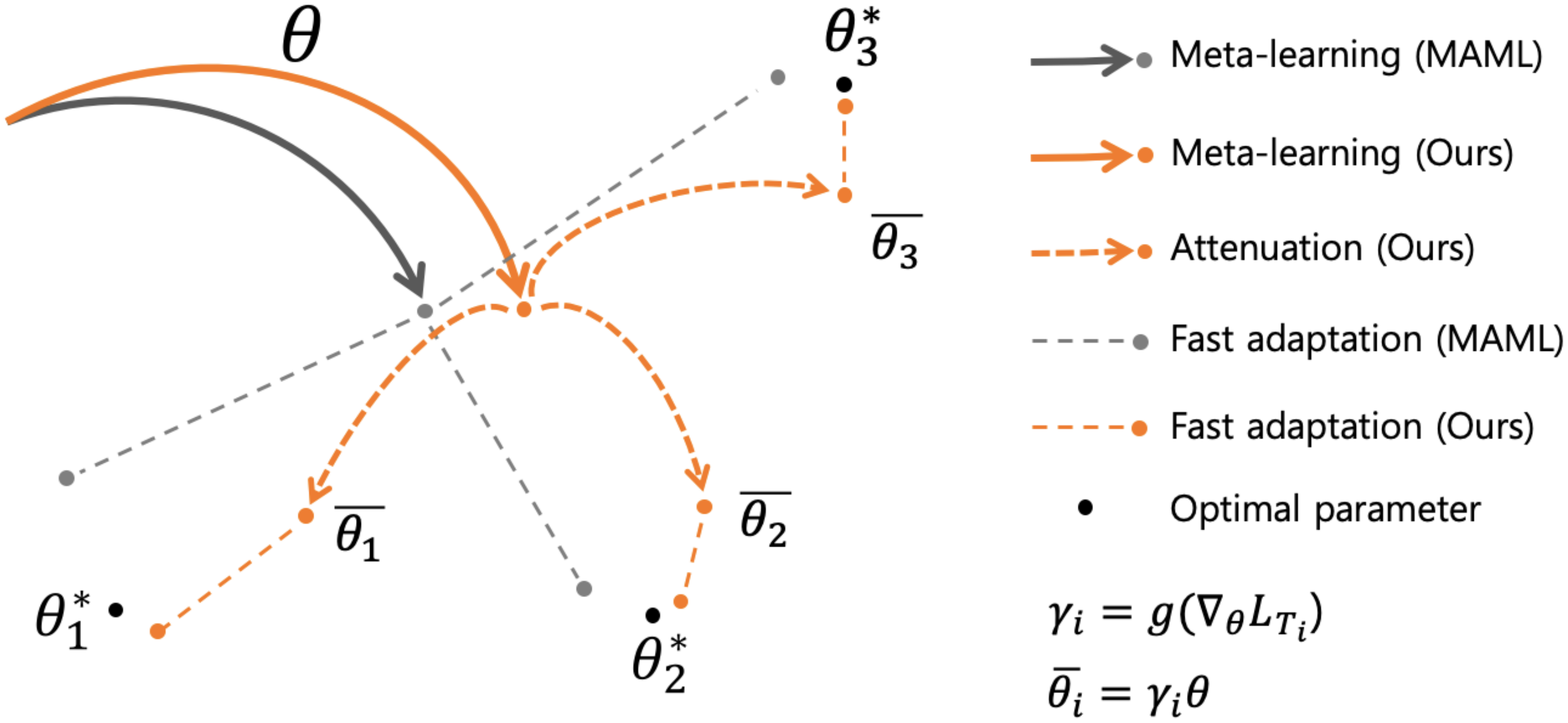}
\end{center}
\vspace{-1cm}
\caption{When there is a large degree of \textit{conflict}, the updated initialization ends up in the location neither of tasks desires. Such undesired (hence compromised) initialization location can make learning difficult during fast adaptation to each task. Our method makes the fast adaptation easier by minimizing the influence of the compromised initialization for each task, through attenuation parameter $\gamma$ generated by the task-conditioned network $g$. This makes the optimization landscape smoother and hence helps achieve better generalization to unseen examples.}
\vspace{-0.35cm}
\label{fig:conflict_analysis}
\end{figure}

Recent deep learning models demonstrate outstanding performance in various fields; however, they require supervised learning with a tremendous amount of labeled data. 
On the other hand, humans are able to learn concepts from only few examples. Considering the cost of data annotation, the capability of humans to learn from few examples is desirable.

When there are concerns for overfitting in few-data regime, data augmentation and regularization techniques are often used. Another commonly used technique is to fine-tune a network pre-trained on large labelled data from another dataset or task~\cite{razavian2014cnn,simonyan2015very}. 
Fine-tuning often does provide adaptation without overfitting even in few-data regime, however at the cost of computation due to many update iterations~\cite{triantafillou2018meta}.
In contrast, meta-learning tackles the problem systematically via two stages of learners: a meta-learner learns common knowledge across a distribution of tasks, which is then used for a learner to quickly learn task-specific knowledge with few examples. 
A popular instance is the model-agnostic meta-learning (MAML)~\cite{finn2017model}, where a meta-learner is formulated such that it learns a common initialization that encodes the common knowledge across tasks.
\begin{figure*}
\begin{center}
\subfloat[loss landscape]{
    \includegraphics[width=0.3\linewidth]{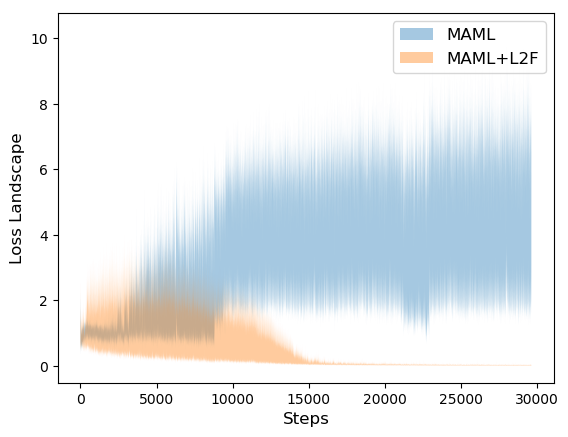}
    \label{fig:loss}
}
\subfloat[gradient predictiveness]{
    \includegraphics[width=0.3\linewidth]{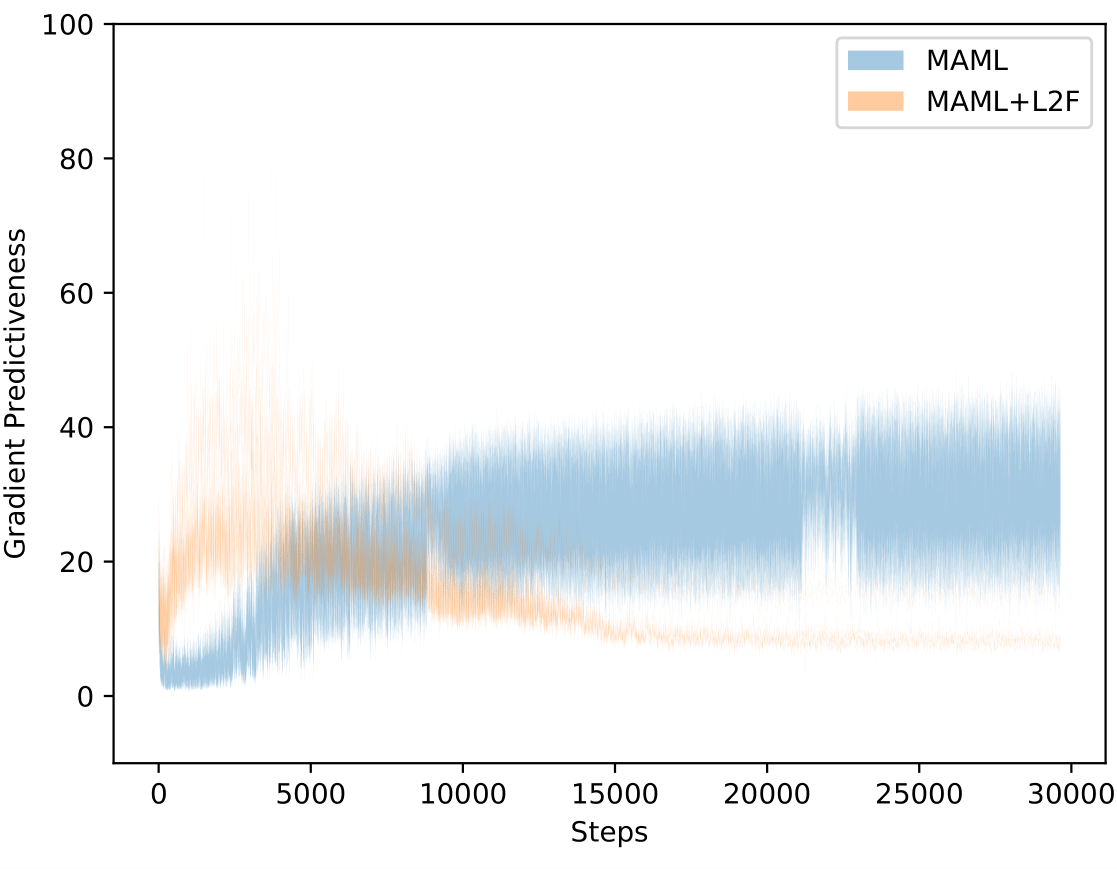}
    \label{fig:gradient}
}
\subfloat[``effective'' $\beta$ smoothness]{
    \includegraphics[width=0.345\linewidth]{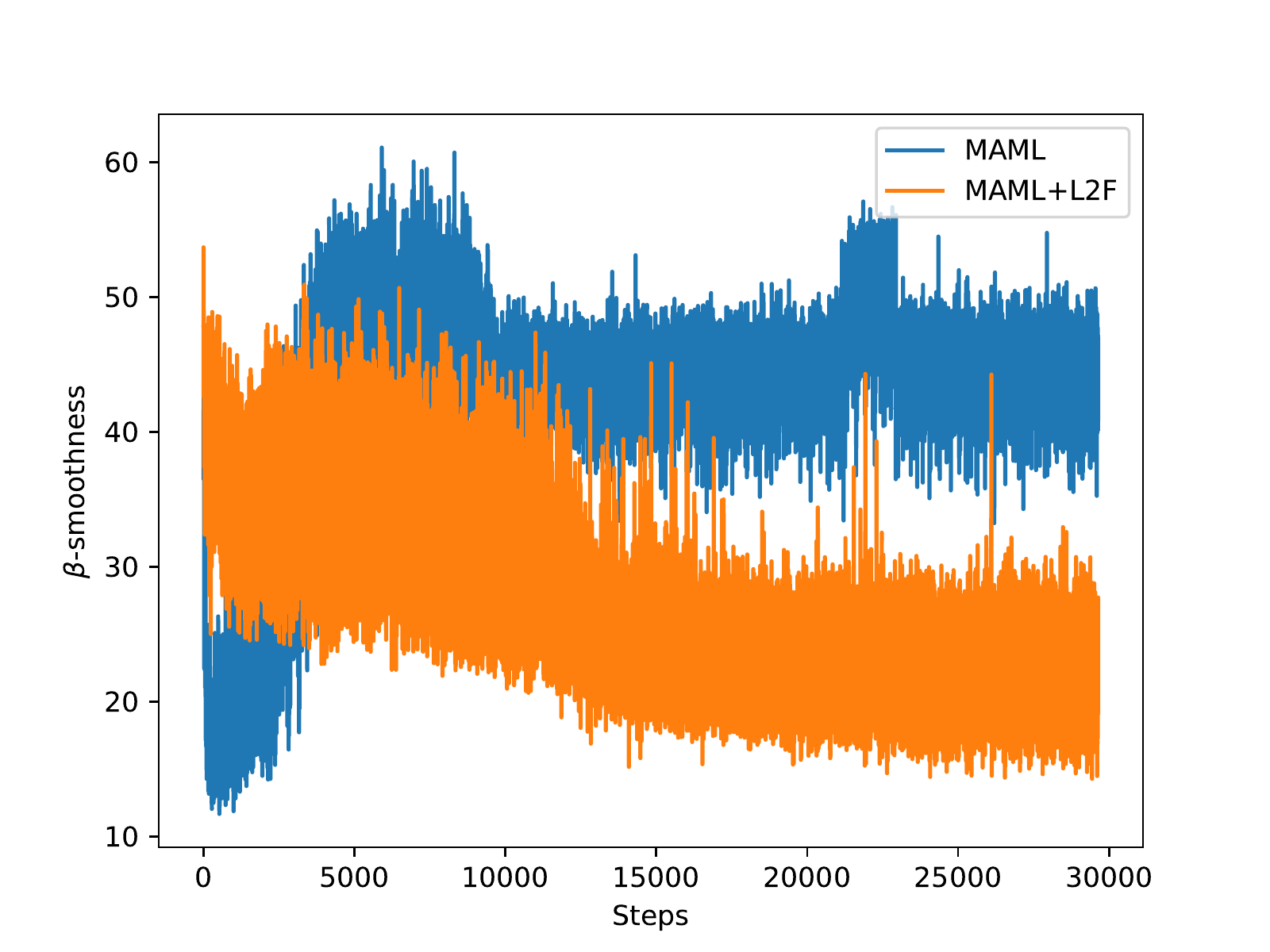}
    \label{fig:beta}
}
\end{center}
\vspace{-1em}
\caption{Visualization of optimization landscape: In \cite{santurkar2018how}, they analyze the stability and smoothness of the optimization landscape by measuring Lipschitzness and the ``effective'' $\beta$-smoothness of loss. 
We use these measurements to analyze learning dynamics for both MAML and our proposed method during training on 5-way 5-shot miniImageNet classification tasks, i.e. investigating fast-adaptation (or inner-loop) optimization. 
At each inner-loop update step, we measure variations in loss (a), the $l_2$ difference in gradients (b), and the maximum difference in gradient over the distance (c) as we move to different points along the computed gradient for that gradient descent. We take an average of these values over the number of inner-loop updates and plot them against training iterations.
The thinner shade in plots (a) and (b) and the lower the values in plot (c) indicate the smoother loss landscape and thus less training difficulty \cite{santurkar2018how}.}
\vspace{-0.35cm}
\label{fig:landscape}
\end{figure*}

The assumption of the existence of a task distribution may justify MAML for seeking a common initialization among tasks. 
But, there still exists variations among tasks, some of which may lead to the disagreement among tasks on the location of the initialization. We call such disagreement \textit{conflict} and formally define it in this paper. 
Some of prior knowledge encoded in such compromised initialization is useful for one task but may be irrelevant or even detrimental for another.
Consequently, a learner struggles to learn new concepts quickly with the prior knowledge that conflicts with information from new examples, as illustrated in Figure ~\ref{fig:conflict_analysis}. Such learning difficulty can manifest as the sharp loss landscape and thereby poor generalization to new examples~\cite{li2018visualizing,santurkar2018how}. Motivated by our hypothesis, we analyze and indeed observe the sharp landscape during fast adaptation to new examples (as shown in Figure \ref{fig:landscape}) and suggest that the learned initialization by MAML is a ``bad'' location. 

One solution for a meta-learner would be to simply \textit{forget} the part of the initialization that hinders adaptation to the task, minimizing its influence. 
This raises two questions: Where do these \textit{conflicts} occur? To what extent?
We hypothesize that the degree of conflict varies among layers of a neural network, especially CNN, since deeper layers learn more task-specific knowledge or class-specific knowledge in classification~\cite{zeiler2014visualizing}. 
To test the hypothesis, we measure \textit{conflict} at each layer and observe that \textit{conflict} is indeed more severe at deeper layers, as shown in Figure~\ref{fig:maml_analysis}\subref{fig:layer_conflict}.
We also observe that the amount of agreement between the learned initialization and the initialization desired by a given task differs for each task in Figure~\ref{fig:maml_analysis}\subref{fig:task_conflict}. 
Thus, we argue that \textit{conflicts} occur at two levels: task and layer.

Motivated by the observation, we propose to learn selective \textit{forgetting} by applying a task-and-layer-wise \textit{attenuation} on MAML initialization, controlling the influence of prior knowledge for each task and layer. 
For each task, we argue that initialization weights and its gradients (obtained from support examples of task), together, encode information about optimization specific to a task, and thus propose to condition on them to generate attenuation parameters. 
As for layer-wise attenuation, we generate an attenuation parameter for each layer. 
The proposed method, named L2F (Learn to Forget), indeed improves the quality of the initialization (illustrated by a smoother loss landscape in Figure~\ref{fig:landscape}) and consistent performance improvement across different domains, managing to maintain the simplicity and generalizability of MAML.
\section{Related Work}
Meta-learning aims to learn across-task prior knowledge to achieve fast adaptation to specific tasks~\cite{bengio1992optimization,hochreiter2001learning,schmidhuber1987evolutionary,schmidhuber1992learning,thrun2012learning}. 
Recent meta-learning systems can be broadly classified into three categories: metric-based, network-based, and optimization-based. 
The goal of metric-based system is to learn relationship between query and support examples by learning an embedding space, where similar classes are closer and different classes are further apart~\cite{koch2015siamese,NIPS2017_6996,Sung_2018_CVPR,vinyals2016matching}. 
Network-based approaches encode fast adaptation into network architecture, for example, by generating input-conditioned weights~\cite{munkhdalai2017meta,oreshkin2018tadam} or employing an external memory~\cite{munkhdalai2018rapid,santoro2016meta}. 
On the other hand, optimization-based systems adjust optimization for fast adaptation~\cite{finn2017model,ravi2017optimization,nichol2018first}.

Among optimization-based systems, MAML~\cite{finn2017model} has recently received interests, owing to its simplicity and generalizability. 
The generalizability stems from its model-agnostic algorithm that learns across-task initialization. 
The initialization aims to encode prior knowledge that helps the model quickly learn and achieve good generalization performance over tasks on average. 
While MAML boasts the simplicity, it shows relatively low performance on few-shot learning.

There has been several works that tried to improve the performance, especially on few-shot classification~\cite{antoniou2019how,lee2017gradient,Li2017meta,zhang2018metagan,jiang2019learning}. 
However, none of these methods tackles the problem with the sharing of the starting point of adaptation to different tasks. 
Recently, there has been a few works~\cite{oreshkin2018tadam,andrei2019meta,vuorio2018toward} that try to achieve task-wise model or initialization through their proposed task embeddings.
The metric-based system has a similar issue with MAML, and thus TADAM~\cite{oreshkin2018tadam} proposes to learn task embeddings, which are then used to generate affine transformation parameters that transform the features.

In this work, we focus on analyzing the problems of MAML and improving its performance, while maintaining its generalizability.  LEO~\cite{andrei2019meta} tries to solve the issue with the shared initialization by learning task embeddings through relation network, which are then used to generate input-dependent initializations in low-dimensional latent space. 
Another work that tries to relax the constraint on sharing the initialization is Multimodal MAML~\cite{vuorio2018toward}, where they propose to learn task embeddings and transform the MAML initialization with affine parameters. 

In contrast to~\cite{vuorio2018toward,andrei2019meta} that only focus on making the initialization task-dependent, we approach the problem from the perspective of optimization and provide a new insight that the quality of MAML initialization is compromised due to \textit{conflicts} among tasks on the location of the initialization in optimization landscape. 
Such compromised initialization will hinder fast adaptation and is illustrated by sharp loss landscape in Figure \ref{fig:landscape}. 
Motivated by the phenomenon of \textit{conflicts}, we argue that we only need to attenuate (\textit{forget}) the compromised part of the initialization. 
In fact, a large portion of the performance boost comes from the attenuation, not from the task-conditioned transformation (see Table \ref{tab:attenuation}). 

From the perspective of optimization, we also provide more effective and efficient task embedding.
Previous works~\cite{andrei2019meta,vuorio2018toward} try to achieve task-wise initializations through learning task embeddings directly from the input. However, learning such task embeddings without any task label is difficult and require specialized techniques, such as relation network~\cite{andrei2019meta} and metric learning~\cite{oreshkin2018tadam} that may not be applicable in other complex problems such as in reinforcement learning. 
We argue and observe that the amount of \textit{conflicts} varies among tasks, hinting that \textit{conflicts} can be used to identify tasks.
Since \textit{conflicts} between the desired initialization by task and the learned initialization can be described with gradients (see Section~\ref{sec:conflict}), we demonstrate that gradients itself give task-specific optimization information and thus can be used to represent tasks.
Because gradients are easily obtainable and model-agnostic, not only do we achieve effective task-wise initialization but also manage to maintain the simplicity and generalizability of MAML.

Overall, our proposed method greatly improves the performance of MAML while managing to maintain the simplicity and generalizability of MAML. 
Owing to its generalizability, we further show that not only does our method demonstrate a consistent improvement across domains, including reinforcement learning; but also our method can be easily applied to other MAML-based methods.
\section{Proposed Method}
\subsection{Problem Formulation}

\begin{figure*}
\begin{center}
\subfloat[Degree of conflict at each layer]{
    \includegraphics[width=0.234\linewidth]{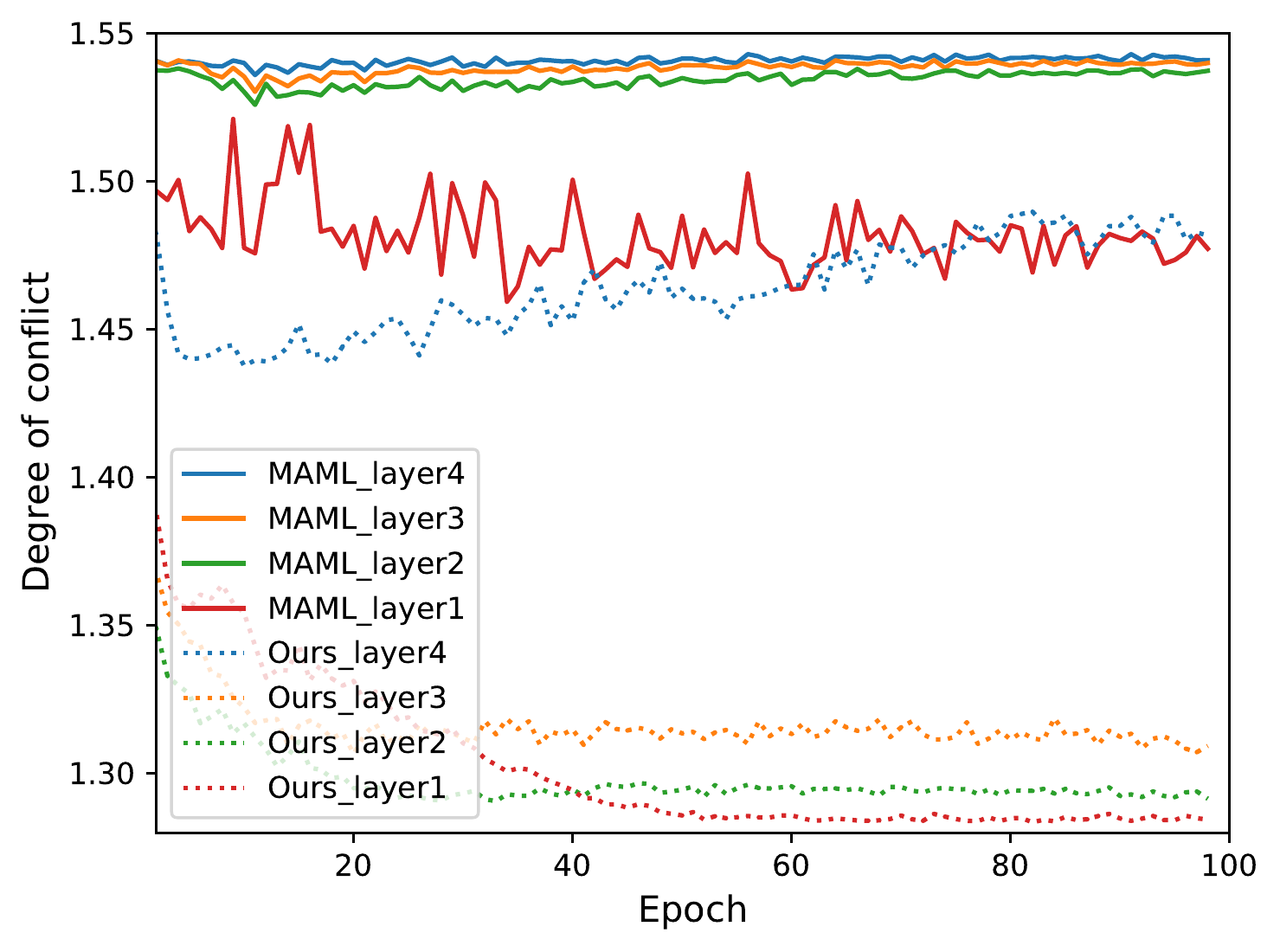}
    \label{fig:layer_conflict}
}
\subfloat[Manual $\gamma$ for each layer]{
    \includegraphics[width=0.226\linewidth]{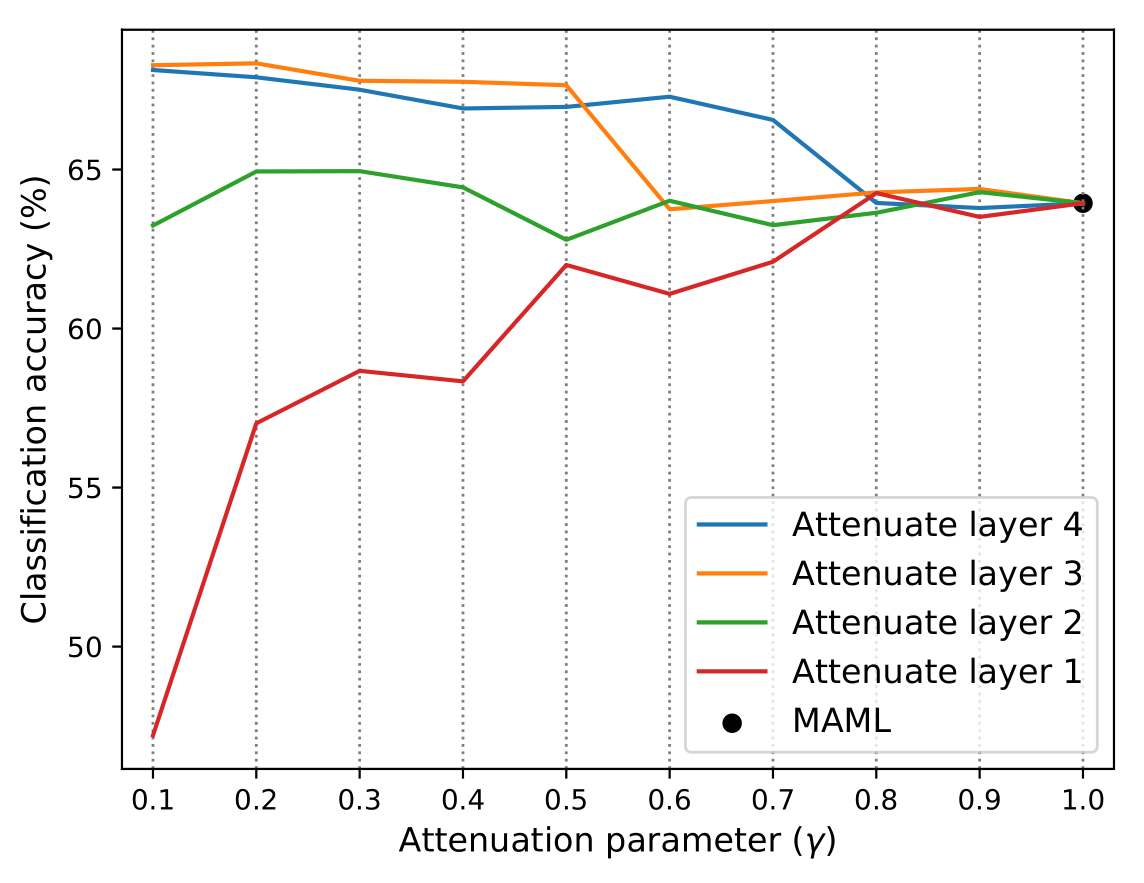}
    \label{fig:gamma_exp}
}
\subfloat[Degree of conflict for each task]{
    \includegraphics[width=0.232\linewidth]{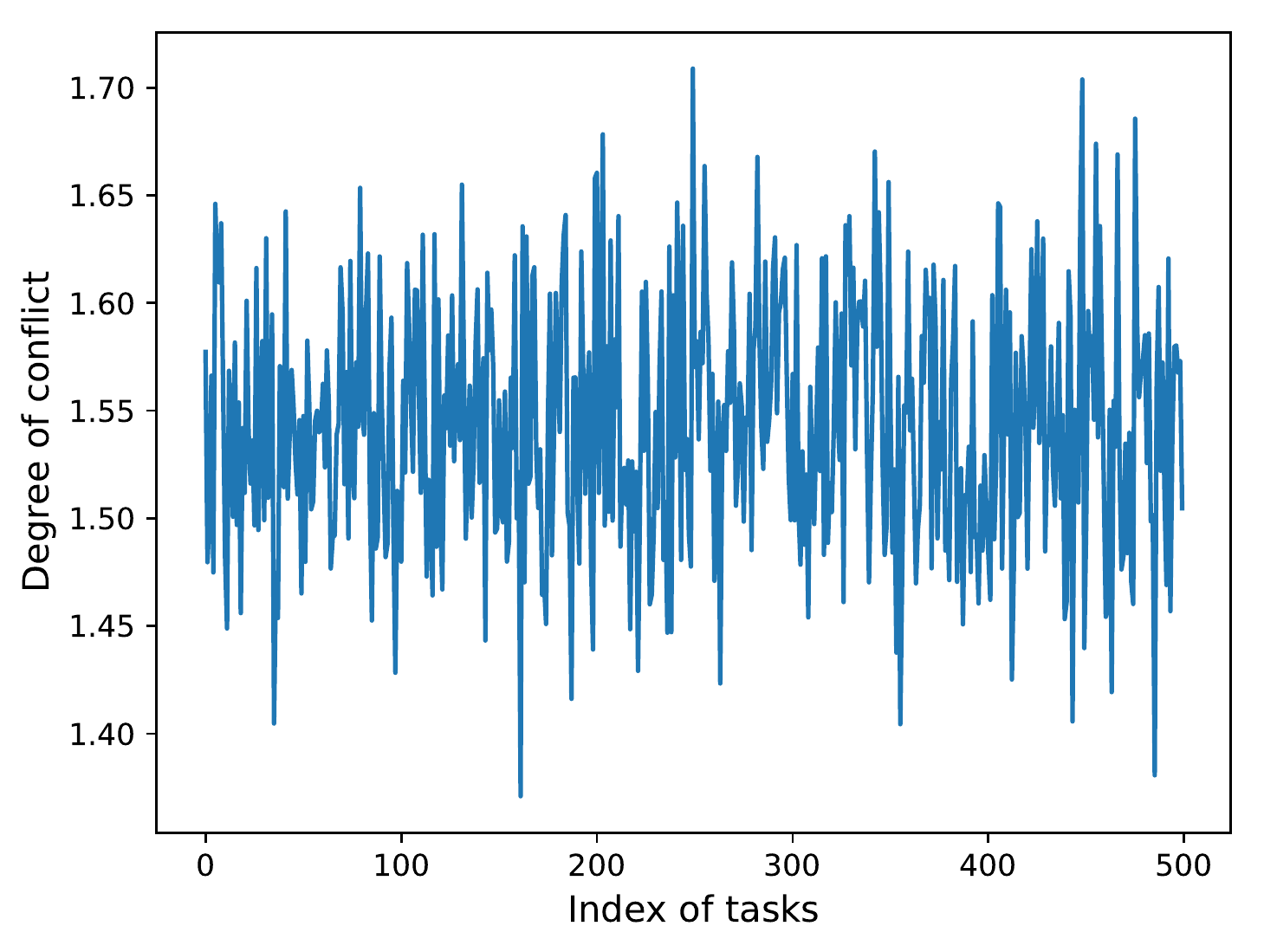}
    \label{fig:task_conflict}
}
\subfloat[Generated $\gamma$ for each task]{
    \includegraphics[width=0.23\linewidth]{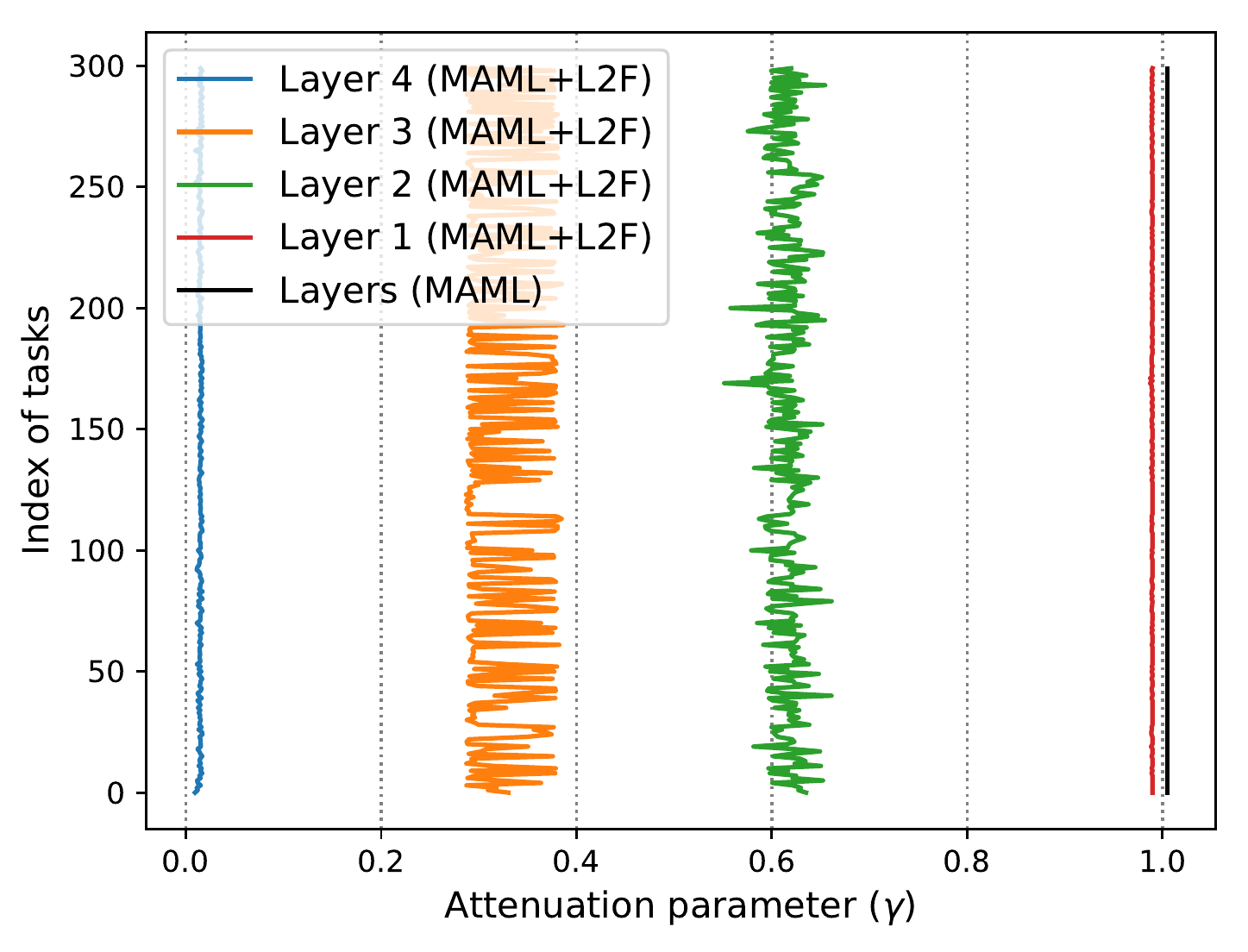}
    \label{fig:gamma_visual}
}
\end{center}
\vspace{-1em}
\caption{Analysis on degree of conflict and attenuation: (a) Throughout training, degree of conflict is measured and observed to vary among layers. For MAML, deeper layers exhibit greater extent of conflict, which aligns with the observation that deeper layers encode more task specific features~\cite{zeiler2014visualizing}. After applying L2F to MAML, conflict is observed to have decreased greatly. (b) Manual attenuation of an initialization by different levels (the lower $\gamma$, the stronger attenuation) for each layer affects the classification accuracy of a 4-layer CNN on miniImageNet. The figure suggests that deeper layers prefer the stronger attenuation. This supports our argument that the larger degree of conflict suggests the initialization quality is more compromised and that the compromised part needs to be minimized. (c) The degree of conflict between each meta-train task and the MAML initialization is observed to vary. This indicates that the amount of prior knowledge that is useful is different for each task. (d) Different attenuation parameters $\gamma$ are generated by the proposed method for each meta-test task, especially for middle-level layers. This suggests that the degree of conflict varies for each task, especially in middle-level layers.}
\vspace{-0.2cm}
\label{fig:maml_analysis}
\end{figure*}
Before introducing the proposed method, we start with the formulation of a generic meta-learning algorithm. We assume there is a distribution of tasks $p(\mathcal{T})$, from which meta-learning algorithm aims to learn the prior knowledge, represented by a model with parameters $\theta$. 
Tasks, each of which is sampled from $p(\mathcal{T})$, are split into three disjoint sets: meta-training set, meta-validation set, and meta-test set. In $k$-shot learning, a task $\mathcal{T}_i$ is first sampled from the meta-training set, followed by sampling $k$ number of examples $\mathcal{D}_{\mathcal{T}_i}$ from $\mathcal{T}_i$. 
These $k$ examples are then used to quickly adapt a model with parameters, $\theta$. 
Then, new examples $\mathcal{D}^\mathcal{'}_{\mathcal{T}_i}$ are sampled from the same task $\mathcal{T}_i$ to evaluate the generalization performance on unseen examples with the corresponding loss function, $\mathcal{L}_{\mathcal{T}_i}$. 
The feedback from the loss is then used to adjust the model parameters $\theta$ to achieve better generalization. 
Finally, the meta-validation set is used for model selection, while the meta-test set is used for the final evaluation on the selected model.
\subsection{Model-Agnostic Meta-Learning}
To tackle the problem of fast adaptation to unseen tasks with few examples, we borrow the philosophy and the methodology from MAML~\cite{finn2017model}. 
MAML encodes prior knowledge in an initialization and seeks for a ``good'' common initial set of values for weights of a neural network across tasks. 
Formally, given a network $f_\theta$ with weights $\theta$, MAML learns a set of initial weight values, $\theta$, which will serve as a good starting point for fast adaptation to a new task $\mathcal{T}_i$, sampled from a task distribution $p(\mathcal{T})$. 
Given few examples $\mathcal{D}_{\mathcal{T}_i}$ and a loss function $\mathcal{L}_{\mathcal{T}_i}$ from the task $\mathcal{T}_i$, the network weights are adapted to $\mathcal{T}_i$ during inner-loop update as follows:
\begin{equation}
\theta'_i = \theta - \alpha\nabla_{\theta} \mathcal{L}^{\mathcal{D}_{\mathcal{T}_i}}_{\mathcal{T}_i}(f_\theta).
\label{eq:inner_update}
\end{equation}
To give feedback on the generalization performance of the model with adapted weights $\theta’_i$ to each task, the model is evaluated on new examples, $\mathcal{D}^{\mathcal{'}}_{\mathcal{T}_i}$ sampled from the same task $\mathcal{T}_i$. 
The feedback, manifested in the form of loss gradients, is used to update the initialization $\theta$ so that better generalization is achieved:
\begin{equation}
\theta  \leftarrow \theta - \eta \nabla_{\theta} \sum_{\mathcal{T}_i}{\mathcal{L}^{\mathcal{D}^\mathcal{'}_{\mathcal{T}_i}}_{\mathcal{T}_i}(f_{\theta'_i})}.
\label{eq:outer_update}
\end{equation}
\begin{algorithm}[t]
\caption{Proposed Meta-Learning}
\label{algo:algorithm}
\begin{algorithmic}[1]
\REQUIRE Task distribution $p(\mathcal{T})$
\REQUIRE Learning rates $\alpha, \eta$
\STATE Randomly initialize $\theta, \phi$
\STATE Let $\theta$ = $\{\theta^j\}^{j=1...l}$ where $j$ is the layer index and $l$ is the number of layers of a network
\WHILE{not converged}
	\STATE Sample a batch of tasks $\mathcal{T}_i \sim p(\mathcal{T})$
	\FOR {each task $\mathcal{T}_i$}    
		\STATE Sample examples $(\mathcal{D}_{\mathcal{T}_i}, \mathcal{D}^\mathcal{'}_{\mathcal{T}_i})$ from $\mathcal{T}_i$
		\STATE Compute $\mathcal{L}^{\mathcal{D}_{\mathcal{T}_i}}_{\mathcal{T}_i}(f_\theta)$ by evaluating $\mathcal{L}_{\mathcal{T}_i}$ with respect to $\mathcal{D}_{\mathcal{T}_i}$ 
		\STATE Compute attenuation parameter $\gamma$ for each layer: \{$\gamma_i^j\}^{j=1...l} = g_\phi(\nabla_{\theta} \mathcal{L}^{\mathcal{D}_{\mathcal{T}_i}}_{\mathcal{T}_i}(f_{\theta}))$, 
		\STATE Compute attenuated initialization: $\bar{\theta}_i^j = \gamma_i^j \theta^j$
		\STATE Initialize $\theta'_i = \{\bar{\theta}_i^j\}^{j=1...l}$
		\FOR {number of inner-loop updates}
			\STATE Compute $\mathcal{L}^{\mathcal{D}_{\mathcal{T}_i}}_{\mathcal{T}_i}(f_{\theta'_i})$ by evaluating $\mathcal{L}_{\mathcal{T}_i}$ with respect to $\mathcal{D}_{\mathcal{T}_i}$
			\STATE Perform gradient descent to compute adapted weights: $\theta'_i = \theta'_i - \alpha\nabla_{\theta'_i}\mathcal{L}^{\mathcal{D}_{\mathcal{T}_i}}_{\mathcal{T}_i}(f_{\theta'_i})$
		\ENDFOR
		\STATE Compute $\mathcal{L}^{\mathcal{D}^\mathcal{'}_{\mathcal{T}_i}}_{\mathcal{T}_i}(f_{\theta'_i})$ by evaluating $\mathcal{L}_{\mathcal{T}_i}$ with respect to $\mathcal{D}^\mathcal{'}_{\mathcal{T}_i}$
	\ENDFOR
	\STATE Perform gradient descent to update weights: $(\theta, \phi) \leftarrow (\theta, \phi) - \eta \nabla_{(\theta, \phi)}\sum_{\mathcal{T}_i}\mathcal{L}^{\mathcal{D}^\mathcal{'}_{\mathcal{T}_i}}_{\mathcal{T}_i}(f_{\theta'_i})$
\ENDWHILE
\end{algorithmic}
\end{algorithm}
\vspace{-0.35cm}
\subsection{Definition of Conflict}\label{sec:conflict}
While MAML is elegantly simple, its limitation comes from the very fact that the initialization is shared across a distribution of tasks. 
Despite the goal of MAML, which is to learn a ``good'' starting point for fast adaptation to new tasks, the shared initialization, in fact, hinders the fast learning process. This is illustrated by sharp optimization landscape during fast adaptation in Figure \ref{fig:landscape}.
This is mainly due to disagreement between tasks on the location of a ``good'' starting point. We call such disagreement \textit{conflict}. 

At each training iteration, each task $\mathcal{T}_i$ takes the initialization closer to the desired location via gradient: $\boldsymbol{u}_i = -\nabla_{\theta}\mathcal{L}^{\mathcal{D'}}_{\mathcal{T}_i}(f_{\theta'_i})$ during meta-update. 
However, since MAML shares the initialization, the update is made via gradients accumulated over a batch of tasks $\sum_{i}{\boldsymbol{u_i}}$ as in Equation (\ref{eq:outer_update}).
Hence, in the example of two tasks, the \textit{conflict} occurs between tasks $\mathcal{T}_i$ and $\mathcal{T}_j$ when their gradient directions, i.e. directions of $\boldsymbol{u}_i$ and $\boldsymbol{u}_j$, differ. 
The more their directions differ, the more the initialization update diverges from $\boldsymbol{u}_i$ and $\boldsymbol{u}_j$, pointing towards the location that is not desirable for both $\mathcal{T}_i$ and $\mathcal{T}_j$. 
We refer to this phenomenon as \textit{compromise in the initialization}. 

We define \textit{the degree of conflict} among tasks to be the average angle between $\boldsymbol{u}_i$ and $\sum_{i}{\boldsymbol{u}_i}$, which is measured as the average absolute arccosine of the dot product of the normalized vectors, $\mathbb{E}_{\mathcal{T}_i \sim p(\mathcal{T})}[\left|cos^{-1}(\boldsymbol{\hat{u}}_i \cdot \boldsymbol{v}) \right|]$, where $\boldsymbol{\hat{u}}_i$ is $\frac{\boldsymbol{u}_i}{\norm{\boldsymbol{u}_i}}$ and $\boldsymbol{v}$ is $\frac{ \sum_{i}{\boldsymbol{u}_i}}{\norm{ \sum_{i}{\boldsymbol{u}_i}}}$.
Figure \ref{fig:maml_analysis}\subref{fig:layer_conflict} measures the \textit{degree of conflict} at each epoch and demonstrates that the \textit{conflict} is indeed more prominent in deeper layers, which aligns with the observation that the deeper layers encode more task-specific features~\cite{zeiler2014visualizing}. 
\subsection{Learning to Forget}
When \textit{the degree of conflict} is high, we say the initialization is more \textit{compromised}, and hence the more difficult it is to learn new tasks quickly, as illustrated by sharp loss landscape in Figure \ref{fig:landscape}. 
This suggests that the learner finds some part of the initialization to be irrelevant or even detrimental for learning a given task. 
We thus propose to discard such compromised part of the prior knowledge via attenuating the initialization parameters $\theta$ directly. 
Then, one may ask which parameter is compromised? 

To answer the question, we refer to the previous finding that lower layers of a CNN encode general knowledge while deeper layers contain more task-specific information~\cite{zeiler2014visualizing}. 
Upon this observation, we hypothesize that lower layers do not need much attenuation while deeper layers do. 
To support our hypothesis, we perform an experiment, shown in Figure \ref{fig:maml_analysis}\subref{fig:gamma_exp}, where we vary the amount of attenuation ($\gamma^j$) on each layer to observe how much each layer benefits. 
As expected, deeper layers favor stronger attenuation while lower layers prefer little to no attenuation.  
This leads to the second question: How much should the parameters be attenuated layer-wise? 

One answer would be to let a model learn to find an optimal set of attenuations. The answers to these two questions lead to our proposal: learn layer-wise attenuation via applying a single learnable parameter $\gamma^j$ on the initialization parameters of each layer $\theta^j$ as follows:
\begin{equation}
\bar{\theta^j} = \gamma^j \theta^j,
\label{eq:affine_transform}
\end{equation}
where $j$ is the layer index of a neural network. 
The attenuated initialization $\bar{\theta}$ serves as a new starting point for fast adaptation to tasks.
Although this may reduce the extent of compromise that may exist in the original MAML initialization, one may ask if the amount of unnecessary or contradicting information in the initialization is equal across tasks. 

Surely, the degree of agreement and disagreement with others differs for different tasks. 
This can be observed in Figure \ref{fig:maml_analysis}\subref{fig:task_conflict}, where the measured degree of conflict is observed to vary for each task. 
As a result, there is no consensus between tasks on what the best attenuation is for layer 2, as indicated by different attenuation preferred by each task in Figure \ref{fig:maml_analysis}\subref{fig:gamma_visual}. 
To resolve such conflict, in addition to the layer-wise attenuation, we propose a task-dependent attenuation. 
But, this poses another question: What information can be used to make attenuation task-dependent? 

We turn to gradients $\nabla_{\theta} \mathcal{L}^{\mathcal{D}_{\mathcal{T}_i}}_{\mathcal{T}_i}(f_{\theta})$ for the answer. Gradients, used for fast adaptation via gradient descents, not only hold task-specific information but also encode the quality of the initialization with respect to the given task $\mathcal{T}_i$ from the perspective of optimization. Thus, we propose to compute gradient $\nabla_{\theta} \mathcal{L}^{\mathcal{D}_{\mathcal{T}_i}}_{\mathcal{T}_i}(f_{\theta})$ at the initialization and condition a network $g_\phi$ on it to generate the task-dependent attenuation:
\begin{equation}
\gamma_i = g_{\phi}(\nabla_{\theta} \mathcal{L}^{\mathcal{D}_{\mathcal{T}_i}}_{\mathcal{T}_i}(f_{\theta})), 
\label{eq:affine_params}
\end{equation}
where $\gamma_i = \{\gamma_i^j\}$ is the set of layer-wise gammas for the $i$-th task and $g_\phi$ is a 3-layer MLP network of parameters $\phi$, with a sigmoid at the end to facilitate attenuation. 
For the network $g_\phi$ to generate layer-wise gammas, the network is conditioned on the layer-wise mean of gradients.

After the initialization is adapted to each task, the network undergoes fast adaptation as in Equation (\ref{eq:inner_update}) and the initialization is updated as in Equation (\ref{eq:outer_update})  during training. 
The overall training procedure is summarized in Algorithm~\ref{algo:algorithm}.

\section{Experiments} \label{experiment}
\begin{table}[b]
\vspace{-0.3cm}
   \small 
   \centering
   \scalebox{0.65}{
   \begin{threeparttable}
   \begin{tabular}{lccccc} 
   \toprule[\heavyrulewidth]\toprule[\heavyrulewidth]
          &\multirow{2}{*}{\textbf{Backbone}}& \multicolumn{2}{c}{\textbf{miniImageNet}}\\ 
          &&1-shot&5-shot\\
   \midrule
   Matching Network~\cite{vinyals2016matching}& 4 conv & $43.44 \pm 0.77\%$ & $55.31 \pm 0.73\%$\\
   Meta-Learner LSTM~\cite{ravi2017optimization} & 4 conv & $43.56 \pm 0.84\%$ & $60.60 \pm 0.71\%$\\
   MetaNet~\cite{munkhdalai2017meta} & 5 conv & $49.21 \pm 0.96\%$ & $-$\\
   LLAMA~\cite{grant2018recasting} & 4 conv & $49.40 \pm 0.84\%$ & $-$\\
   Relation Network~\cite{Sung_2018_CVPR} & 4 conv & $50.44 \pm 0.82\%$ & $65.32 \pm 0.70\%$\\
   Prototypical Network~\cite{NIPS2017_6996} & 4 conv & $49.42 \pm 0.78\%$ & $68.20 \pm 0.66\%$\\
   \hdashline
   MAML~\cite{finn2017model}& 4 conv & $48.70 \pm 1.75\%$ & $63.11 \pm 0.91\%$\\
   MAML++~\cite{antoniou2019how}& 4 conv & $52.15 \pm 0.26\%$ & $68.32 \pm 0.44\%$\\
   MAML+L2F (Ours) & 4 conv & $52.10 \pm 0.50\%$ & $69.38 \pm 0.46\%$\\
   \midrule
   MetaGAN~\cite{zhang2018metagan}& ResNet12 & $52.71 \pm 0.64\%$ & $68.63 \pm 0.67\%$\\
   SNAIL~\cite{mishra2018simple} & ResNet12\tnote{*} & $55.71 \pm 0.99\%$ & $68.88 \pm 0.92\%$\\
   adaResNet~\cite{munkhdalai2018rapid} & ResNet12 & $56.88 \pm 0.62\%$ & $71.94 \pm 0.57\%$\\
   CAML~\cite{jiang2019learning} & ResNet12\tnote{*}& $59.23 \pm 0.99\%$ & $72.35 \pm 0.71\%$\\
   TADAM~\cite{oreshkin2018tadam}  & ResNet12\tnote{*} & $58.5 \pm 0.3\%$ & $76.7 \pm 0.3\%$\\
   \hdashline
   MAML & ResNet12 & $51.03 \pm 0.50\%$ & $68.26 \pm 0.47\%$\\
   MAML+L2F (Ours) & ResNet12 & $ 57.48 \pm 0.49\%$ & $74.68 \pm 0.43\%$\\
   \midrule
   LEO~\cite{andrei2019meta} &WRN-28-10\tnote{*}& $61.76 \pm 0.08\% $ & $77.59 \pm 0.12 \%$\\
   LEO (reproduced) &WRN-28-10\tnote{*}& $61.50 \pm 0.17\% $ & $77.12 \pm 0.07 \%$\\
   LEO+L2F (Ours) &WRN-28-10\tnote{*}& $\boldsymbol{62.12 \pm 0.13\%}$ & $\boldsymbol{78.13 \pm 0.15\%}$\\
   \bottomrule[\heavyrulewidth] 
   \end{tabular}
   \begin{tablenotes}
   \item[*] a pre-trained network.
   \end{tablenotes}
   \end{threeparttable}
   }
   \caption{Test accuracy on 5-way miniImageNet classification} 
   \label{tab:result_mini}
\end{table}
\begin{table}[t]
   \small 
   \centering
   \scalebox{0.8}{
   \begin{threeparttable}
   \begin{tabular}{lccccc}
   \toprule[\heavyrulewidth]\toprule[\heavyrulewidth]
          &\multirow{2}{*}{\textbf{Backbone}}& \multicolumn{2}{c}{\textbf{tieredImageNet}} \\ 
          &&1-shot&5-shot\\
   \midrule
   MAML & 4 conv & $49.06 \pm 0.50\%$ & $67.48 \pm 0.47\%$\\
   MAML+L2F (Ours) & 4 conv & $54.40 \pm 0.50\%$ & $73.34 \pm 0.44\%$\\
   MAML & ResNet12  & $58.58 \pm 0.49\%$ & $71.24 \pm 0.43\%$\\
   MAML+L2F (Ours) & ResNet12 & $63.94 \pm 0.48\%$ & $77.61 \pm 0.41\%$\\
   \midrule
   LEO &WRN-28-10\tnote{*}& $66.33 \pm 0.05\%$ & $81.44 \pm 0.09\%$\\
   LEO (reproduced) &WRN-28-10\tnote{*}& $67.02 \pm 0.11\%$ & $82.29 \pm 0.16\%$\\
   LEO+L2F (Ours) &WRN-28-10\tnote{*}& $\boldsymbol{68.00 \pm 0.11\%}$ & $\boldsymbol{83.02 \pm 0.08\%}$\\
   \bottomrule[\heavyrulewidth] 
   \end{tabular}
   \begin{tablenotes}
   \item[*] a pre-trained network.
   \end{tablenotes}
   \end{threeparttable}
   }
   \caption{Test accuracy on 5-way tieredImageNet classification} 
   \label{tab:result_tiered}
   \vspace{-0.5cm}
\end{table}
In this section, we demonstrate the effectiveness and generalizability of our method through extensive experiments on various problems, including few-shot classification, regression, and reinforcement learning. 
\subsection{Few-Shot Classification}
Two well-known datasets, miniImageNet and tieredImageNet are used for the classification test, both of which are extracted from ImageNet dataset while taking into account for few-learning scenarios. 
miniImageNet is constructed by randomly selecting 100 classes from the ILSVRC-12 dataset, with each class consisting of 600 images of size 84 $\times$ 84~\cite{vinyals2016matching}. 
The constructed dataset is divided into 3 disjoint subsets: 64 classes for training, 16 for validation, and 20 for test as in~\cite{ravi2017optimization}.

tieredImageNet is a larger subset with 608 classes with 779,165 images of size 84 $\times$ 84 in total. 
Classes are grouped into 34 categories, according to ImageNet hierarchy. 
These categories are then split into 3 disjoint sets: 20 categories for training, 6 for validation, and 8 for test. 
According to~\cite{ren2018meta}, this minimizes class similarity between training and test and thus makes the problem more challenging and realistic.  
Experiments for tieredImageNet and miniImageNet are conducted under typical settings: 5-way 1-shot and 5-way 5-shot classification. For more experiments on other datasets, such as FC100~\cite{oreshkin2018tadam}, CIFAR-FS~\cite{bertinetto2019meta}, and Meta-Dataset~\cite{triantafillou2018meta}, please see the supplementary materials.
\vspace{-0.3cm}
\subsubsection{Results}
\vspace{-0.3cm}
The results of our proposed approach, other baselines and existing state-of-the-art approaches on the miniImageNet and tieredImageNet are presented in Table \ref{tab:result_mini} and Table \ref{tab:result_tiered}, respectively. 
The proposed method improves MAML by a large margin. 
We note that our proposed approach remains model-agnostic and achieves better or comparable accuracy to the state-of-the-art approaches with the same backbone, even without fine-tuning.
To show generalization of the contribution, we apply L2F to the state-of-the-art MAML-based system LEO and demonstrate the performance improvement, achieving the new state-of-the-art performance.
\vspace{-0.5cm}
\subsubsection{Ablation Studies}
\begin{table}[H]
   \vspace{-0.3cm}
   \small
   \centering
   \begin{tabular}{ccc}
   \toprule[\heavyrulewidth]\toprule[\heavyrulewidth]
          Inner-loop&\multirow{2}{*}{MAML} &\multirow{2}{*}{MAML+L2F(Ours)}\\
          update steps&&\\
   \midrule
   1 & $56.93 \pm 0.32\%$ & $68.16 \pm 0.47\%$\\
   2 & $55.63 \pm 0.50\%$ & $66.85 \pm 0.49\%$\\
   3 & $58.79 \pm 0.49\%$ & $68.61 \pm 0.46\%$\\
   4 & $62.72 \pm 0.45\%$ & $68.66 \pm 0.43\%$\\
   5 & $63.94 \pm 0.41\%$ & $69.38 \pm 0.46\%$\\
   6 & $64.54 \pm 0.46\%$ & $-$\\
   \bottomrule[\heavyrulewidth] 
   \end{tabular}
   \caption{Ablation studies on inner-loop update steps on  5-way 5-shot miniImageNet classification.} 
   \label{tab:inner_step}
   \vspace{-0.3cm}
\end{table}
\textbf{Inner-loop update steps}
One may argue that the comparisons are not fair because there is one extra adjustment to initialization parameters before inner-loop updates. Table \ref{tab:inner_step} shows ablation studies on the number of inner-loop updates for the proposed and the baseline to demonstrate that the performance gain is not due to an extra number of adjustments to parameters. Rather, the benefits come from \textit{forgetting} the unnecessary information, helping the learner quickly adapt to new tasks.
\begin{table}[b]
\vspace{-0.3cm}
   \small
   \centering
   \begin{tabular}{cc}
   \toprule[\heavyrulewidth]\toprule[\heavyrulewidth]
    Attenuation Scope & Accuracy\\
   \midrule
   None (MAML, our reproduction)& $63.94\pm0.48\%$\\
   parameter-wise & $64.7 \pm 0.43\%$\\
   filter-wise & $65.35 \pm 0.48\%$\\
   layer-wise & $\boldsymbol{68.49 \pm 0.41\%}$\\
   network-wise & $67.84 \pm 0.46\%$\\
   \midrule
   MAML+L2F (Ours) & $\boldsymbol{69.38 \pm 0.46\%}$\\
   \bottomrule[\heavyrulewidth] 
   \end{tabular}
   \caption{Ablation studies on attenuation scope. Except MAML+L2F, all models learn task-independent attenuation parameters to illustrate the effect of attenuation scope alone, without task-conditioning.} 
   \label{tab:attenuation}
\end{table}

\textbf{Attenuation Scope}
One may be curious and ask: Is layer-wise attenuation the best way to go? 
Thus, we analyze different scopes of attenuation; a single attenuation parameter for the whole network, or an individual attenuation parameter for each layer, each filter, and each weight of the network. 
To focus on investigating which scope of attenuation is most beneficial, we remove the task-dependent part and make the attenuation parameters learnable (with values initialized to be 1), rather than generated by the network $g_\phi$. 

We perform an ablation study with a 4-layer CNN in 5-way 5-shot classification setting on miniImageNet and present results in Table \ref{tab:attenuation}. 
As expected, the layer-wise attenuation gave the most performance gain. Weight-wise or filter-wise attenuation parameter may have finer control, but these parameters have limited scope in that they do not have information about conflicts that occur at the level of layers or network. 
On the other hand, layer-wise and network-wise parameters gain information about conflicts in neighbor weights as gradients pass through different weights/filters to reach the same attenuation parameter, since the attenuation parameter is shared by these weights/filters. In the meantime, network-wise parameters do not have enough control and thus perform worse than the layer-wise parameters. 
In the trade-off between control and information gain, layer-wise has shown to strike the right balance.

\textbf{Effect of Task-Conditioning}
Table \ref{tab:attenuation} reports lower performance of layer-wise attenuation model, compared to our full model, MAML+L2F. 
The only difference between the layer-wise attenuation model and ours is that the layer-wise attenuation model lacks the task-conditioning. 
One can observe that the most performance gain in our method comes from the attenuation, alluding to the importance of attenuation.
Regardless, the task-conditioning does improve the performance as well. 

\textbf{Representation of Task Embedding}
To verify that gradients contain high-quality information about tasks, we condition the network $g$ on the mean of class prototypes from the pre-trained prototypical network~\cite{NIPS2017_6996}(similar to TADAM~\cite{oreshkin2018tadam}) as task representation.
Table \ref{tab:input} demonstrates that our method with gradients as task representation performs similarly or slightly better than the one with the mean of class prototypes. 
This exhibits the effectiveness of gradients as task representation from the perspective of the optimization, especially because gradients are simple to obtain and model-agnostic while class prototypes are high-dimensional and not applicable across different domains.
\begin{table}[t]
   \small
   \centering
   \begin{tabular}{lcc}
   \toprule[\heavyrulewidth]\toprule[\heavyrulewidth]
          &\multicolumn{2}{c}{\textbf{miniImageNet}} \\ 
          &5-shot\\
   \midrule
   Features (class prototype) & $68.73\pm0.46\%$\\
   Gradients (Ours, MAML+L2F) & $69.48 \pm 0.46\%$\\
   \bottomrule[\heavyrulewidth] 
   \end{tabular}
   \caption{Ablation studies on types of representation for task embedding} 
   \label{tab:input}
\end{table}
  \textbf{Effect of Attenuation}
To analyze how much performance gain comes from each part of L2F (i.e. \textit{forgetting} and task-dependency), we apply each module separately to MAML and present results in Table \ref{tab:module}. 
Since the investigation on effectiveness of task-dependency has already been presented in Table \ref{tab:attenuation}, we now focus on the effectiveness of the attenuation, compared to other variant transformations. 
To that end, we explore different types of task-dependent transformations of the initialization. 
We start with the simple superset of the attenuation: $\gamma$ without sigmoid (Model 3) such that $\gamma_i$ is no longer restricted to be between $0$ and $1$, and hence does not facilitate attenuation. We also explore a more flexible option: affine transformation (Model 2), where the network $g_\phi$ generates two sets of parameters $\gamma_i, \delta_i$ without sigmoid, which will modulate $f_\theta$ via $\gamma^j_i \theta^j + \delta^j_i$.
\begin{table}[t]
   \small
   \centering 
   \scalebox{0.8}{
   \begin{tabular}{ccc} 
   \toprule[\heavyrulewidth]\toprule[\heavyrulewidth]
   Model & Description    & Accuracy\\
   \midrule
   1&MAML (our reproduction) &$63.94\pm0.48\%$\\
   2&MAML + task-dependent non-sigmoided $\gamma^j_i, \delta^j_i$ & $66.22 \pm 0.47\%$\\
   3&MAML + task-dependent non-sigmoided $\gamma^j_i$ & $67.56 \pm 0.47\%$\\
   \midrule
   Ours&MAML + L2F (task-dependent sigmoided $\gamma^j_i$) & $\boldsymbol{69.38 \pm 0.46\%}$\\
   \bottomrule[\heavyrulewidth] 
   \end{tabular}
   }
   \caption{Ablation studies on task-conditioned transformation to illustrate the effectiveness of attenuation.} 
   \label{tab:module}
   \vspace{-0.3cm}
\end{table}

Table \ref{tab:module} illustrates that MAML gains performance boost throughout different types of task-dependent transformation, suggesting the benefits of the task-dependency. 
It is reasonable to expect that more flexibility of transformation (Model 2 and 3) would allow for tasks to bring the initialization to more appropriate location for fast adaptation. Interestingly, the classification accuracy drops as more flexibility is given to the transformation of the initialization. This seeming contradiction underlines the necessity of attenuation (sigmoided $\gamma^j_i$ in our model), rather than just na\"ive transformation, of the initialization to \textit{forget} the compromised part of the prior knowledge encoded in the initialization.

We would like to stress that MAML with task-independent layer- or network-wise attenuation in Table~\ref{tab:attenuation} performs better than other task-conditioned transformations in Table~\ref{tab:module}.
This suggests that it is more important to \textit{forget} the compromised initialization than making it task-adaptive.

\subsection{Regression}
\label{sec:regression}

\begin{figure*}[t]
\begin{center}
\subfloat[5, 10-shot regression]{
\includegraphics[width=0.242\linewidth]{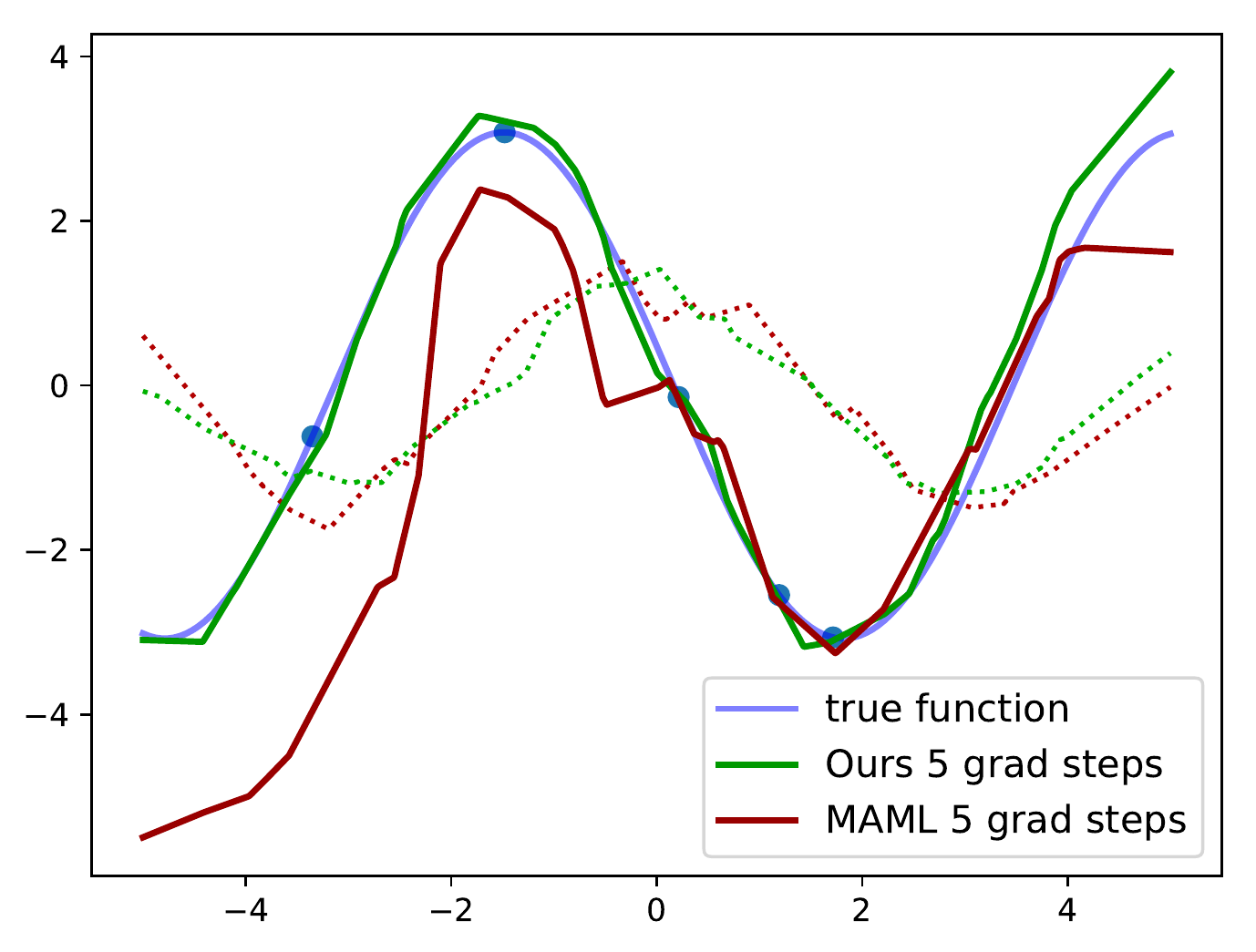}
\includegraphics[width=0.242\linewidth]{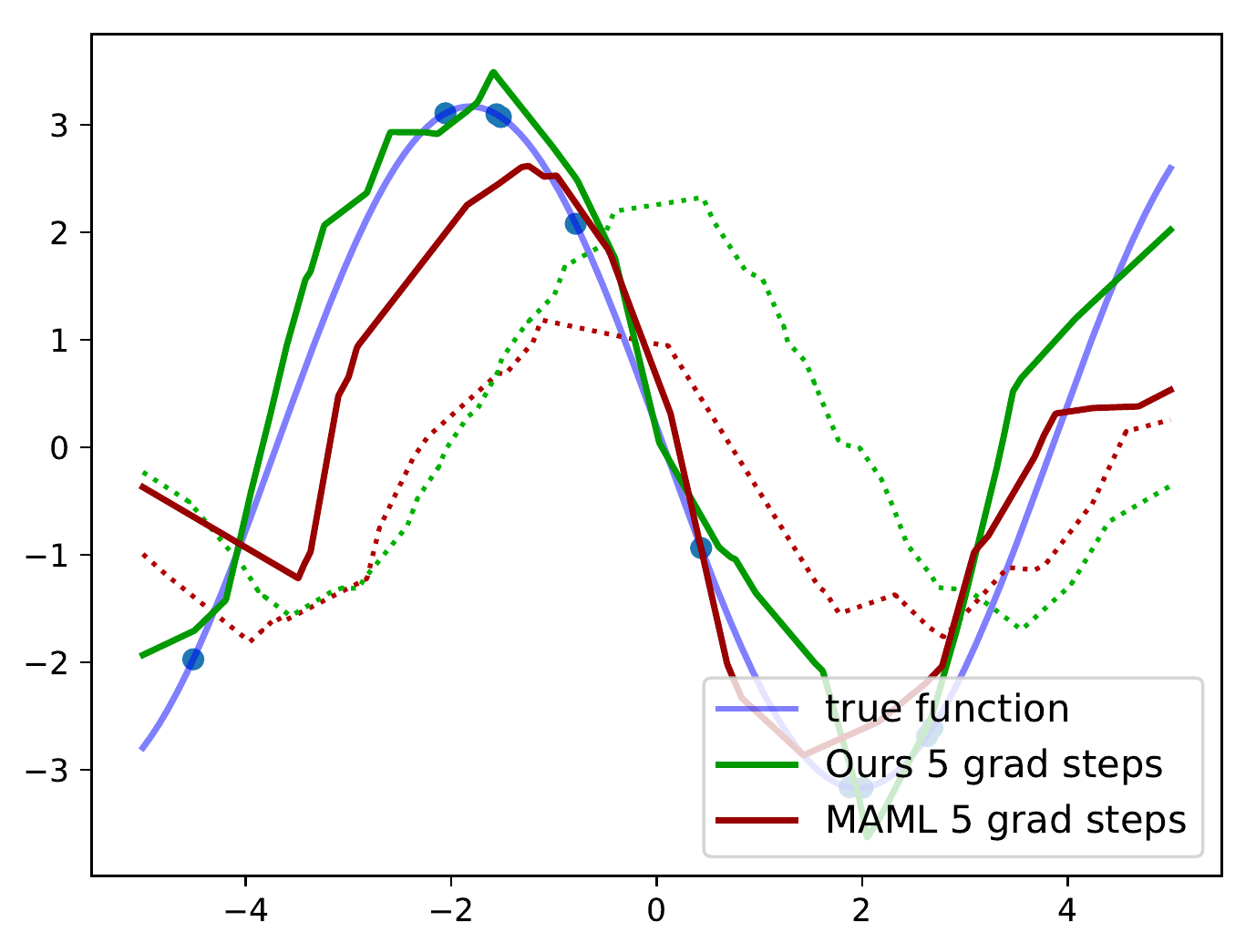}
\label{fig:regression_fig1}
}
\subfloat[Non-overlapped task distributions on 5-shot]{
\includegraphics[width=0.242\linewidth]{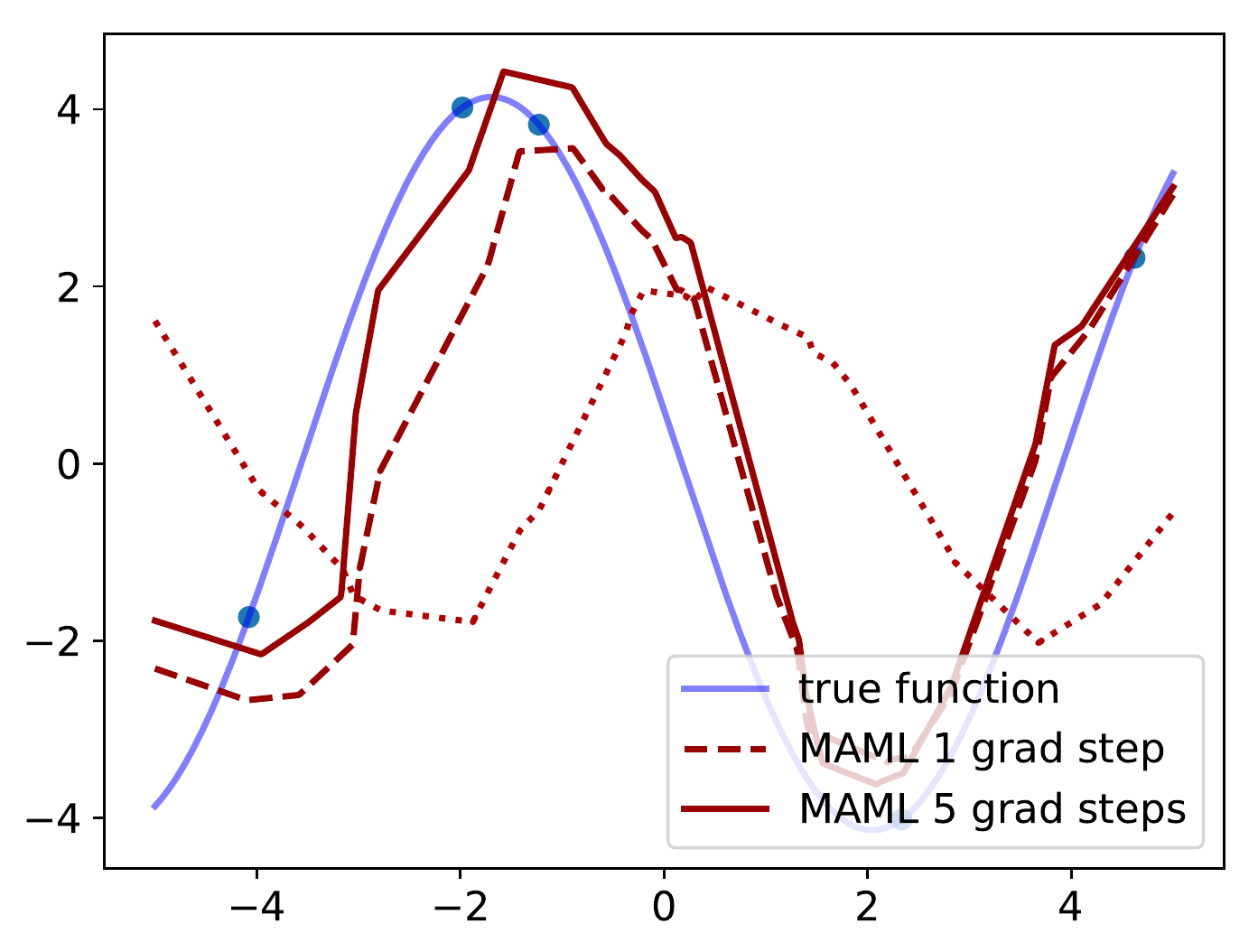}
\includegraphics[width=0.242\linewidth]{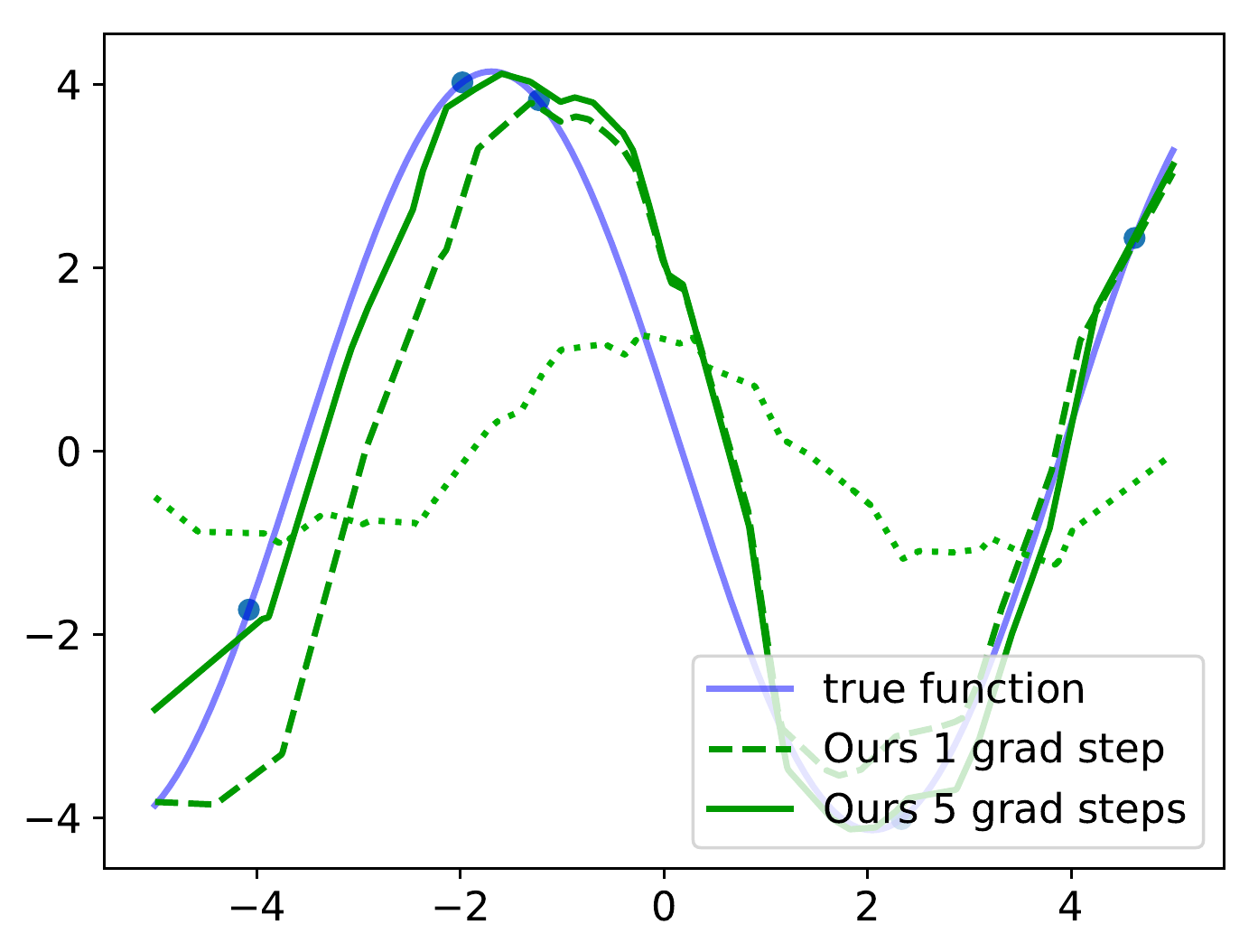}
\label{fig:regression_fig2}
}
\end{center}
\vspace{-1.5em}
\caption{MAML + L2F (Ours) vs MAML on Few-shot regression: (a) Tasks are sampled from the same distribution for training and evaluation. (b) Tasks are sampled from the non-overlapped distributions for training and evaluation. 
In both cases, MAML+L2F (Ours) is more fitted to the true function.}
\vspace{-0.5cm}
\label{fig:regression}
\end{figure*}

We investigate the generalizability of the proposed method across domains, starting with evaluating the performance in $k$-shot regression. In $k$-shot regression, the objective is to fit a function, given $k$ samples of points. Following the general settings from~\cite{finn2017model,Li2017meta}, the target function is set to be a sinusoid with varying amplitude and phase between tasks. The sampling range of amplitude, frequency, and phase defines a task distribution and is set to be the same for both training and evaluation. Regression is visualized in Figure \ref{fig:regression}\subref{fig:regression_fig1}, while its prediction, measured in mean-square error (MSE), is presented in Table \ref{tab:regression}. The results demonstrate that our method not only converges faster but also fits to target functions more accurately.

To further stress the generalization of the MAML+L2F initialization, we extensively increase the degree of conflicts between new tasks and the prior knowledge. To that end, we modify the setting such that amplitude, frequency, and phase are sampled from the non-overlapped ranges for training and evaluation (please refer to the supplementary material for details). In Figure \ref{fig:regression}\subref{fig:regression_fig2}, our model exhibits higher accuracy and thus claims the better generalization.
\begin{figure*}[t] \centering
\begin{center}
\subfloat[2D navigation]{
\includegraphics[width=0.24\linewidth]{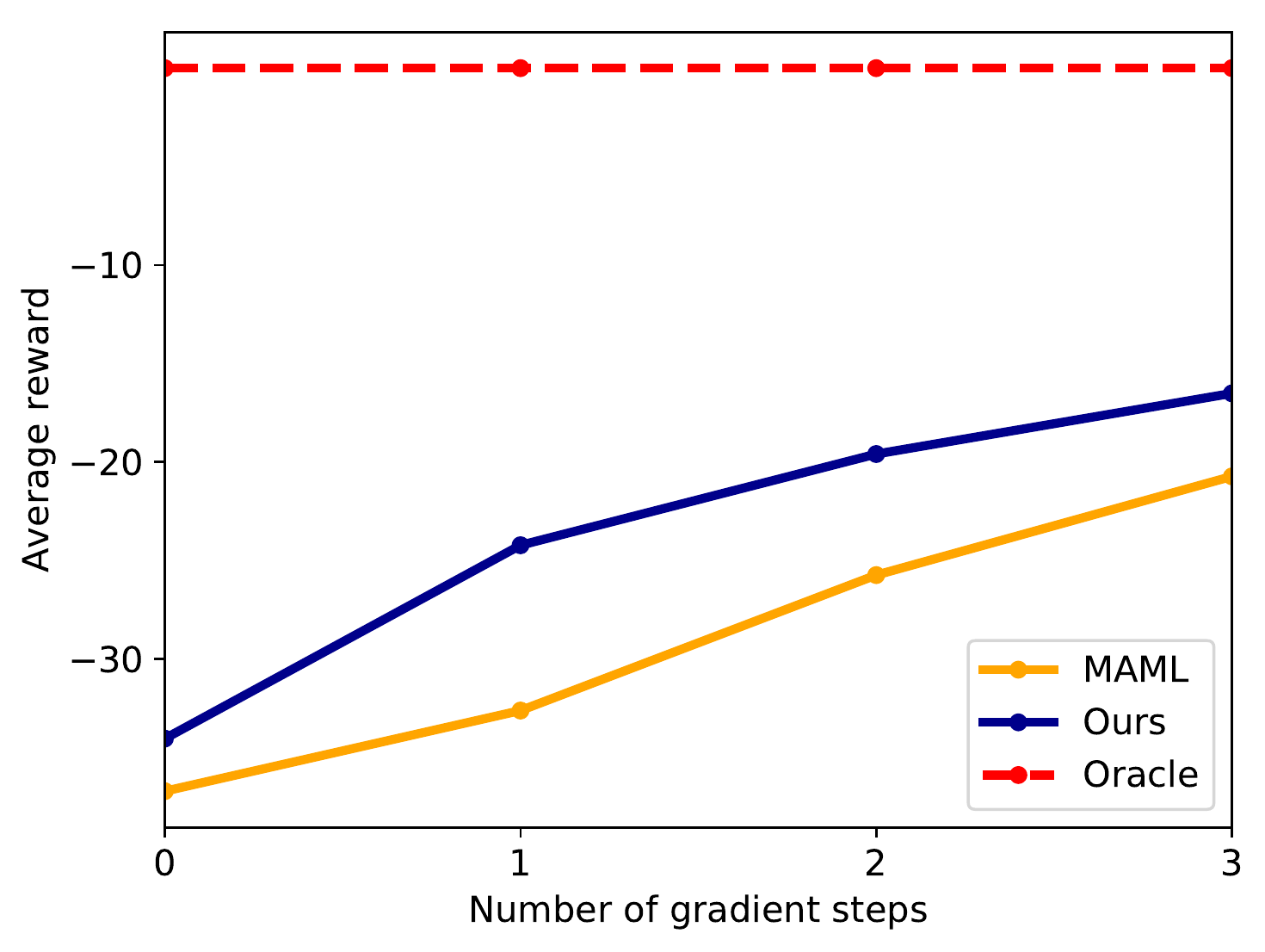}
\includegraphics[width=0.24\linewidth]{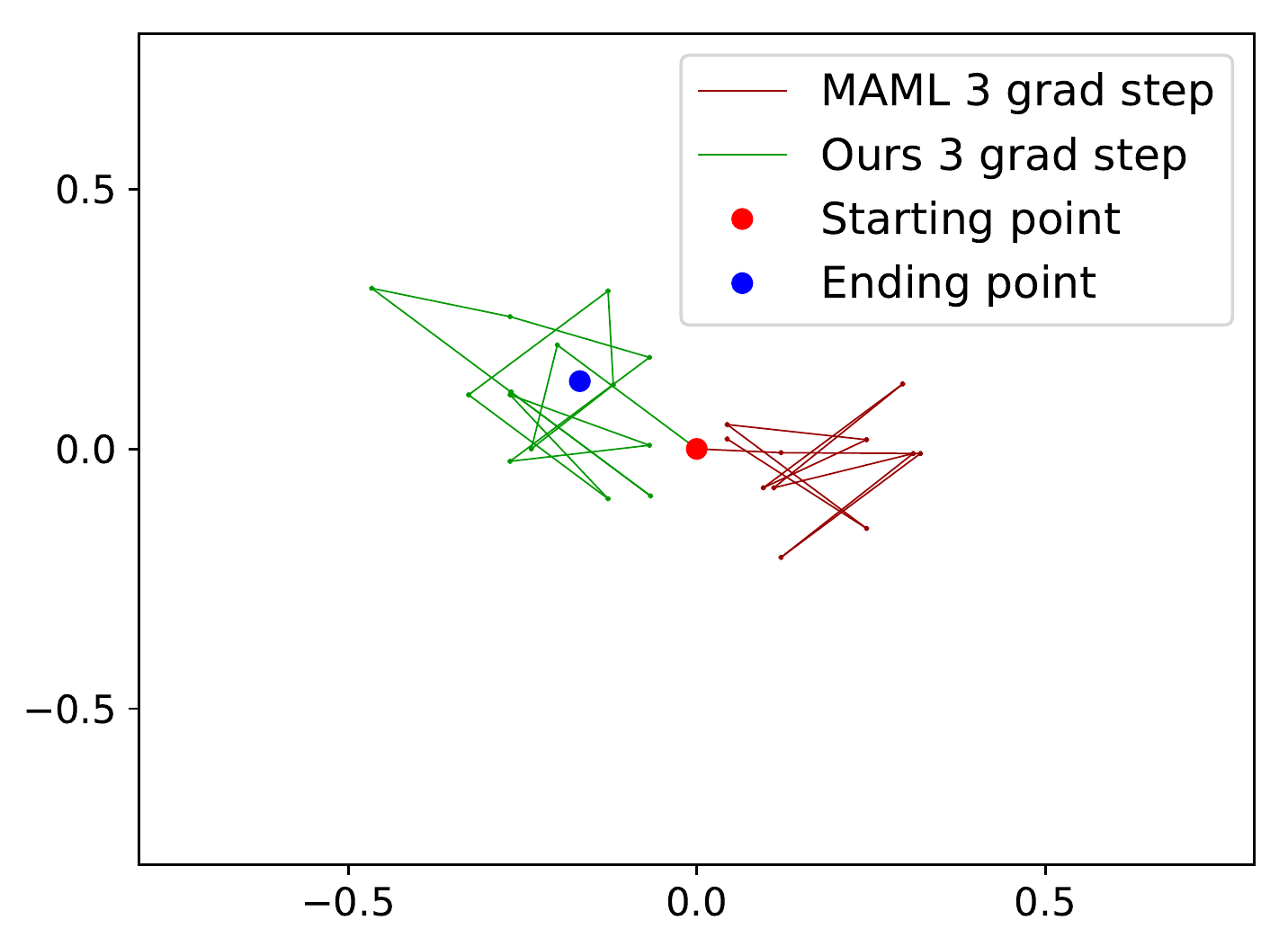}
\label{fig:rl_fig1}
}
\subfloat[Half-cheetah (direction)]{
\includegraphics[width=0.24\textwidth]{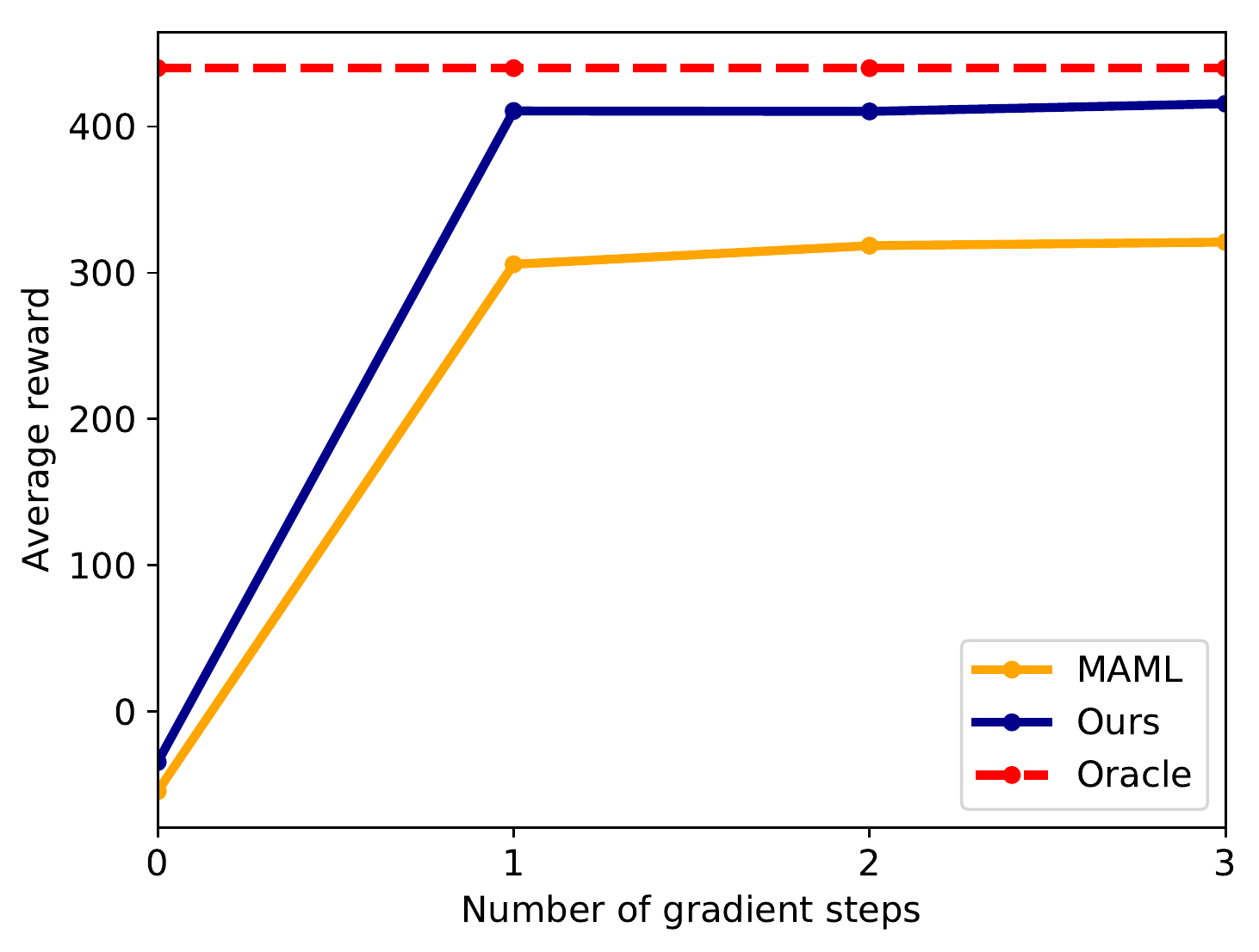}
\label{fig:rl_fig2}
}
\subfloat[Half-cheetah (velocity)]{
\includegraphics[width=0.24\textwidth]{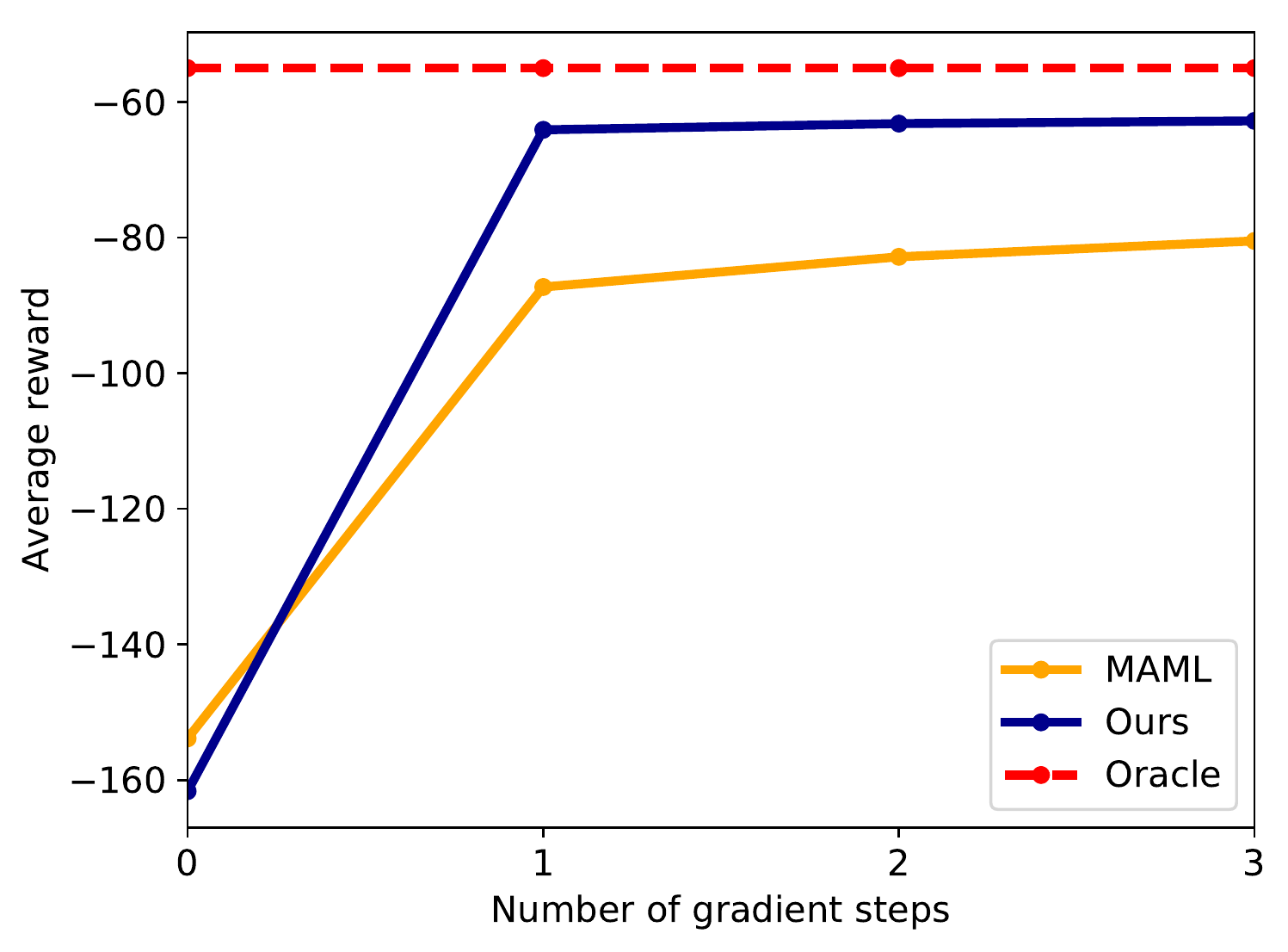}
\label{fig:rl_fig3}
}
\end{center}
\vspace{-1.5em}
\caption{Reinforcement learning results for 3 different environments. The results show that MAML+L2F(Ours) can adapt to each task substantially faster than MAML.}
\vspace{-0.3cm}
\label{fig:rl}
\end{figure*}
\subsection{Reinforcement Learning}
\begin{table}
\small 
\scalebox{0.8}{
\begin{tabular}{@{\extracolsep{5pt}}llrrrr} 
\\[-1.8ex]\hline 
\hline \\[-1.8ex] 
\multicolumn{1}{c}{} & \multicolumn{1}{c}{Models} & \multicolumn{1}{c}{1 step} & \multicolumn{1}{c}{2 steps} & \multicolumn{1}{c}{5 steps}\\
\hline \\[-1.8ex] 
\multirow{ 2 }{*}{ 5-shot training }  &  MAML  &  1.2247  &  1.0268  &  0.8995  \\
  &  MAML+L2F (Ours)  & \textbf{1.0537} & \textbf{0.8426} & \textbf{0.7096} \\
\hline 
\multirow{ 2 }{*}{ 10-shot training }  &  MAML  & 0.9884          & 0.6192          & 0.4072         \\
 &  MAML+L2F (Ours)  & \textbf{0.8069} & \textbf{0.5317} & \textbf{0.3696} \\
\hline
\multirow{ 2 }{*}{ 20-shot training }   & MAML   & 0.6144          & 0.3346          & 0.1817           \\
  &  MAML+L2F (Ours)   & \textbf{0.5475} & \textbf{0.2805} &  \textbf{0.1629} \\
\hline \\[-1.8ex] 
\end{tabular}
}
\caption{MSE averaged over the sampled 100 points with 95$\%$confidence intervals on $k$-shot regression. Our method consistently outperforms across all gradient steps.}
\label{tab:regression}
\vspace{-0.3cm}
\end{table}

To further validate the generalizability of L2F, we evaluate the performance in reinforcement learning, specifically in 2D navigation and locomotion environments from~\cite{duan2016benchmark} as in~\cite{finn2017model}. We briefly outline the task description below (please refer to the supplementary material for details). Figure \ref{fig:rl} presents consistent improvement over MAML across different experiments. This solidifies the generalizability and effectiveness of our proposed method.
\vspace{-0.5cm}
\subsubsection{2D Navigation}
\vspace{-0.1cm}
A 2D navigation task is to move an agent from the starting point to the destination point in 2D space, where the reward is defined as the negative of the squared distance to the destination point. We follow the experiment procedure from~\cite{finn2017model}, where they fix the starting point and only vary the location of destination between tasks. 

Figure \ref{fig:rl}\subref{fig:rl_fig1} presents faster and more precise navigation by our model in both experiment settings, both quantitatively and qualitatively. This solidifies the severity of the conflicts between tasks.
\vspace{-0.5cm}
\subsubsection{Mujoco}
\vspace{-0.3cm}
As a more complex reinforcement-learning environment, we experiment on locomotion with the MuJoCo simulator~\cite{todorov2012mujoco}, where there are two sets of tasks: a robot is required to move in a particular direction in one set and move with a particular velocity in the other. For both experiments,our method outperforms MAML in large margins as shown in Figure \ref{fig:rl}\subref{fig:rl_fig2}, \subref{fig:rl_fig3}.
\subsection{Loss Landscape}
We further validate the effectiveness of our model by illustrating the smoother loss landscape after applying L2F to MAML for the miniImageNet classification tasks, as shown in Figure \ref{fig:landscape}. At the initial stages of training, L2F appears to struggle more, while optimization of MAML seems more stable. 
This may seem contradictory at first but this actually validates our argument about conflicts between tasks even further. 
At the beginning, the MAML initialization is not trained enough and thus does not have sufficient prior knowledge of task distribution yet. 
As training proceeds, the initialization encodes more information about task distribution and encounters conflicts between tasks more frequently. 
As for L2F, the attenuator $g_\phi$ initially does not have enough knowledge about the task distribution and thus generates meaningless attenuation $\gamma_i$, deteriorating the initialization. 
But, the attenuator increasingly encodes more information about the task distribution, generating more appropriate attenuation $\gamma_i$ that corresponds to tasks well. 
The generated $\gamma_i$ accordingly allows for a learner to \textit{forget} the irrelevant part of prior knowledge to help fast adaptation, as illustrated by increasing stability and smoothness of landscape.
\vspace{-3mm}
\section{Conclusion}
In this paper, we argue that forcibly sharing a common initialization in MAML induces conflicts across tasks and thus results in the compromised location of the initialization. The severely sharp loss landscape asserts that such compromise makes the MAML initialization a ``bad'' starting position for fast adaptation. We propose to resolve this discrepancy by facilitating \textit{forgetting} (attenuating) the irrelevant information that may hinder fast adaptation. Specifically, we propose a task-dependent layer-wise attenuation, named L2F, motivated by the observation that the degree of compromise varies between network layers and tasks. Through extensive experiments across different domains, we validate our argument that selective \textit{forgetting} greatly facilitates fast adaptation while retaining the simplicity and generalizability of MAML.




\setcounter{section}{0}
\renewcommand{\thesection}{\Alph{section}}
\renewcommand{\thefigure}{\Alph{figure}}
\renewcommand{\thetable}{\Alph{table}}

\section{Loss Landscape}
\begin{figure*}
\begin{center}
\subfloat[loss landscape]{
    \includegraphics[trim=15 0 20 20, clip,height=0.15\textheight,width=0.3\textwidth]{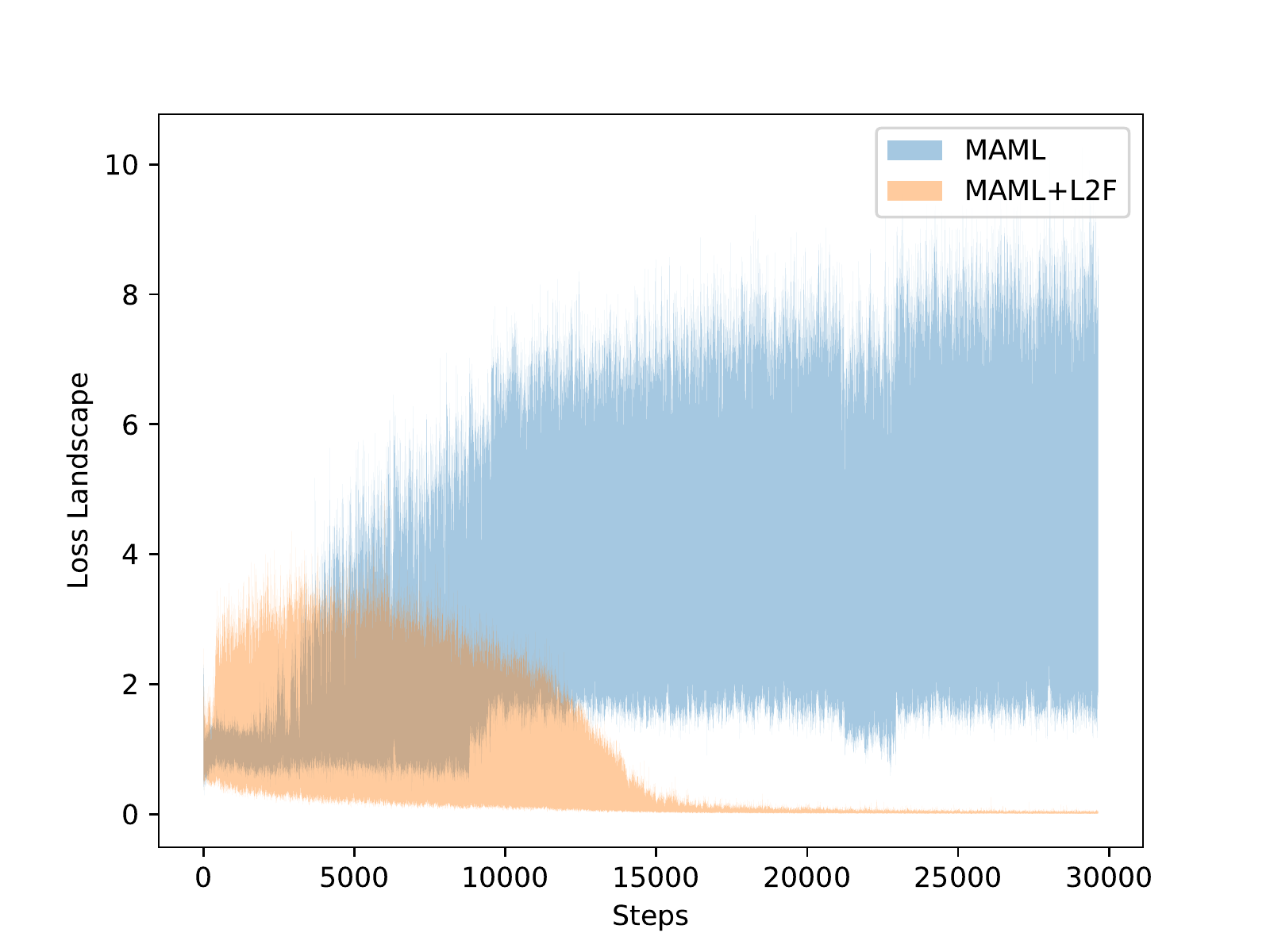}
    \label{sup_fig:loss}
}
\subfloat[gradient predictiveness]{
    \includegraphics[trim=0 20 20 20, clip,height=0.15\textheight,width=0.3\textwidth]{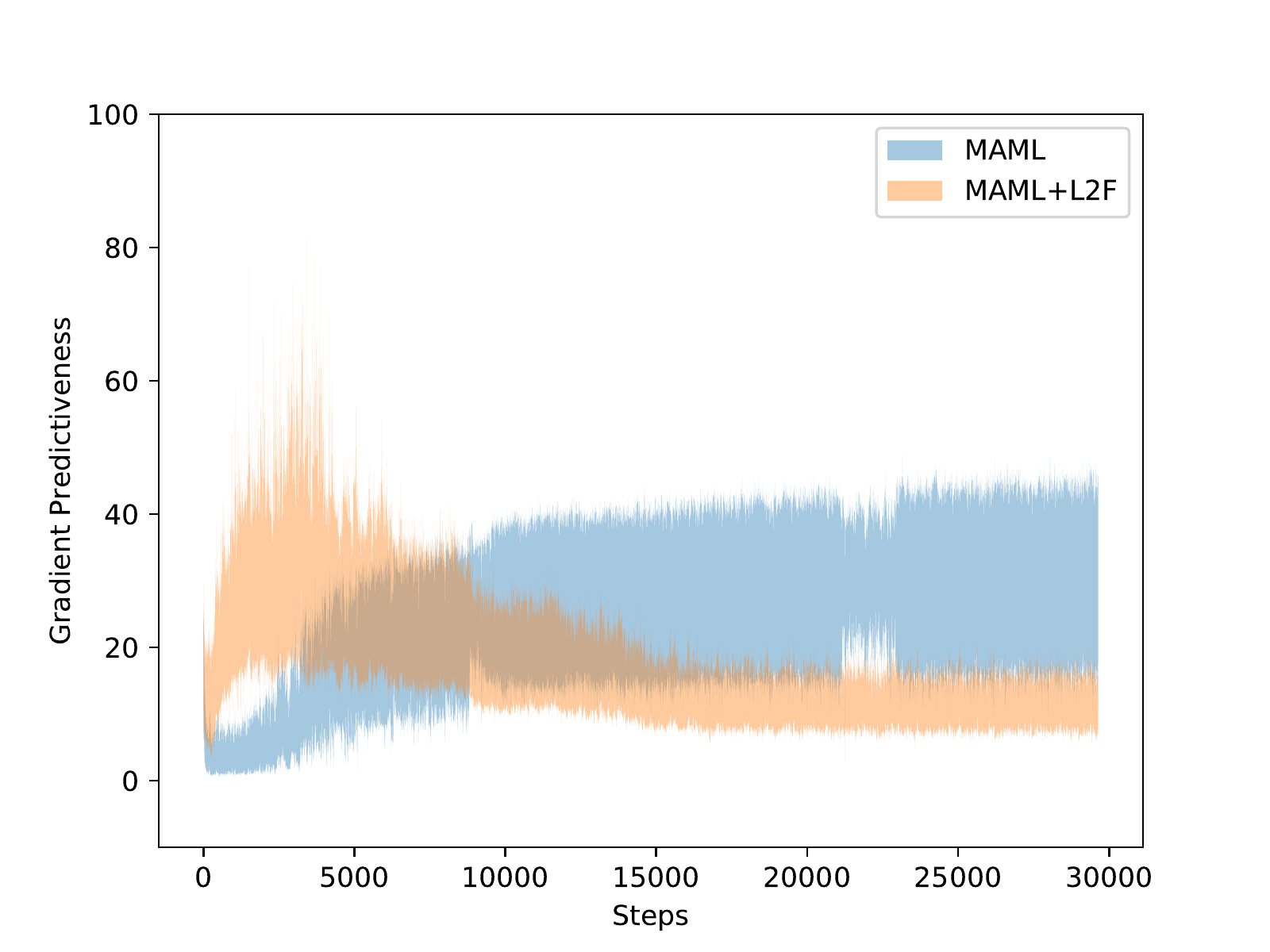}
    \label{sup_fig:gradient}
}
\subfloat[``effective'' $\beta$ smoothness]{
    \includegraphics[trim=0 20 20 20, clip,height=0.15\textheight,width=0.3\textwidth]{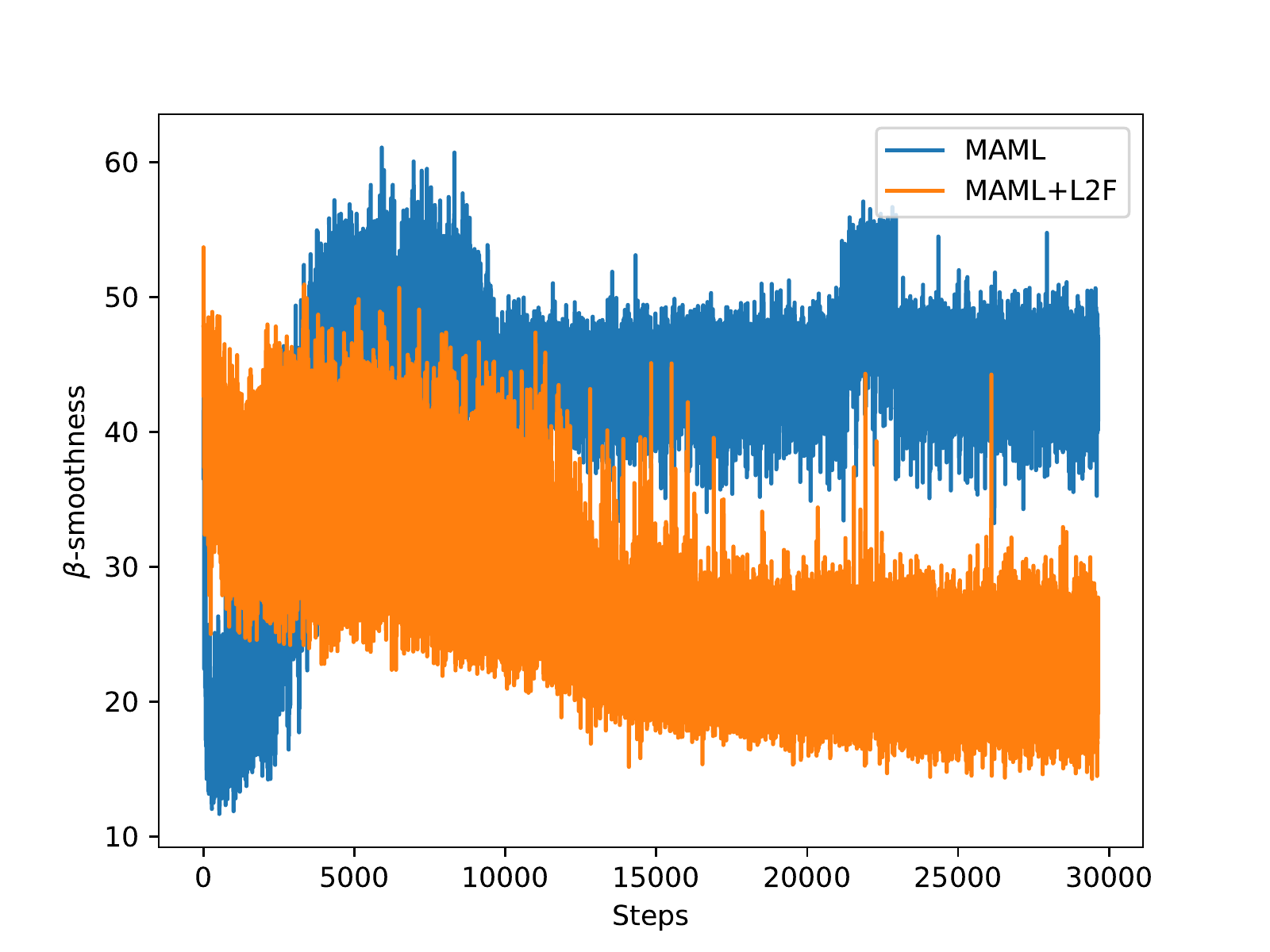}
    \label{sup_fig:beta}
}
\end{center}
\vspace{-2em}
\caption{Analysis of the optimization landscape of the fast adaptation to tasks from the meta-training set. In each subfloat, averaged values are shown for each training iteration.}
\vspace{-1em}
\label{fig:train}
\end{figure*}

\begin{figure*}
\begin{center}
\subfloat[loss landscape]{
    \includegraphics[trim=15 0 20 20, clip,height=0.15\textheight,width=0.3\textwidth]{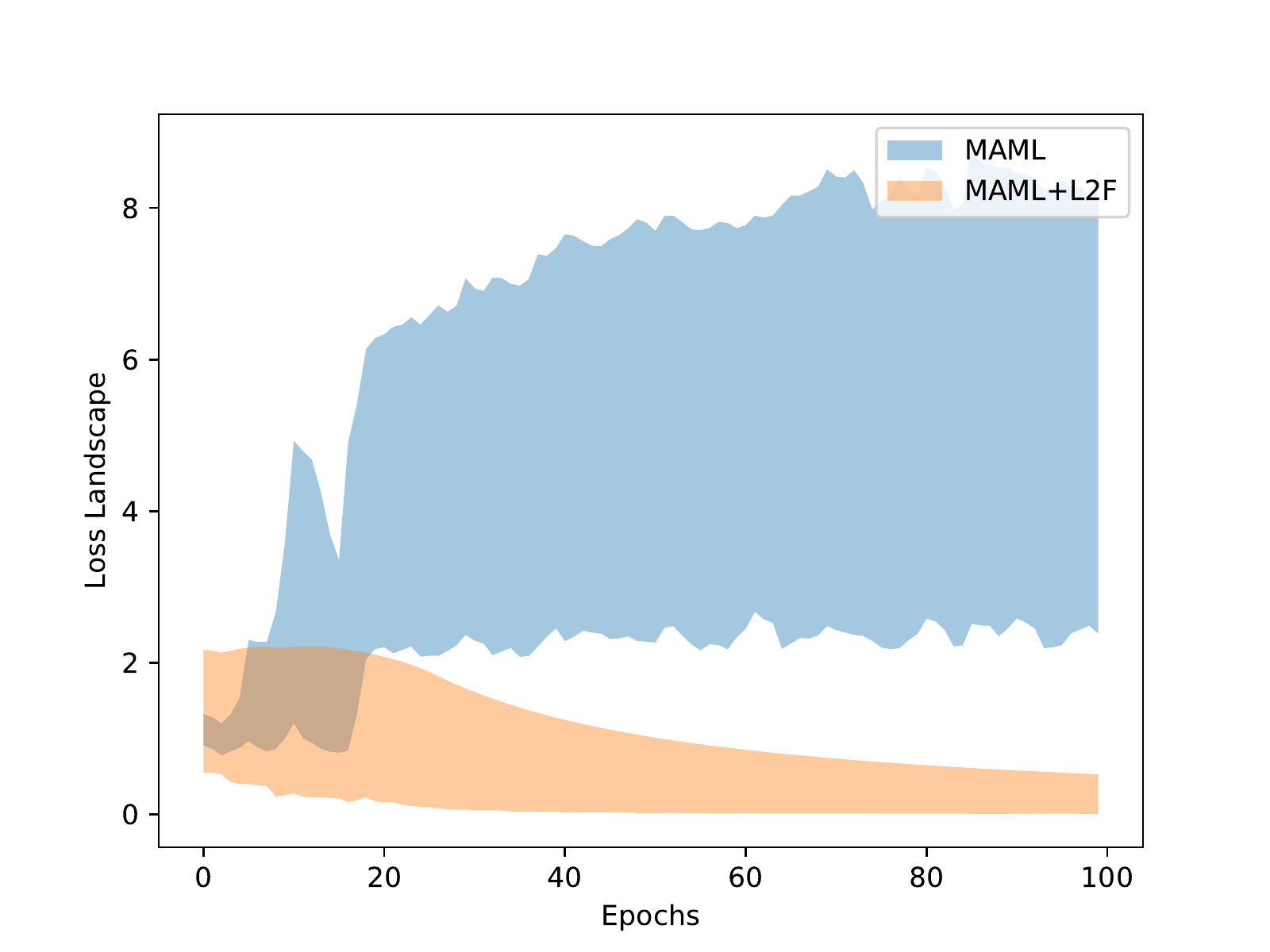}
    \label{fig:val_loss}
}
\subfloat[gradient predictiveness]{
    \includegraphics[trim=20 0 20 20, clip,height=0.15\textheight,width=0.3\textwidth]{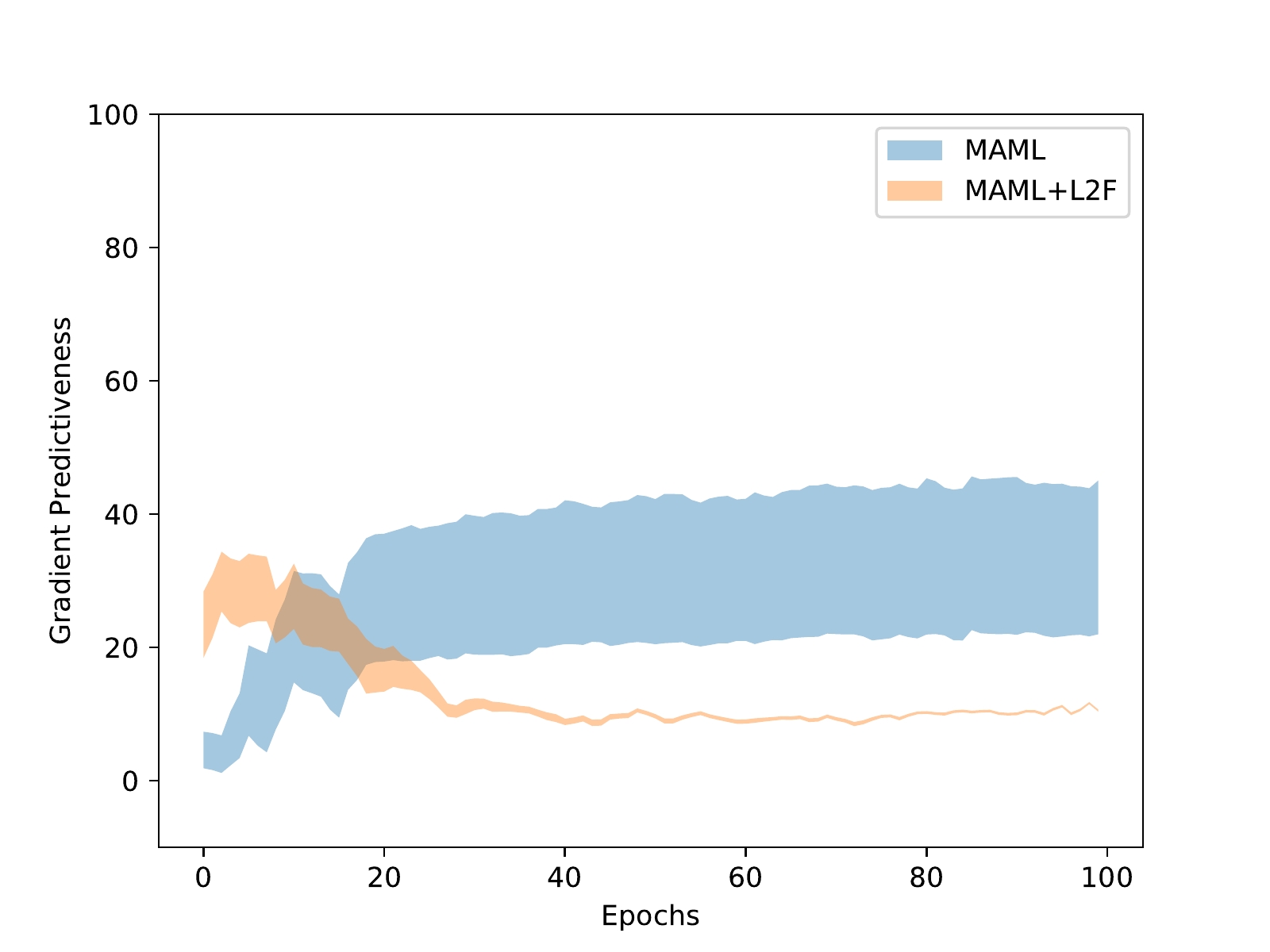}
    \label{fig:val_gradient}
}
\subfloat[``effective'' $\beta$ smoothness]{
    \includegraphics[trim=20 0 20 20, clip,height=0.15\textheight,width=0.3\textwidth]{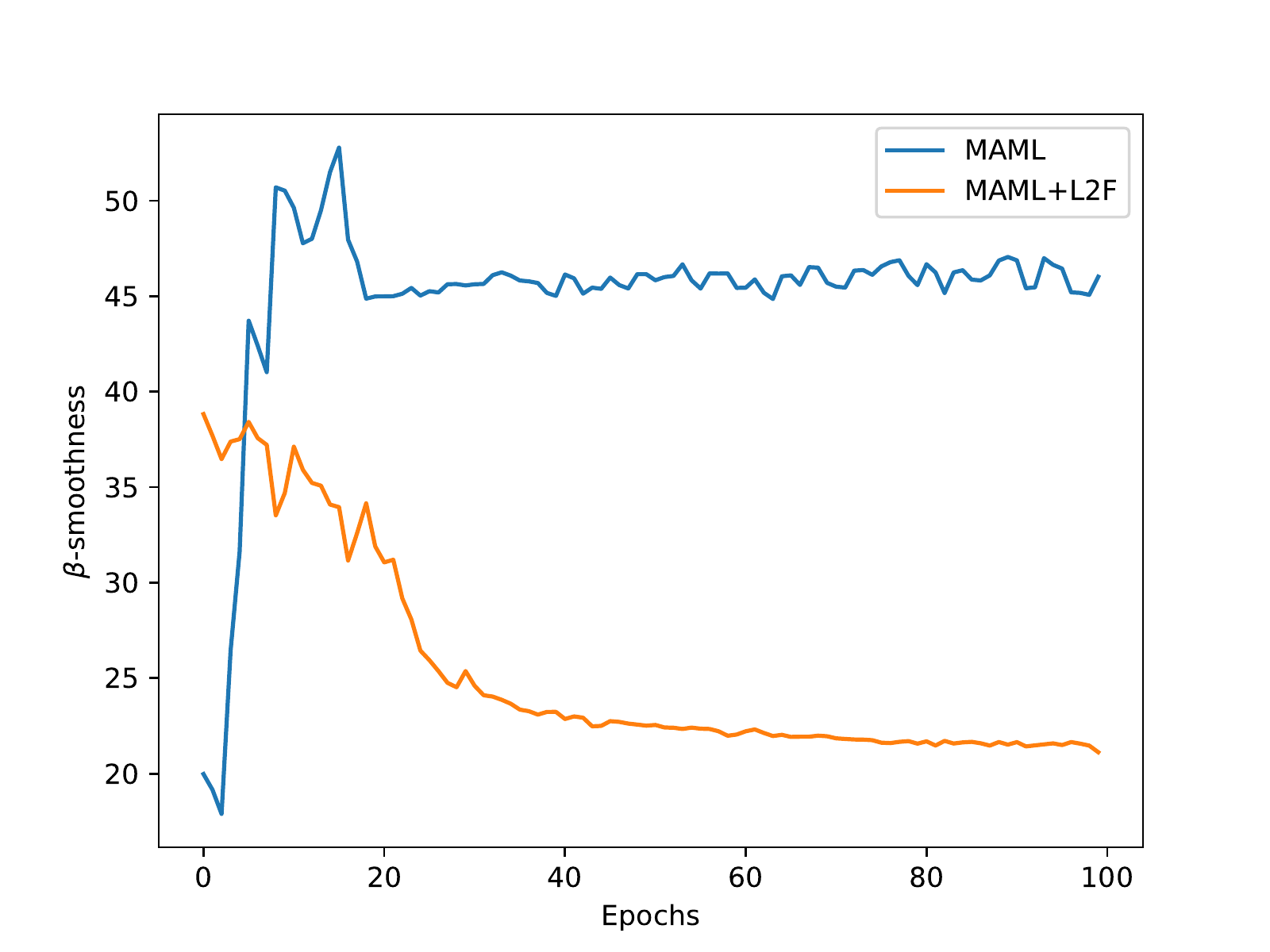}
    \label{fig:val_beta}
}
\end{center}
\vspace{-1em}
\caption{Analysis of the optimization landscape of the fast adaptation to tasks from the meta-validation set. In each subfloat, averaged values are shown for each training epoch.}
\vspace{-1em}
\label{fig:val}
\end{figure*}

In \cite{santurkar2018how}, they analyze the stability and smoothness of the optimization landscape by measuring Lipschitzness and the ``effective'' $\beta$-smoothness of loss. We use these measurements to analyze learning dynamics for both MAML and our proposed method during training on 5-way 5-shot miniImageNet classification tasks. In Figure~\ref{fig:train}, we start with investigating fast-adaptation (or inner-loop) optimization. At each inner-loop update step, we measure variations in loss (Figure~\ref{fig:train}\subref{sup_fig:loss}), the $l_2$ difference in gradients (Figure~\ref{fig:train}\subref{sup_fig:gradient}), and the maximum difference in gradient over the distance (Figure~\ref{fig:train}\subref{sup_fig:beta}), as we move to different points along the computed gradient for that gradient descent. We take an average of these values over the number of inner-loop updates and plot them against training iterations. With a similar approach, we also analyze the optimization stability of fast adaptation to validation tasks at every epoch (Figure~\ref{fig:val}). The measurements were averaged over (the number of validation tasks $\times$ the number of inner-loop update steps).

At the initial stages of training, L2F appears to struggle more, while optimization of MAML seems more stable. 
This may seem contradictory at first but this actually validates our argument about conflicts between tasks even further. 
At the beginning, the MAML initialization is not trained enough and thus does not have sufficient prior knowledge of task distribution yet. 
As training proceeds, the initialization encodes more information about task distribution and encounters conflicts between tasks more frequently.

As for L2F, the attenuator network $g_\phi$ initially does not have enough knowledge about the task distribution and thus generates meaningless attenuation $\gamma$, deteriorating the initialization. But, the attenuator network increasingly encodes more information about the task distribution, generating more appropriate attenuation $\gamma$ that corresponds to tasks well. The generated $\gamma$ accordingly allows for a learner to \textit{forget} the irrelevant part of prior knowledge to help fast adaptation, as illustrated by increasing stability and smoothness of landscape in Figure \ref{fig:train}. 
The similar observation can be made from \ref{fig:val}, illustrating the  generalizability and the robustness of the proposed method to unseen tasks.

\begin{figure*}
\begin{center}
\subfloat[loss landscape]{
    \includegraphics[trim=15 0 20 20, clip,height=0.15\textheight,width=0.3\textwidth]{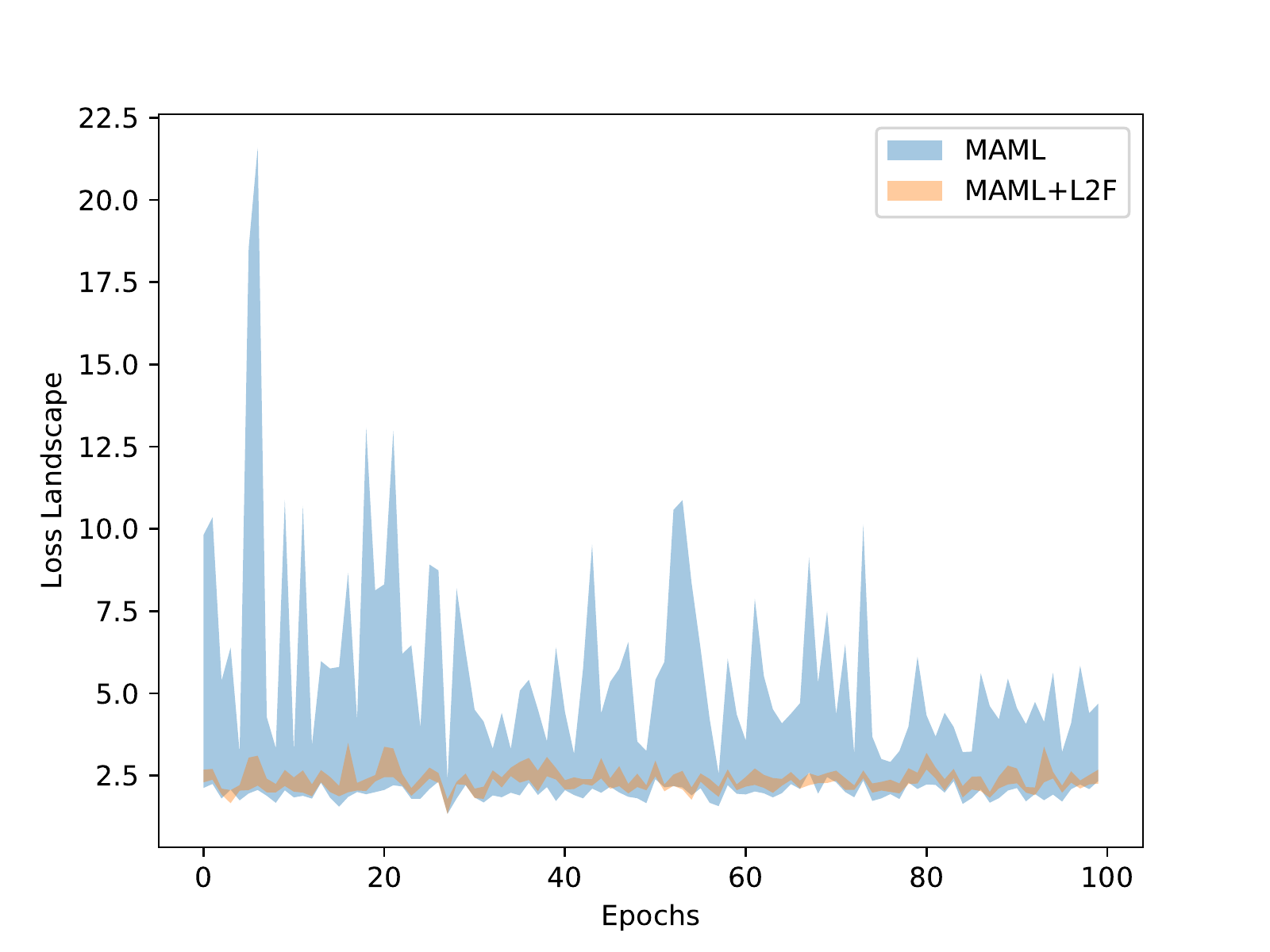}
    \label{sup_fig:loss_2}
}
\subfloat[gradient predictiveness]{
    \includegraphics[trim=20 0 20 20, clip,height=0.15\textheight,width=0.3\textwidth]{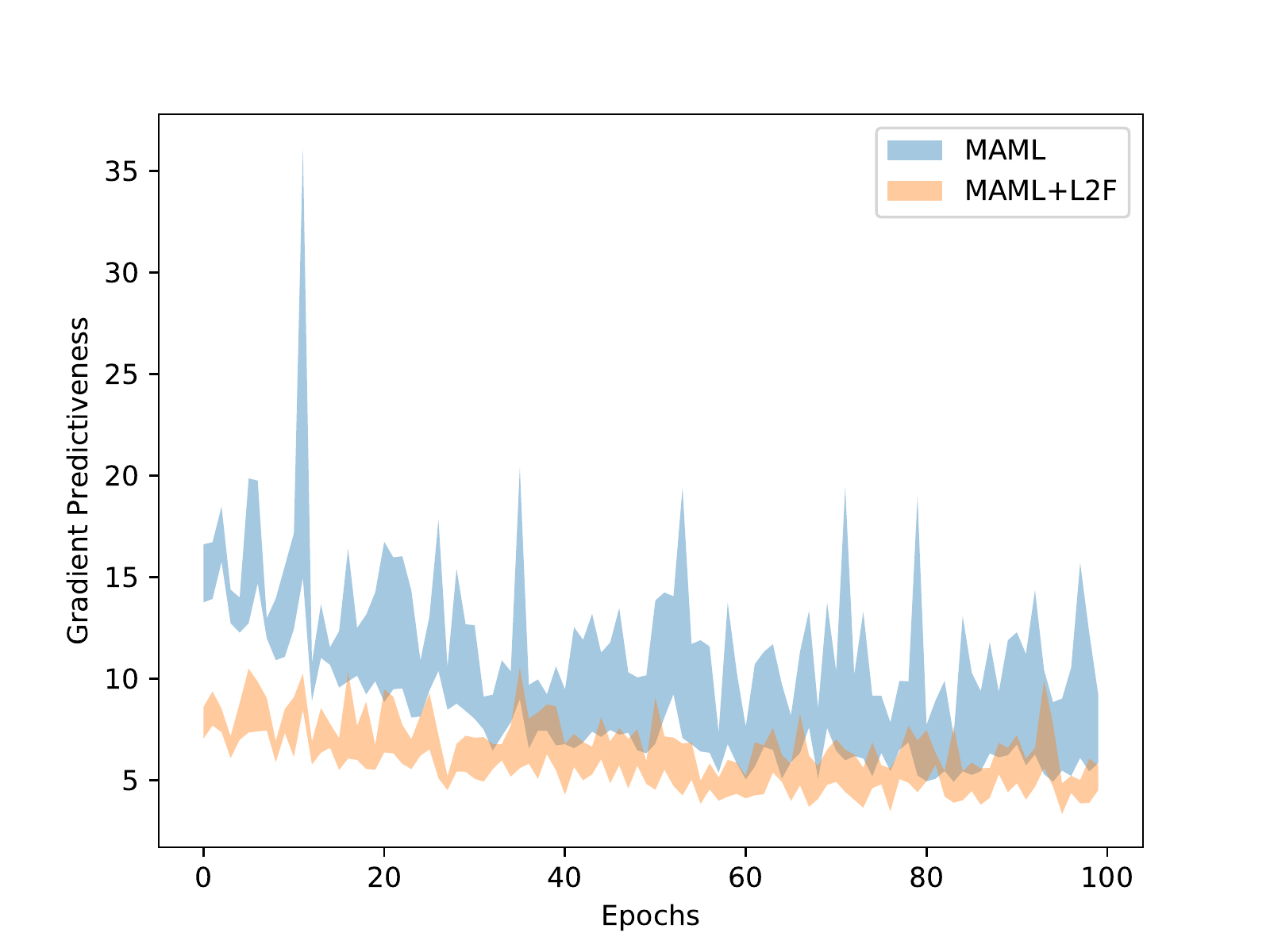}
    \label{sup_fig:gradient_2}
}
\subfloat[``effective'' $\beta$ smoothness]{
    \includegraphics[trim=20 0 20 20, clip,height=0.15\textheight,width=0.3\textwidth]{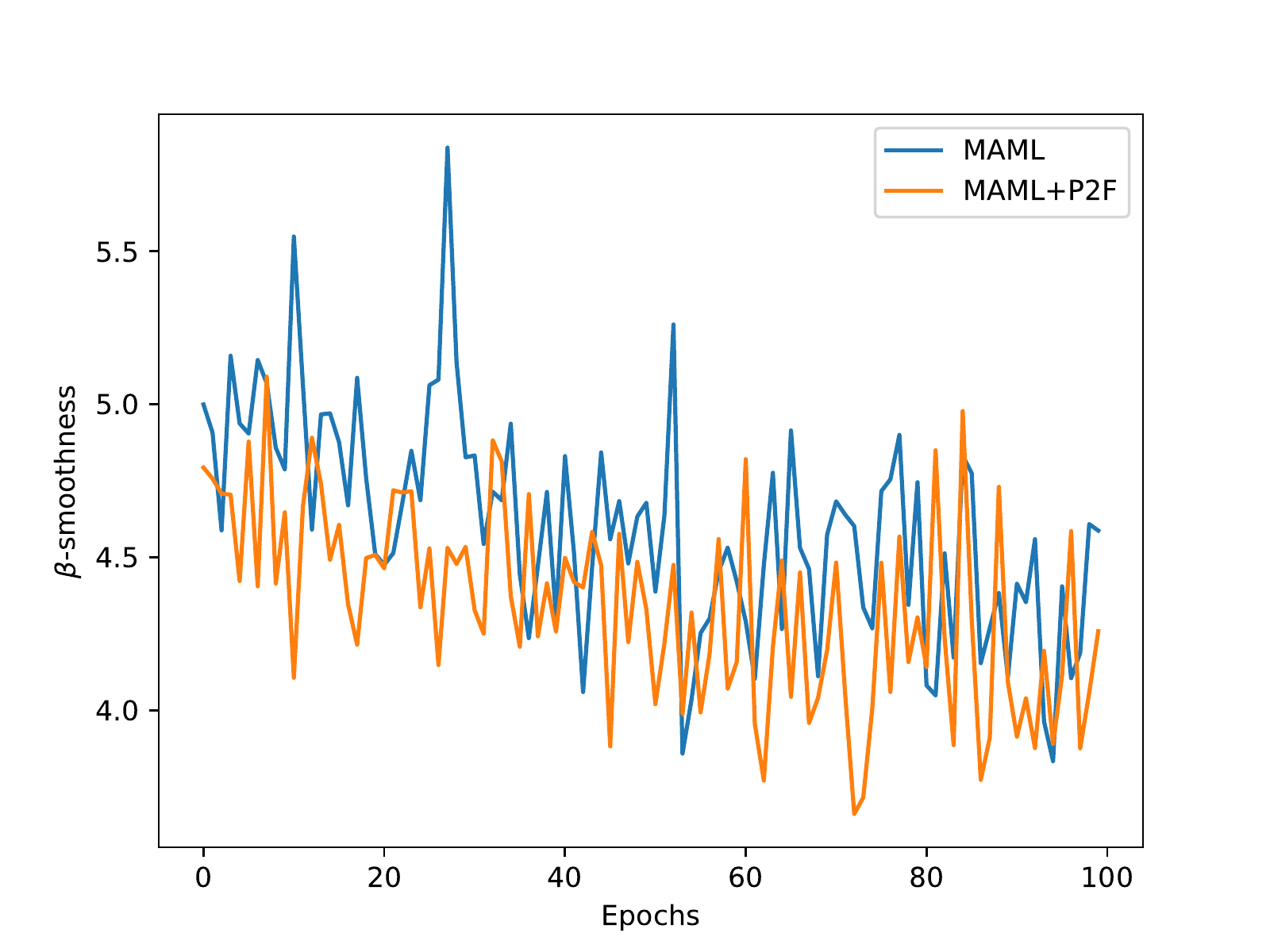}
    \label{sup_fig:beta_2}
}
\end{center}
\vspace{-1em}
\caption{Analysis of the optimization landscape of the initialization learning dynamics.}
\label{fig:init}
\end{figure*}

We also investigate the optimization landscape of learning the initialization $\theta$ itself for both MAML and L2F in Figure \ref{fig:init}. The figure demonstrates that the more stable and smoother landscape is realized by L2F. Because the task-dependent layer-wise attenuation allows for \textit{forgetting} the irrelevant or conflicting part of prior knowledge present in the initialization $\theta$, it lifts a burden of trying to resolve conflicts between tasks from $\theta$, allowing for more stable training of the initialization itself.

\section{Extended Experiments on Classification}
\begin{table}[b]
   \small 
   \centering 
   \scalebox{0.85}{
   \begin{threeparttable}
   \begin{tabular}{lcccccc} 
   \toprule[\heavyrulewidth]\toprule[\heavyrulewidth]
          & \multicolumn{3}{c}{\textbf{FC100}} \\ 
          &1-shot&5-shot&10-shot\\
   \midrule
   MAML\tnote{*} & $35.98 \pm 0.48\%$ & $51.40 \pm 0.50\%$ & $56.13 \pm 0.50\%$\\
   MAML+L2F& $39.46 \pm 0.49\%$ & $53.12 \pm 0.50\%$ & $59.72 \pm 0.49\%$\\
   \midrule
   \bottomrule[\heavyrulewidth] 
   \end{tabular}
   \begin{tablenotes}
   \item[*] Our reproduction.
   \end{tablenotes}
   \end{threeparttable}
   }
   \caption{Test accuracy on FC100 5-way classification} 
   \label{tab:result_fc100}
\end{table}
\begin{table}[b]
   \small 
   \centering 
   \begin{threeparttable}
   \begin{tabular}{lccccc} 
   \toprule[\heavyrulewidth]\toprule[\heavyrulewidth]
          & \multicolumn{3}{c}{\textbf{CIFAR-FS}} \\ 
          &1-shot&5-shot\\
   \midrule
   MAML\tnote{*}  & $53.91 \pm 0.50\%$ & $70.16 \pm 0.46\%$\\
   MAML+L2F  & $57.28 \pm 0.49\%$ & $73.94 \pm 0.44\%$\\
   \bottomrule[\heavyrulewidth] 
   \end{tabular}
   \begin{tablenotes}
   \item[*] Our reproduction.
   \end{tablenotes}
   \end{threeparttable}
   \caption{Test accuracy on CIFAR-FS 5-way classification} 
   \label{tab:result_cifar100}
\end{table}
To further validate that our method consistently provides benefits regardless of scenarios, we compare our method against the baseline on additional datasets that have been recently introduced: FC100 (Fewshot-CIFAR100) \cite{oreshkin2018tadam} and CIFAR-FS (CIFAR100 few-shots) \cite{bertinetto2019meta}. Both aim for creating more challenging scenarios by using low resolution images ($32\times$32, compared to $84\times84$ in miniImageNet \cite{ravi2017optimization} and tieredImageNet \cite{ren2018meta}) from CIFAR100 \cite{krizhevsky2009learning}. These two datasets differ in how they create the train/val/test splits of CIFAR100. While CIFAR-FS follows the procedure that was used for miniImageNet, FC100 aligns more with the goal of tieredImageNet in that they try to minimize the amount overlap between splits by splitting based on superclasses. Table \ref{tab:result_fc100} presents results for FC100 and Table \ref{tab:result_cifar100} for CIFAR-FS. 
We also perform additional experiments on Meta-Dataset~\cite{triantafillou2018meta}, which is a combination of diverse datasets and hence poses more challenging scenarios, where conflicts can occur among tasks more frequently. Table \ref{tab:meta_dataset} shows that our proposed method resolves conflicts better than MAML even under challenging scenarios.

\begin{table*}
   \centering 
   \scalebox{0.7}{
   \begin{tabular}{lccccccccccc} 
   \toprule[\heavyrulewidth]
          Model&ILSVRC&Omniglot&Aircraft&Birds&Textures&Quick Draw&Fungi&VGG Flower&Traffic signs&MSCOCO\\
   \midrule
   MAML&$19.35\pm0.84$&$66.14\pm1.47$&$40.20\pm1.05$&$40.61\pm1.79$&$38.94\pm1.69$&$42.46\pm1.54$&$13.80\pm1.19$&$61.07\pm1.50$&$23.38\pm1.12$&$13.29\pm1.11$\\
   Ours&$25.93\pm1.10$&$72.26\pm1.63$&$53.31\pm1.48$&$42.62\pm1.30$&$49.57\pm1.03$&$50.28\pm1.67$&$20.20\pm1.08$&$64.23\pm1.31$&$31.71\pm1.45$&$19.75\pm0.93$\\
   \bottomrule[\heavyrulewidth] 
   \end{tabular}
   }
   \caption{Test accuracy (\%) of MAML (our reproduction) vs MAML+L2F on Meta-Dataset} 
   \label{tab:meta_dataset}
\end{table*}

\section{Regression}
\subsection{Additional Qualitative results}
\begin{table*}
\centering
\subfloat[Regression]{\label{tab:reg}\scalebox{0.8}{
\begin{tabular}{@{\extracolsep{5pt}} llrrr} 
\vspace{-0.3cm}
\\[-1.8ex]\hline 
\hline \\[-1.8ex] 
\multicolumn{1}{c}{} & \multicolumn{1}{c}{Models} & \multicolumn{1}{c}{1 step} & \multicolumn{1}{c}{2 steps} & \multicolumn{1}{c}{5 steps}\\
\hline \\[-1.8ex] 
\multirow{ 4 }{*}{ 5-shot }  &  MAML  &  1.2247  &  1.0268  &  0.8995  \\
  &  MuMoMAML  & 1.1010 & 0.9291 & 0.8615 \\
  &  MAML++  & 1.2028 & 0.9268 & 0.7547 \\
  &  Ours  & \textbf{1.0537} & \textbf{0.8426} & \textbf{0.7096} \\
\hline  \\[-1.8ex] 
\end{tabular}
}}
\quad
\subfloat[RL]{\label{tab:rll}\scalebox{.8}{
\begin{tabular}{@{\extracolsep{5pt}} llrrrr} 
\vspace{-0.3cm}
\\[-1.8ex]\hline 
\hline \\[-1.8ex] 
\multicolumn{1}{c}{} & \multicolumn{1}{c}{Models} & \multicolumn{1}{c}{1 step} & \multicolumn{1}{c}{2 steps} & \multicolumn{1}{c}{3 steps}\\
\hline \\[-1.8ex] 
\multirow{ 4 }{*}{ 2D Navi }  &  MAML  &  -32.626  &  -25.746  &  -20.734 \\
  &  MuMoMAML  & -25.785 & -23.705 & -19.747 \\
  &  MAML++  & -36.281 & -27.264 & -18.620 \\
  &  Ours  & \textbf{-24.230} & \textbf{-19.598} & \textbf{-16.517} \\
\hline  \\[-1.8ex]
\end{tabular}
}}
\caption{Additional Quantitative Experiments}
\label{tab:additional}
\vspace{-0.5cm}
\end{table*}

\begin{figure*}
\begin{minipage}[b]{.5\textwidth}
\centering
\includegraphics[width=0.48\linewidth]{supple_figures/regression_5/39_test_both_plot.pdf}
\includegraphics[width=0.48\linewidth]{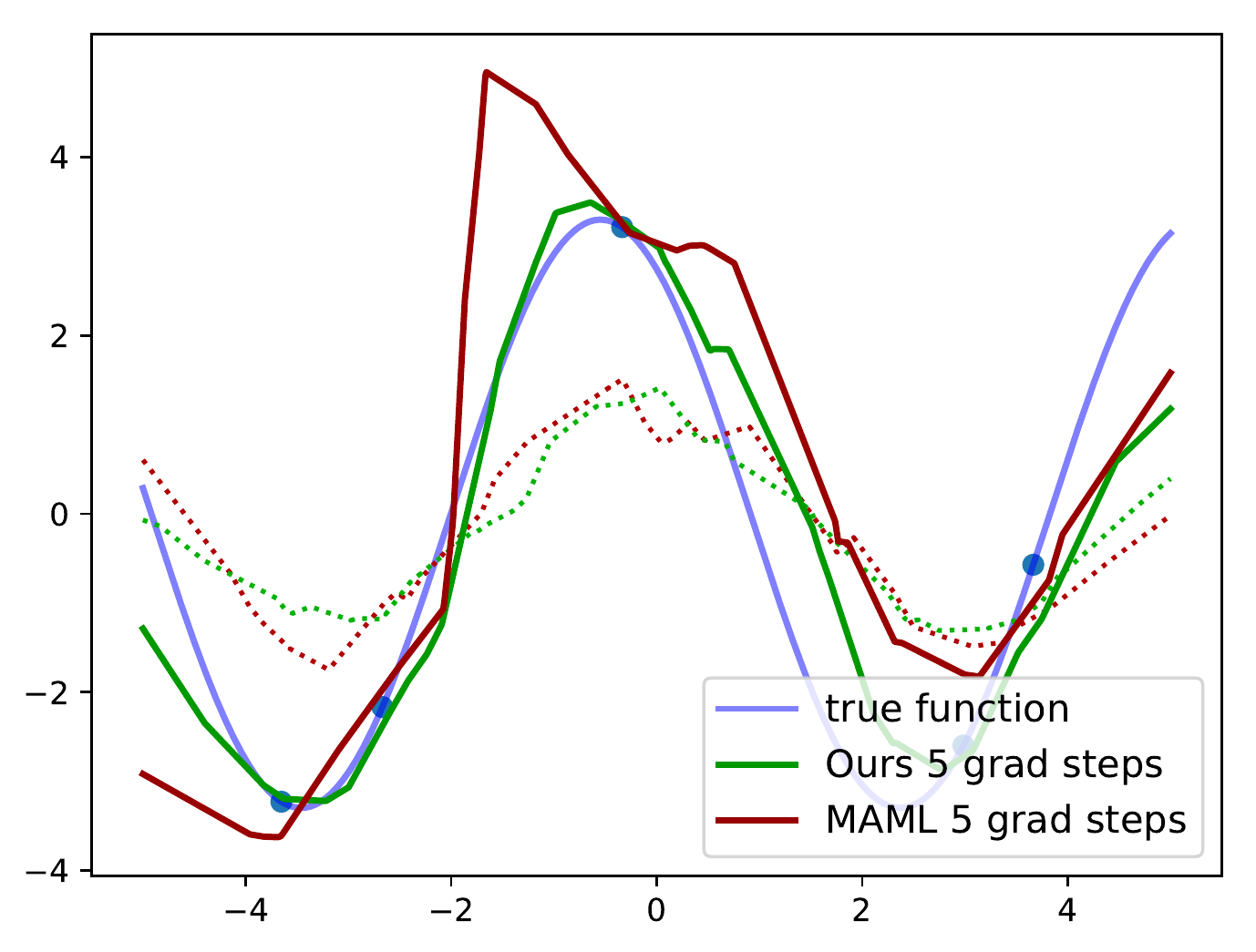}
\vfill
\includegraphics[width=0.48\linewidth]{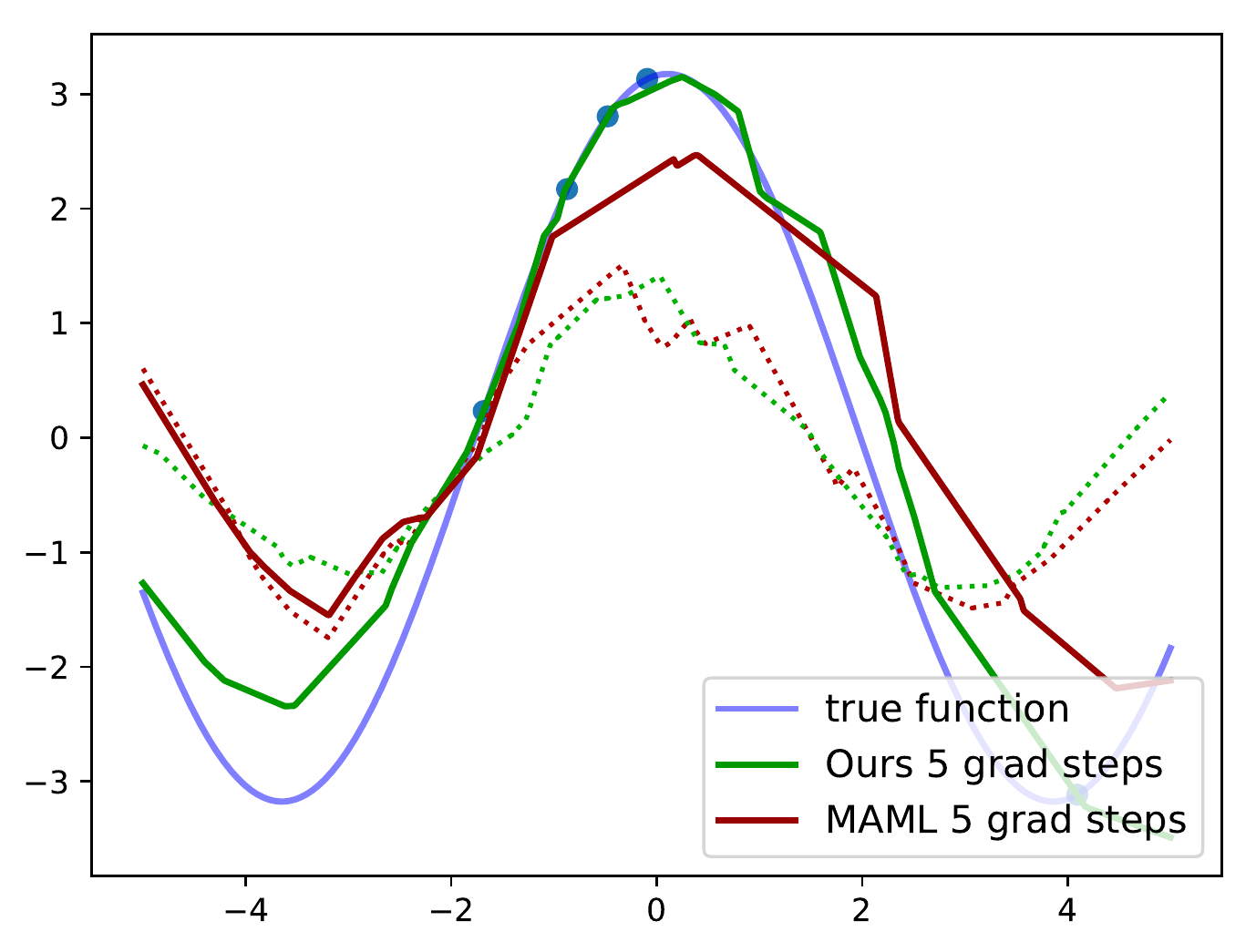}
\includegraphics[width=0.48\linewidth]{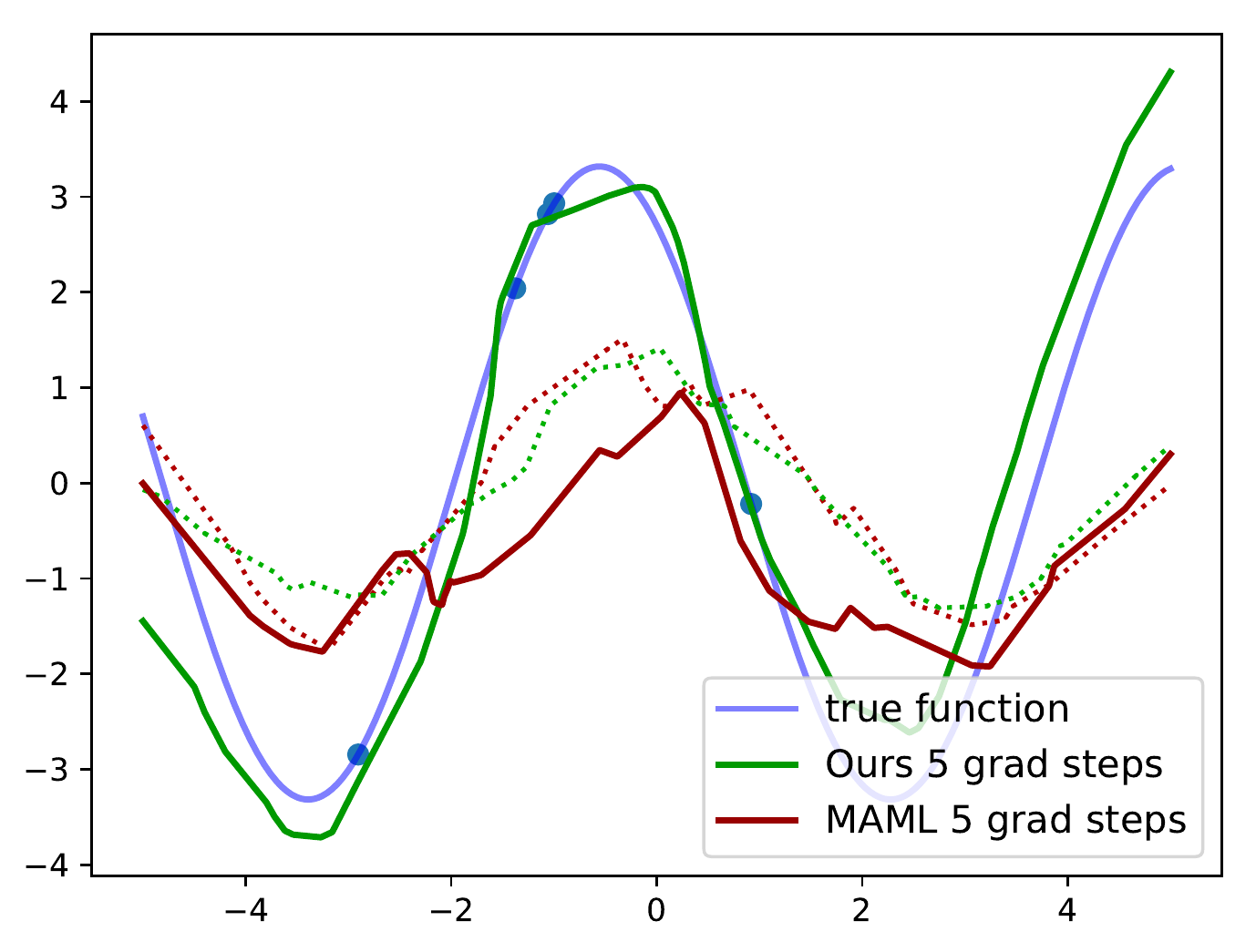}
\vfill
\includegraphics[width=0.48\linewidth]{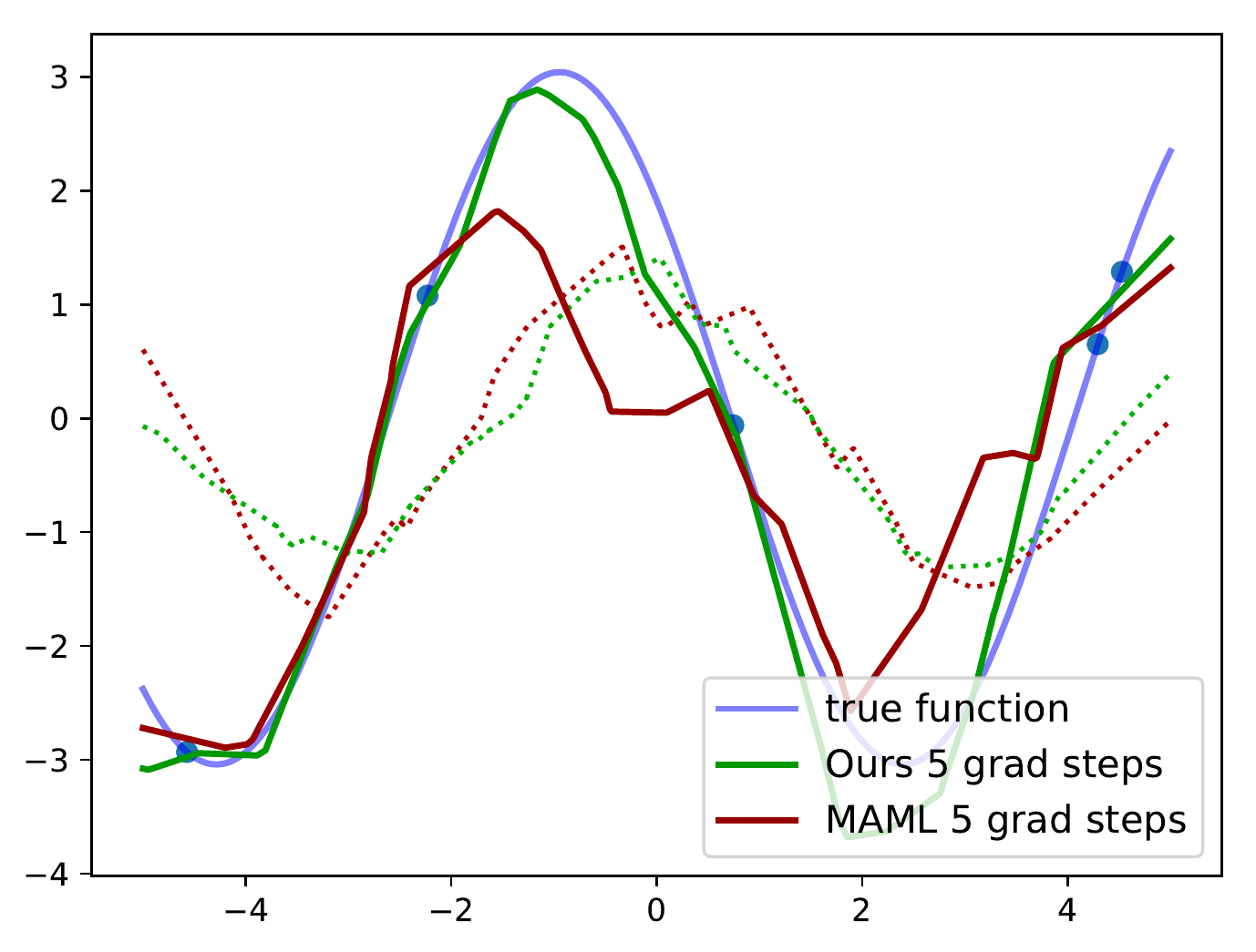}
\includegraphics[width=0.48\linewidth]{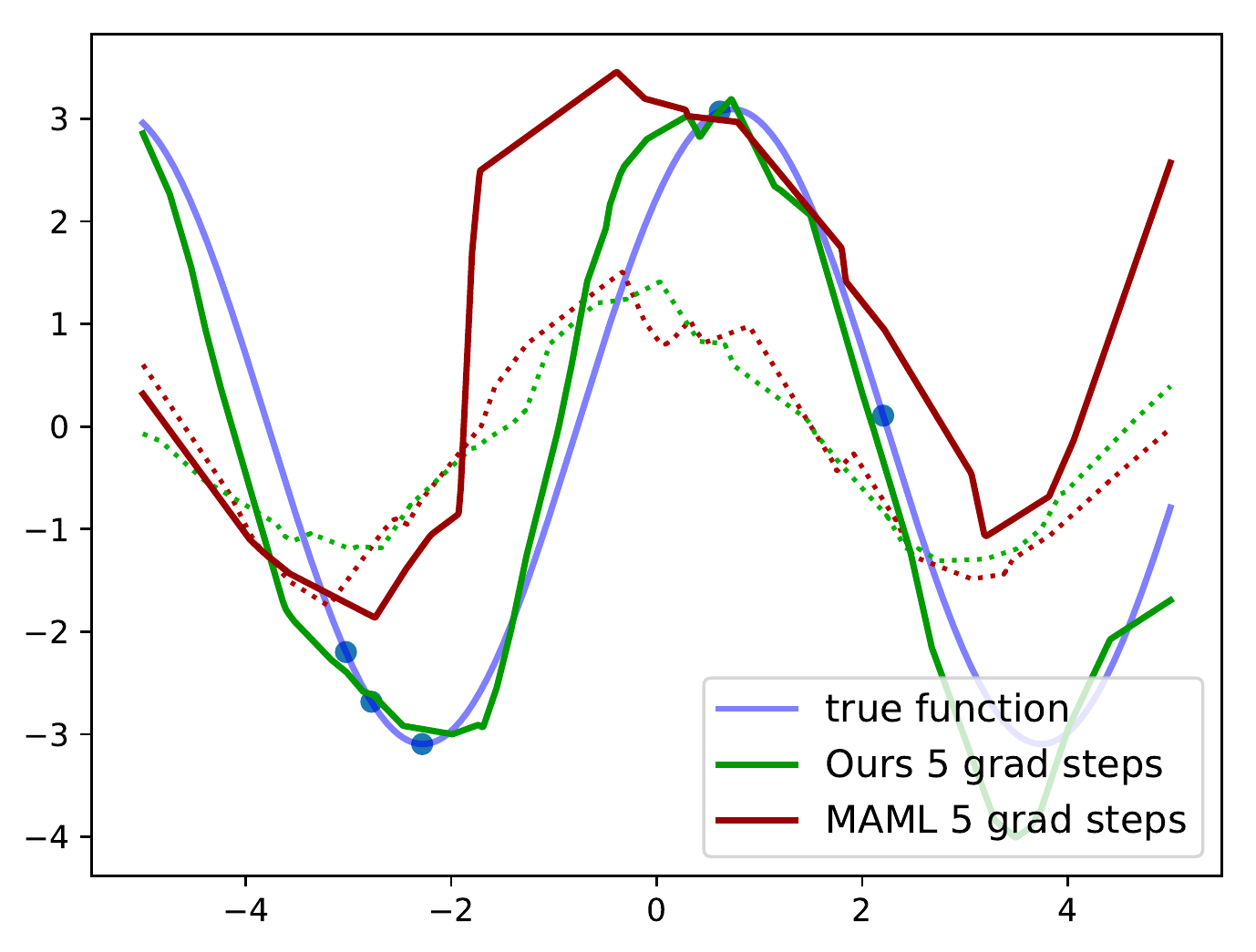}
\vfill
\includegraphics[width=0.48\linewidth]{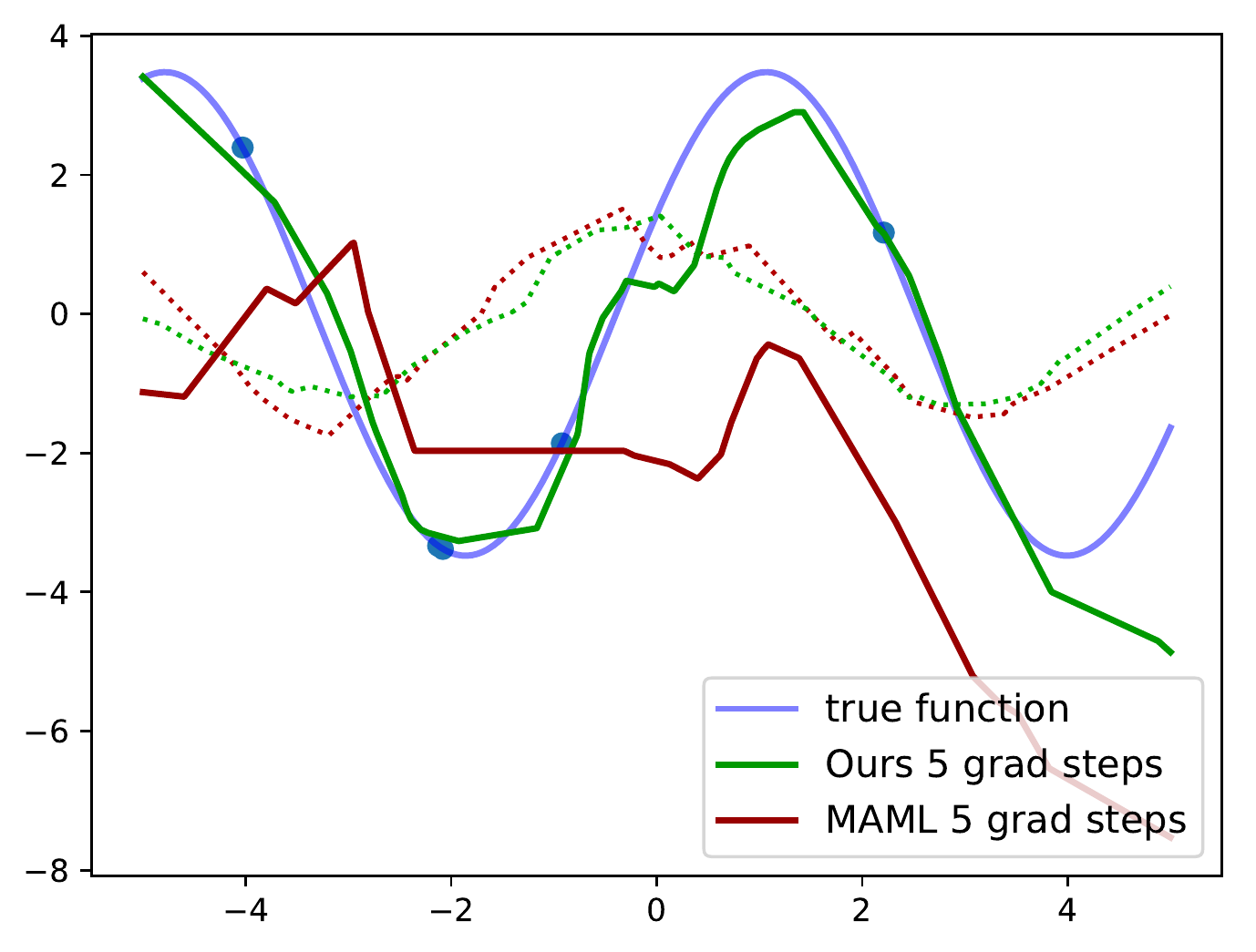}
\includegraphics[width=0.48\linewidth]{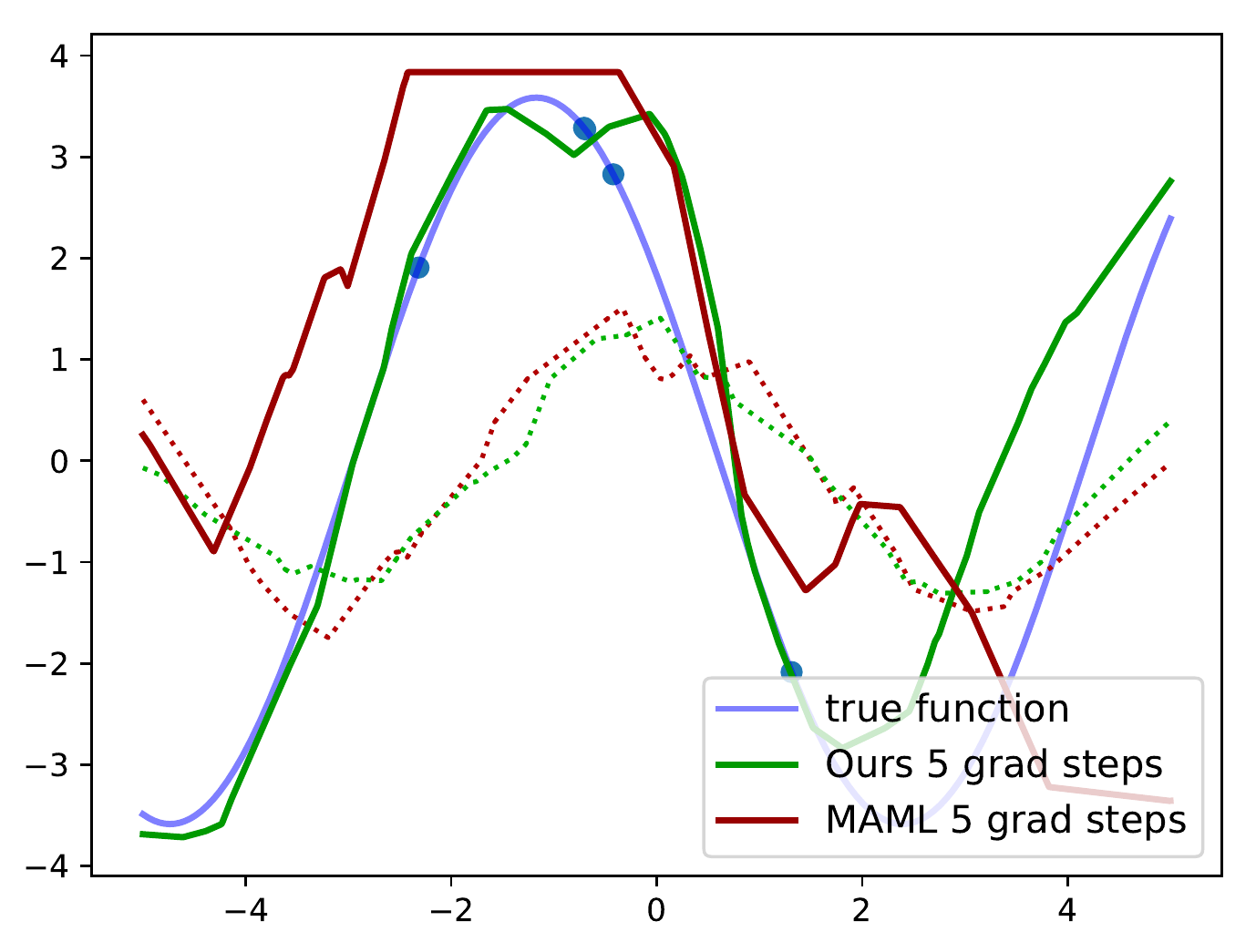}\\
(a) 5 shot regression
\end{minipage}
\begin{minipage}[b]{.5\textwidth}
\centering
\includegraphics[width=0.48\linewidth]{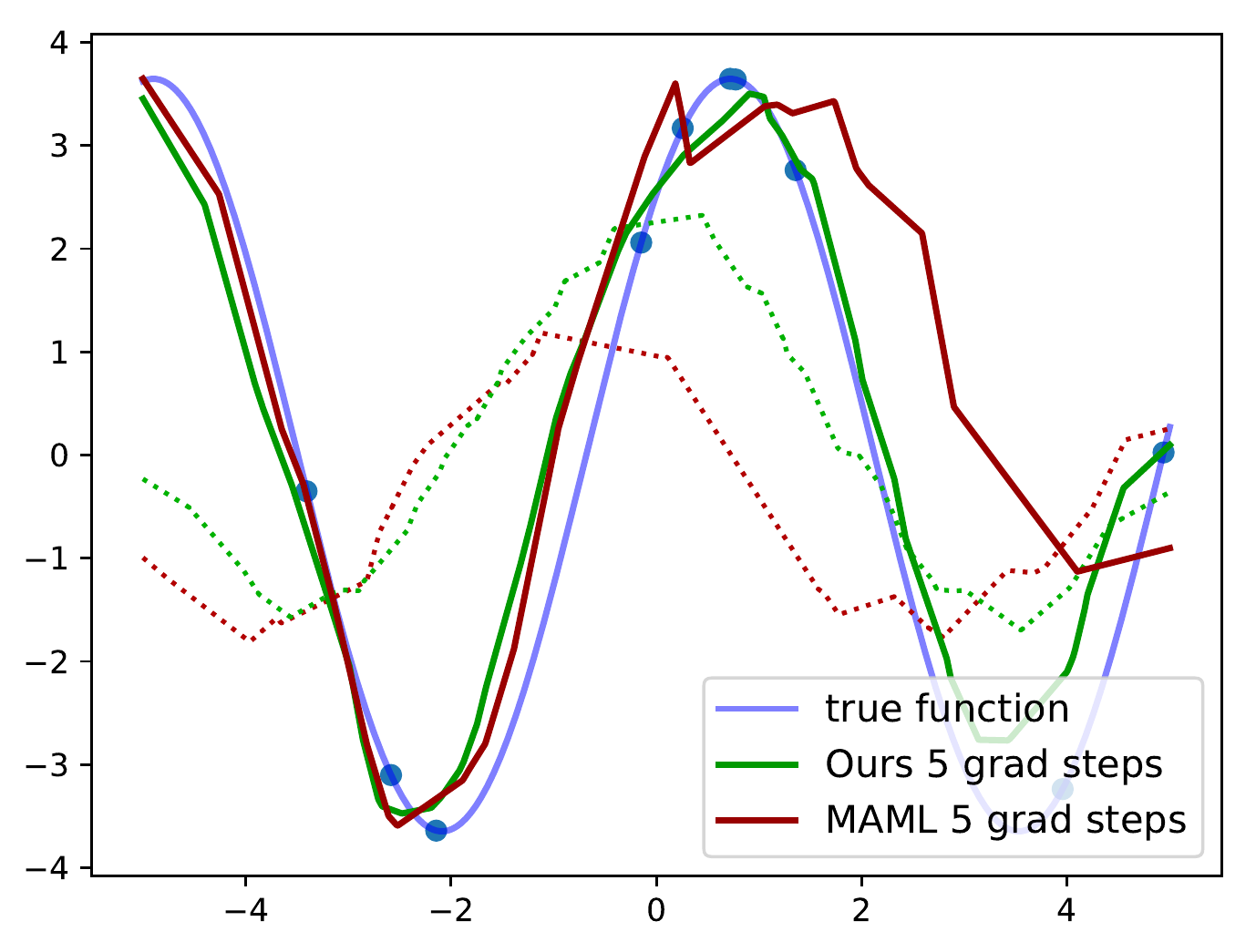}
\includegraphics[width=0.48\linewidth]{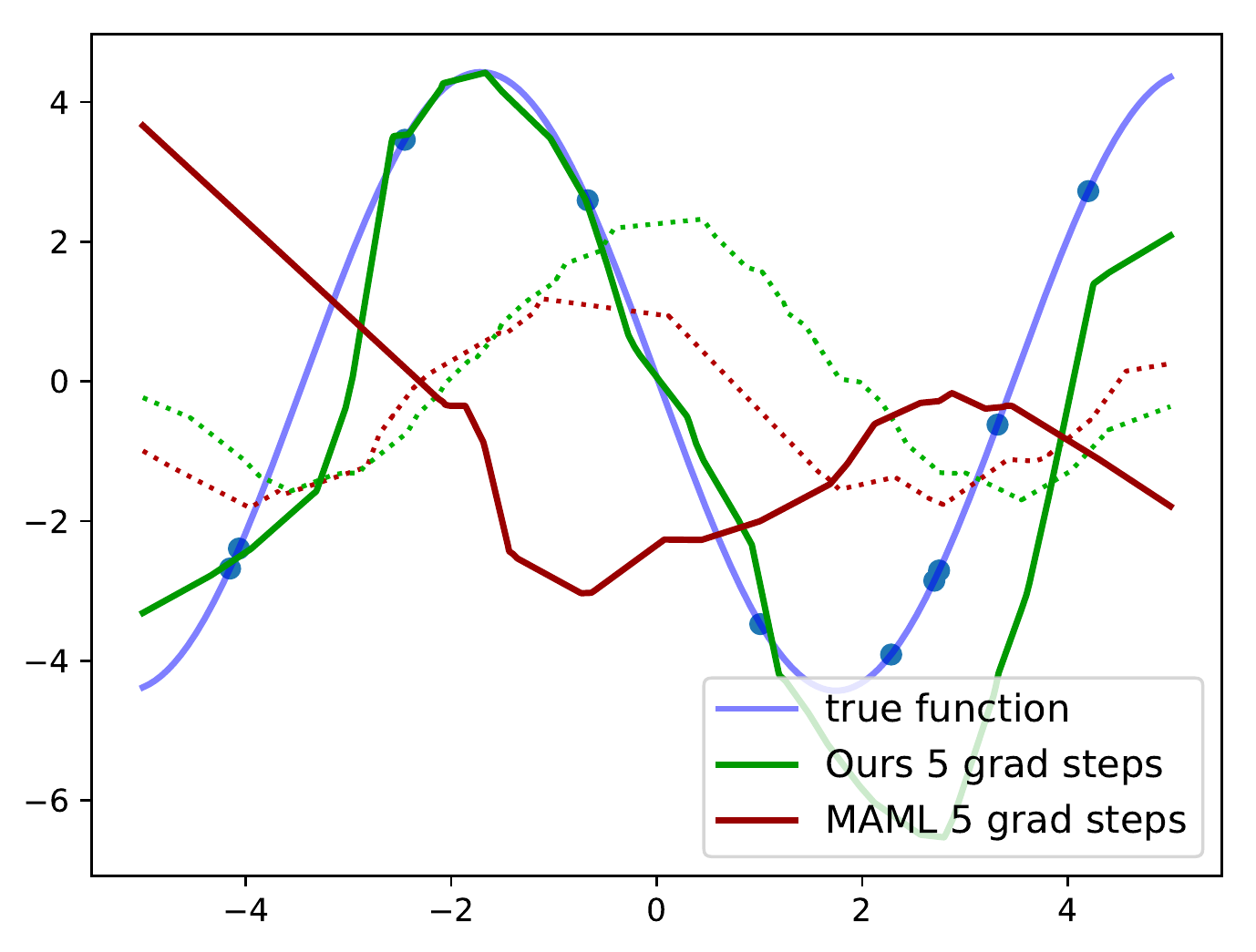}
\vfill
\includegraphics[width=0.48\linewidth]{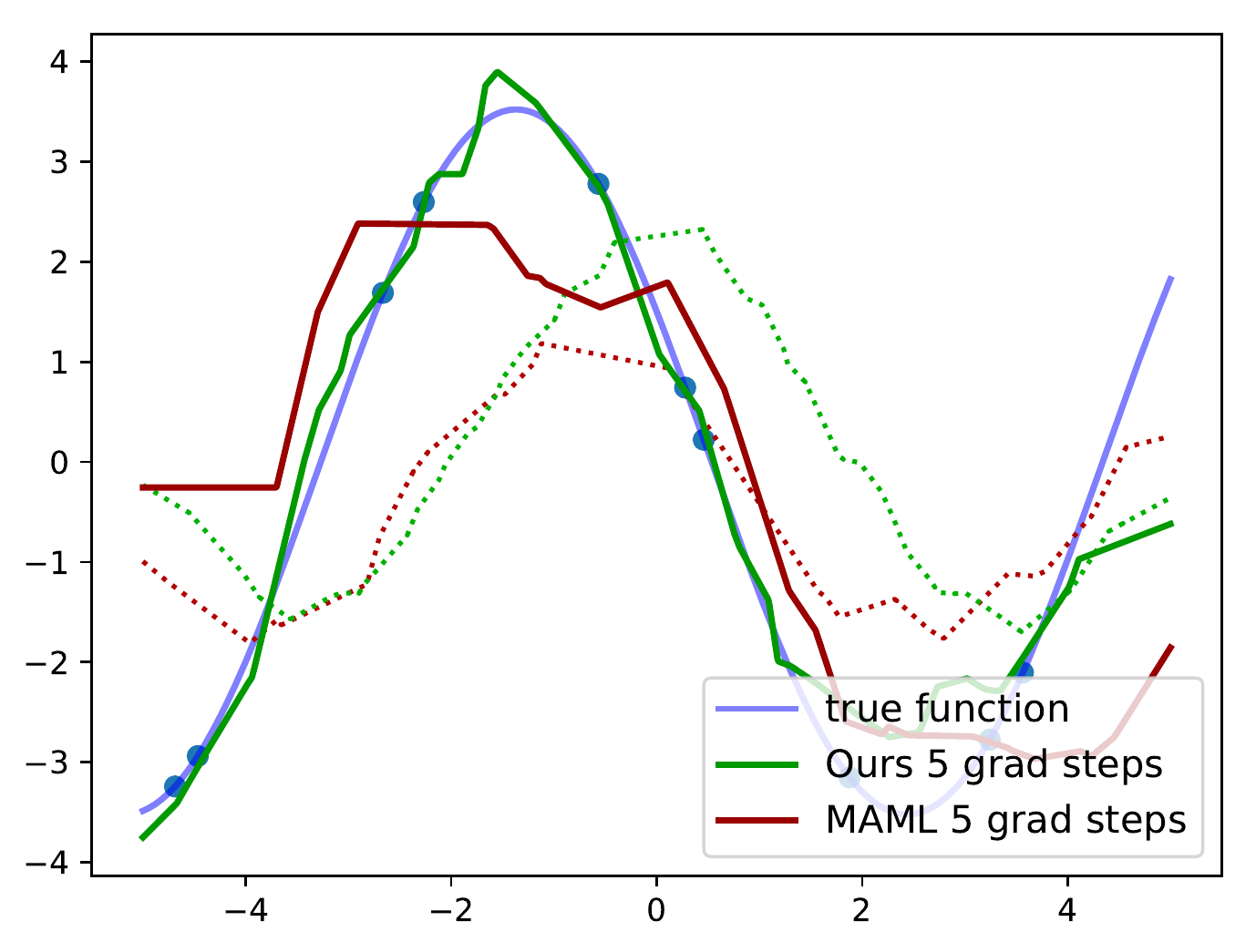}
\includegraphics[width=0.48\linewidth]{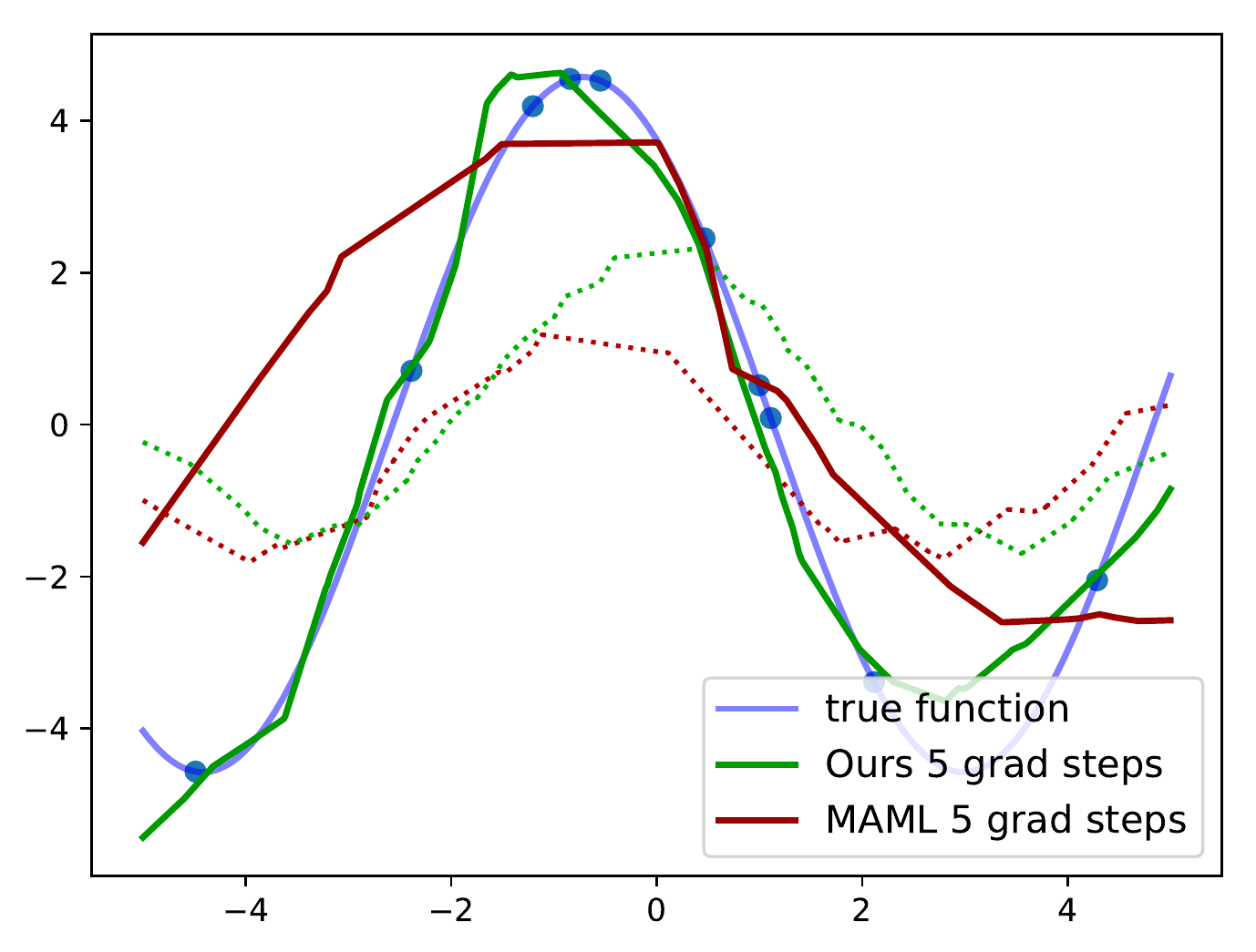}
\vfill
\includegraphics[width=0.48\linewidth]{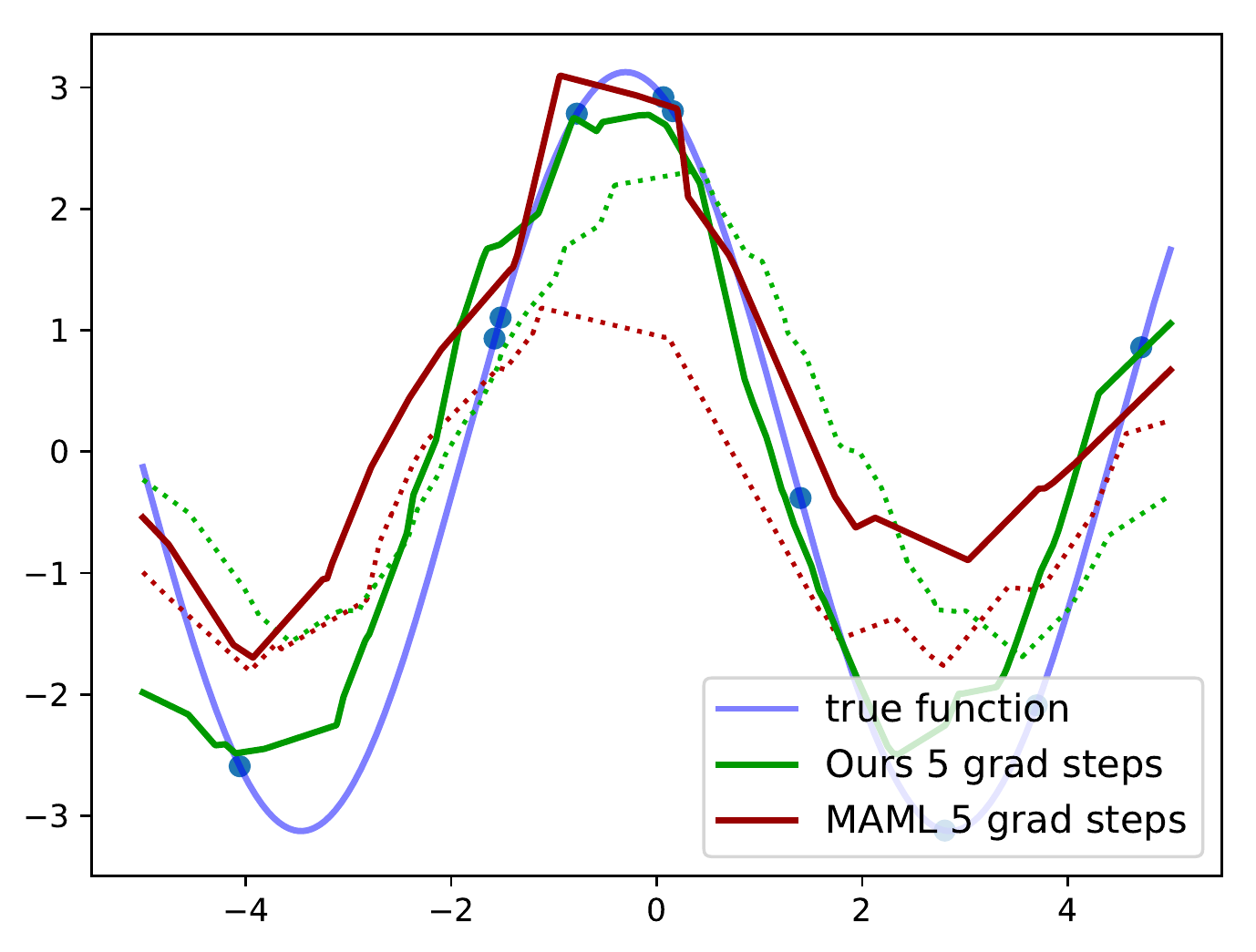}
\includegraphics[width=0.48\linewidth]{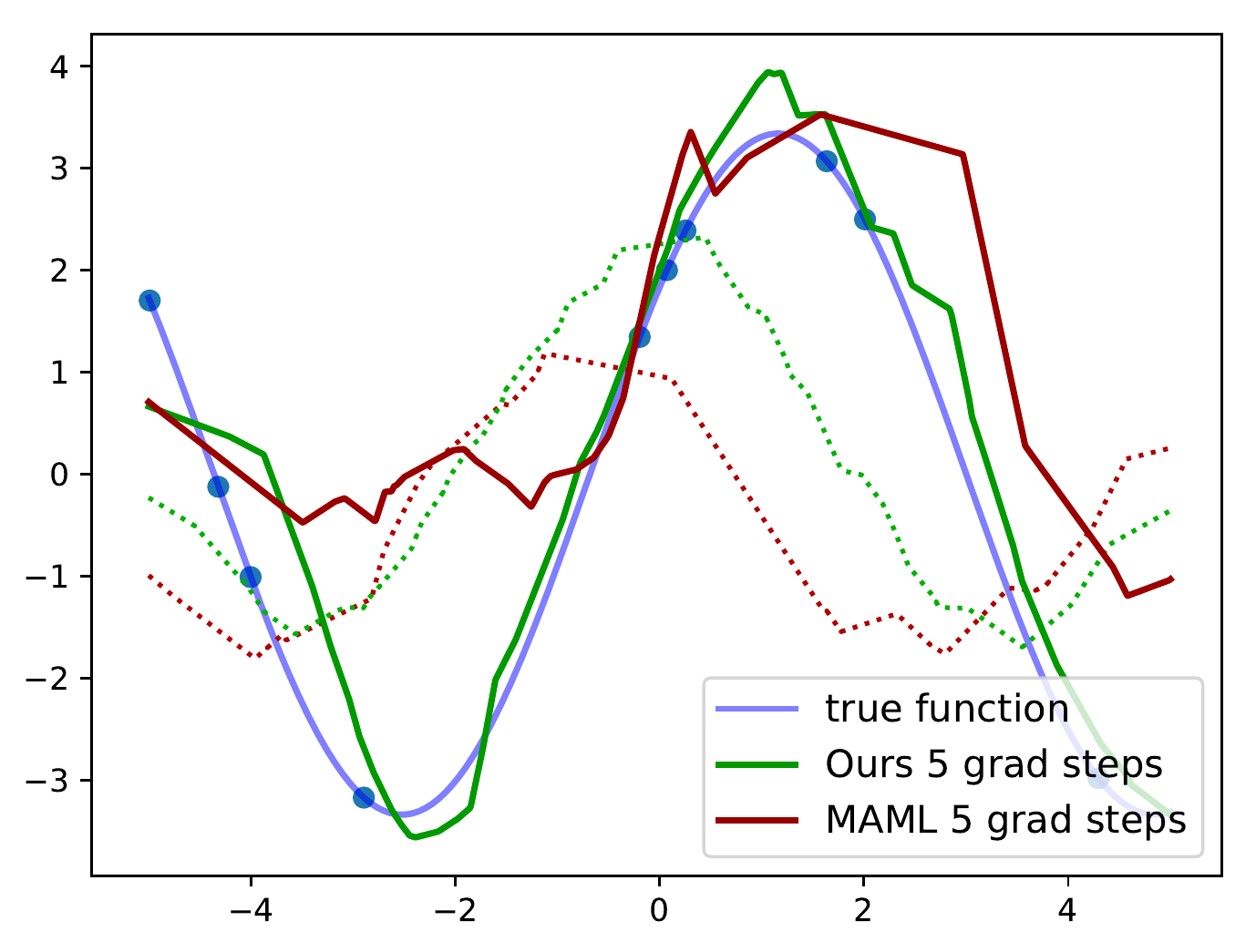}
\vfill
\includegraphics[width=0.48\linewidth]{supple_figures/regression_10/235_test_both_plot.pdf}
\includegraphics[width=0.48\linewidth]{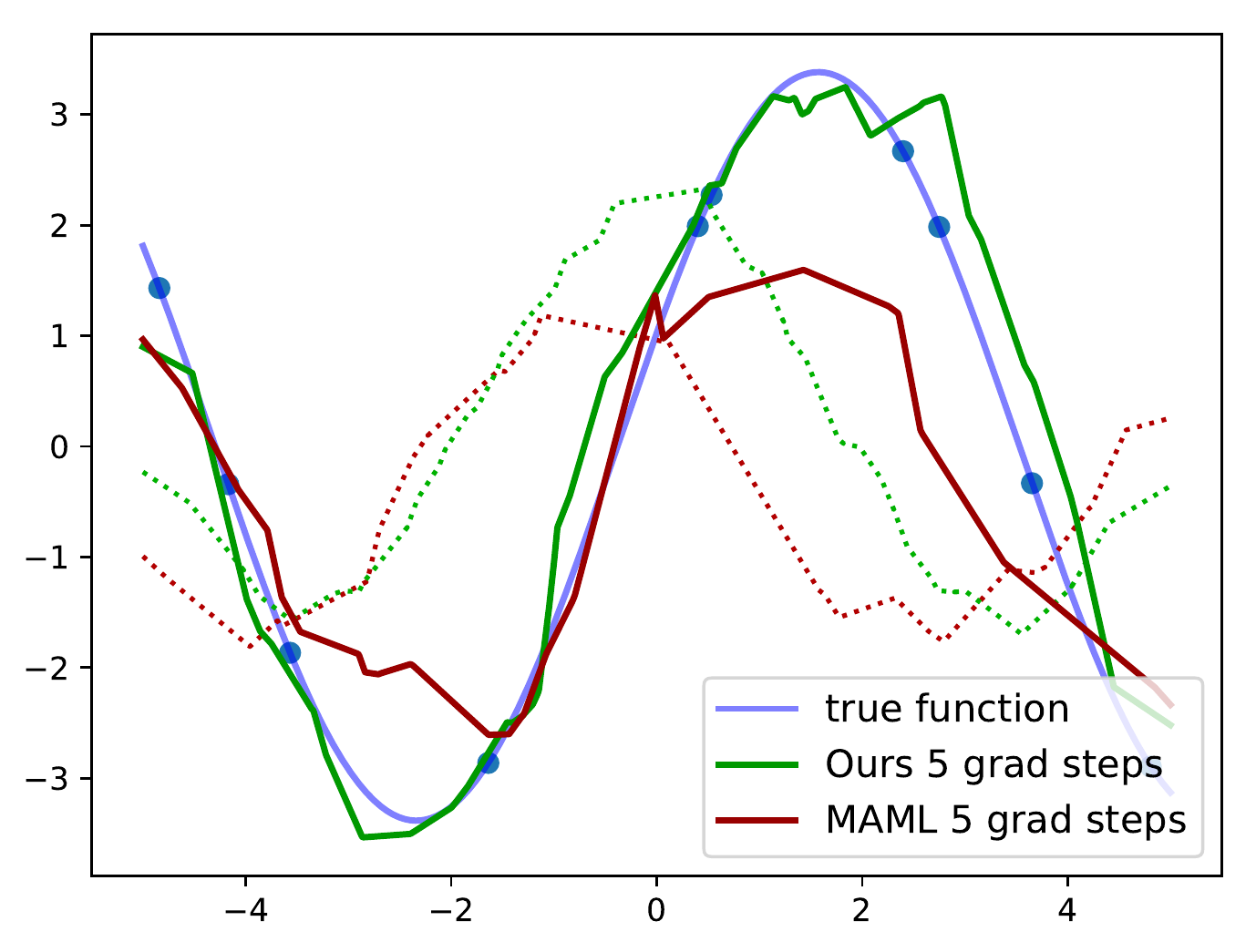}\\
(b) 10 shot regression
\end{minipage}
\caption{Qualitative results for the K$\in [5,10]$-shot sinusoid regression$\left( y (x) = Asin (\omega x + b) \right)$. Parameters are sampled from the same distribution for training and evaluation.}
\vspace{-1em}
\label{fig:regression figure 1}
\end{figure*}

\begin{figure*}
\begin{minipage}[b]{.5\textwidth}
\centering
\includegraphics[width=0.48\linewidth]{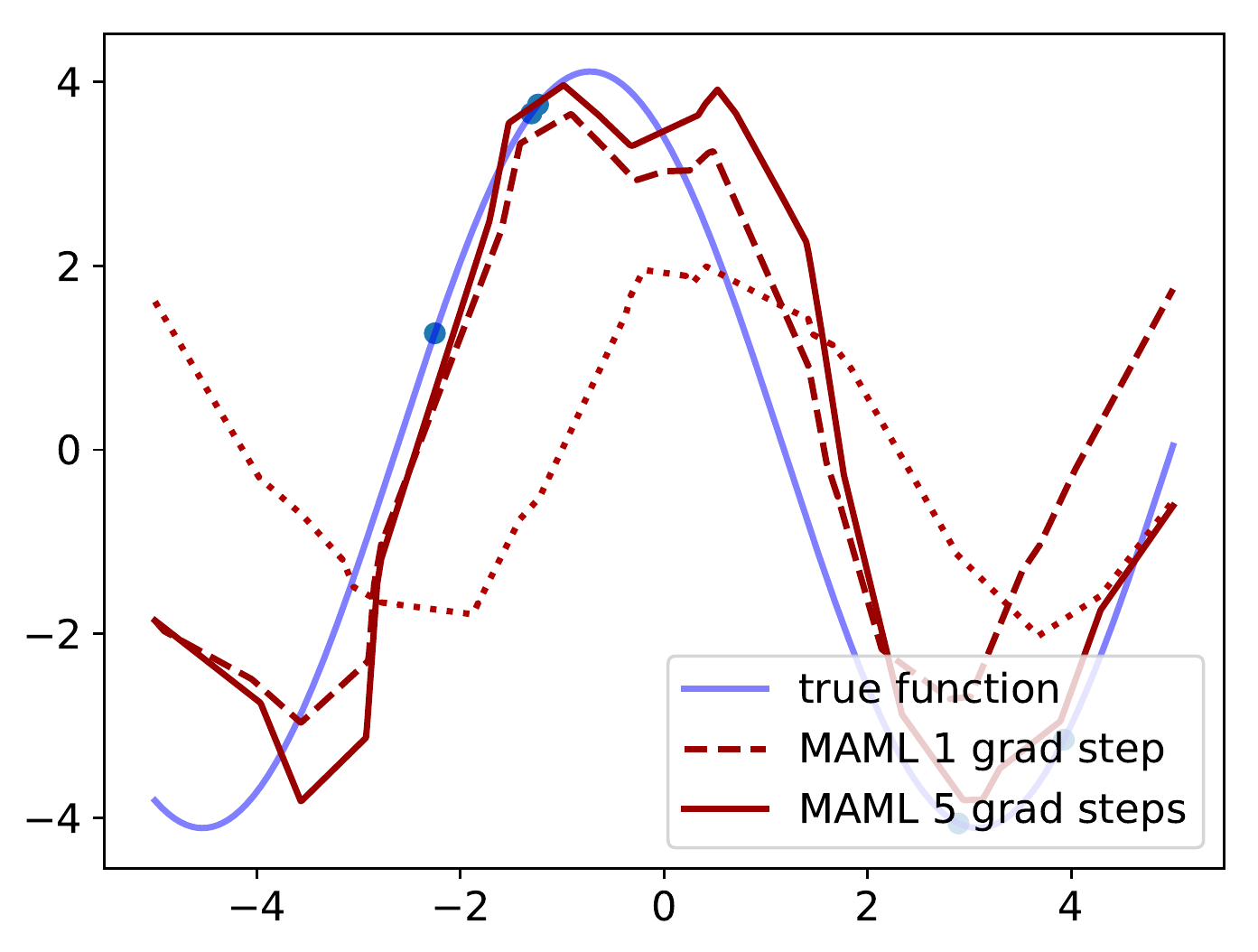}
\includegraphics[width=0.48\linewidth]{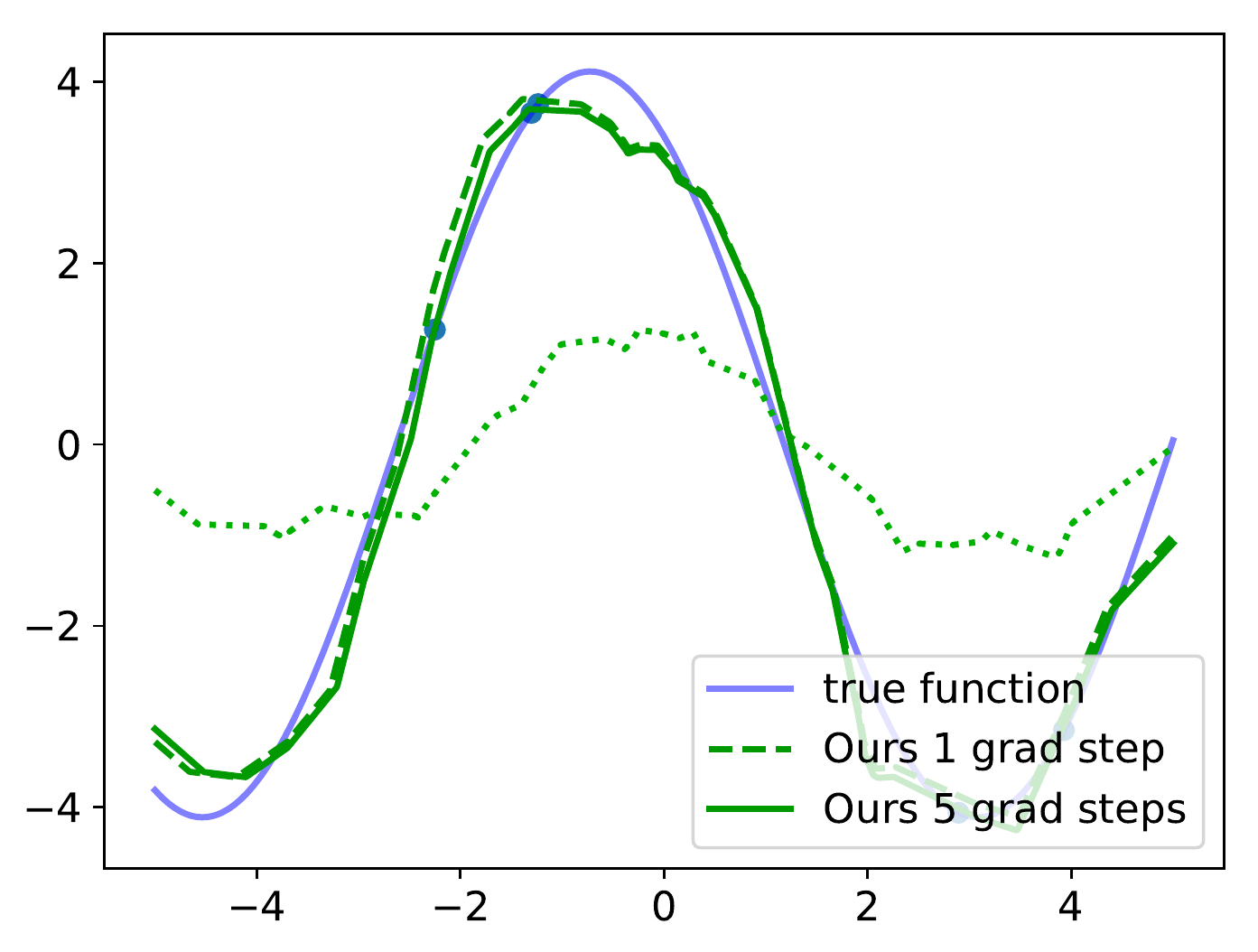}
\vfill
\includegraphics[width=0.48\linewidth]{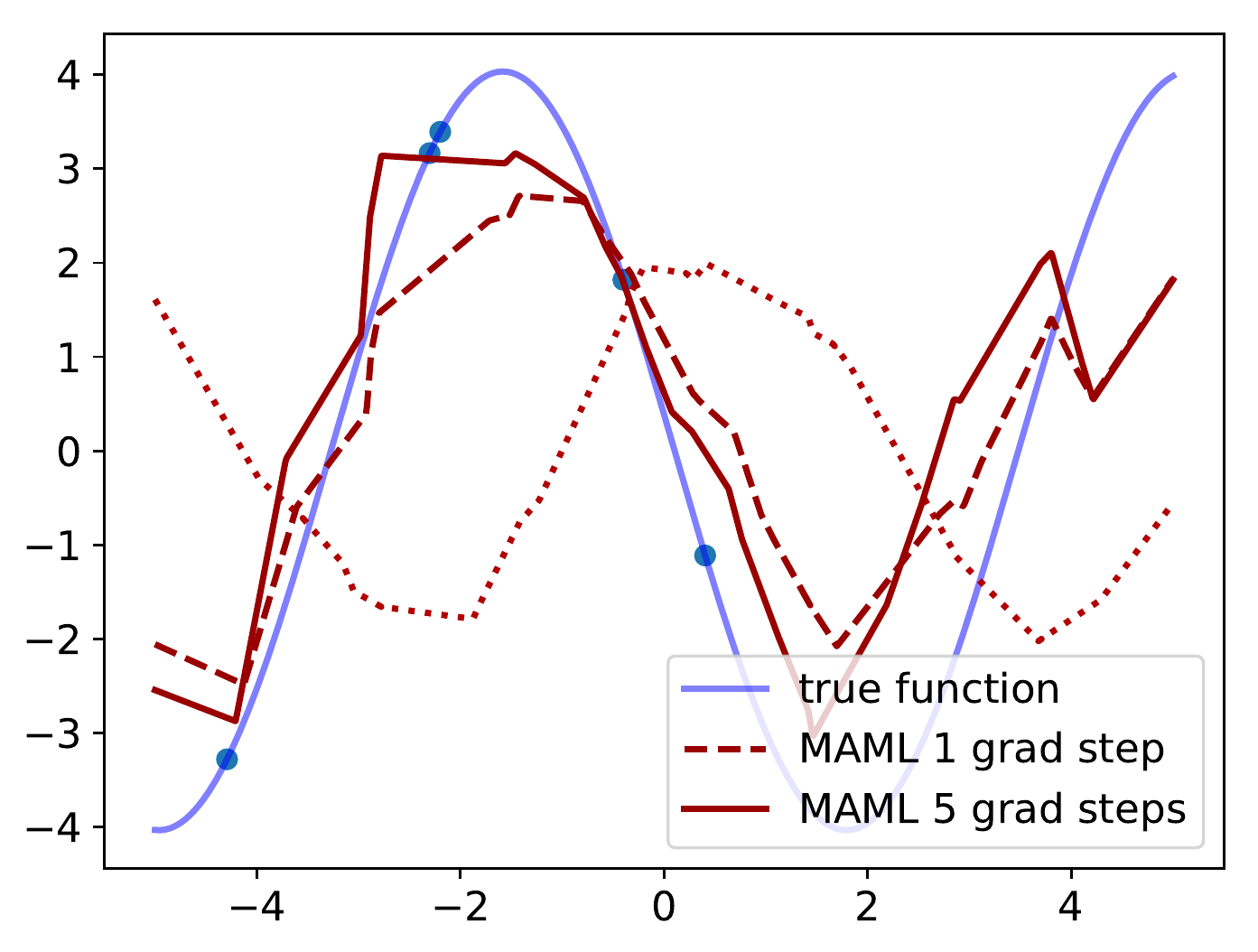}
\includegraphics[width=0.48\linewidth]{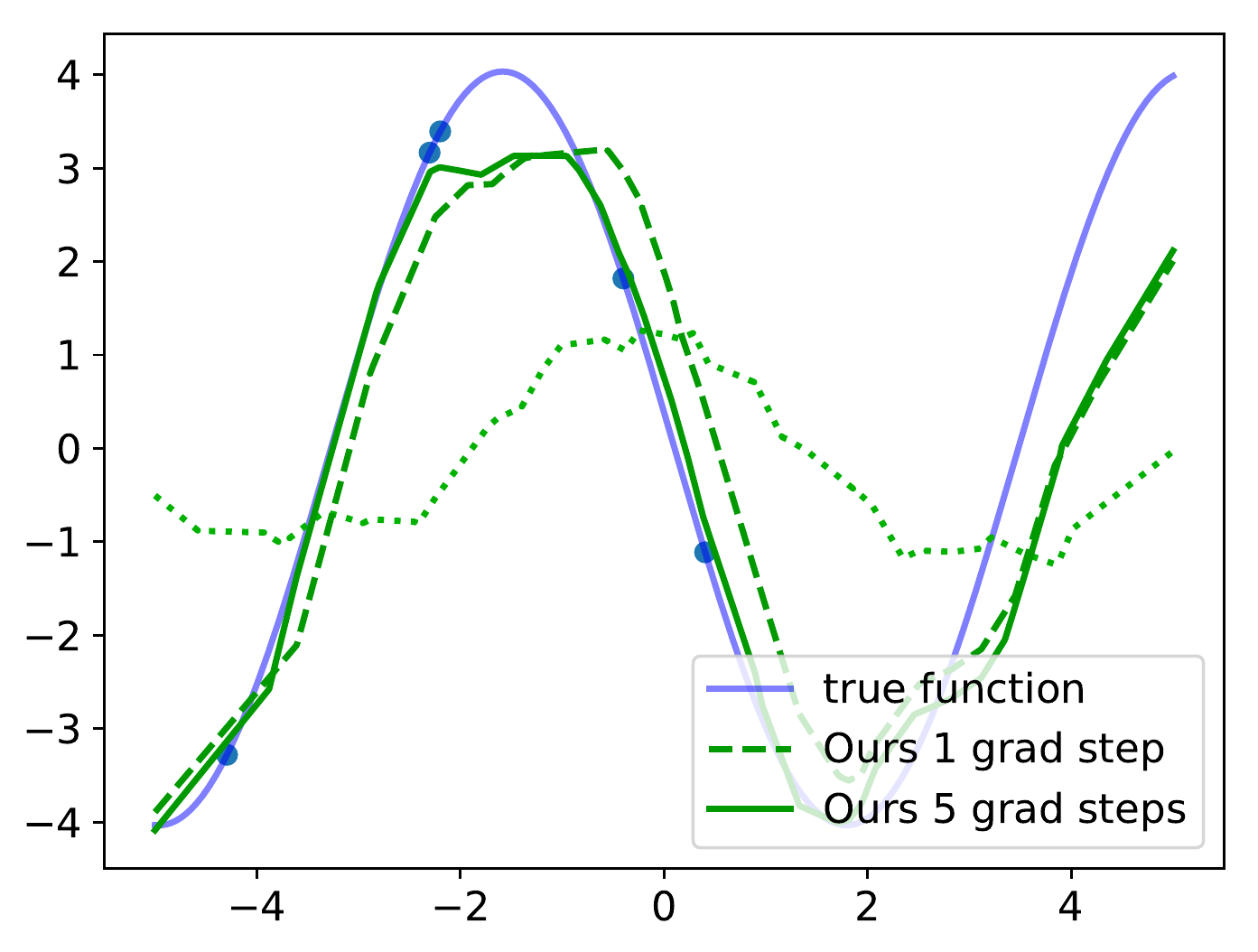}
\vfill
\includegraphics[width=0.48\linewidth]{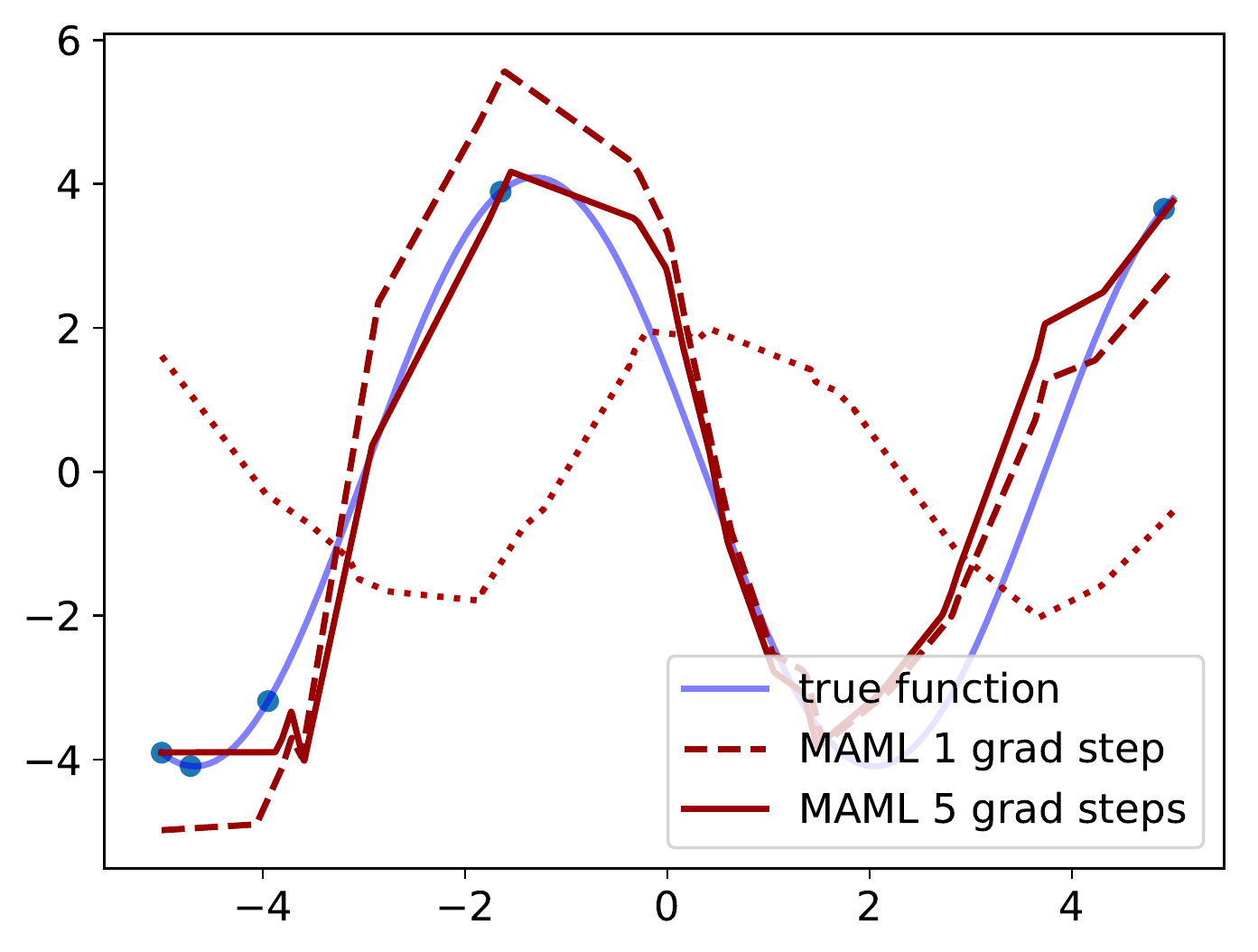}
\includegraphics[width=0.48\linewidth]{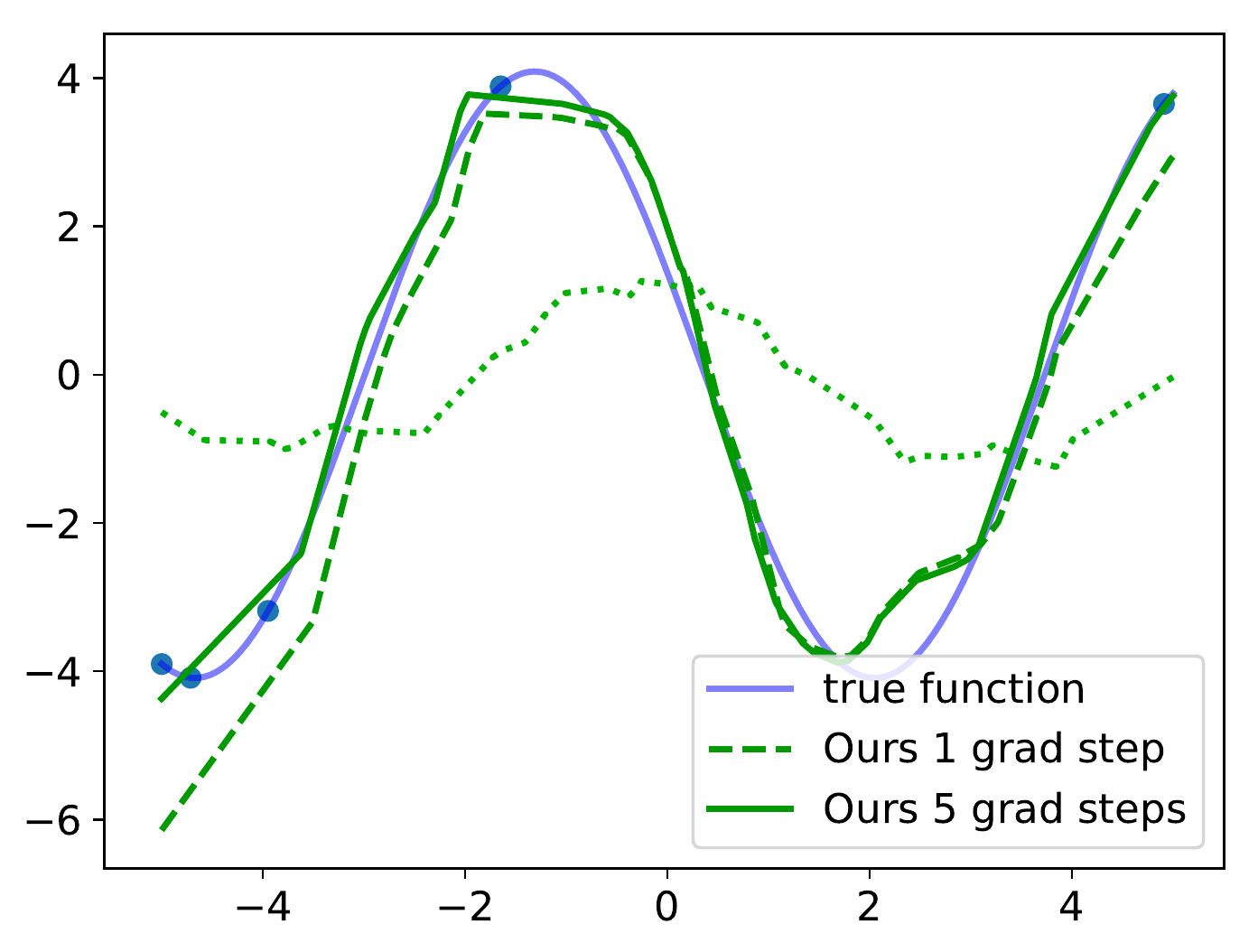}
\vfill
\includegraphics[width=0.48\linewidth]{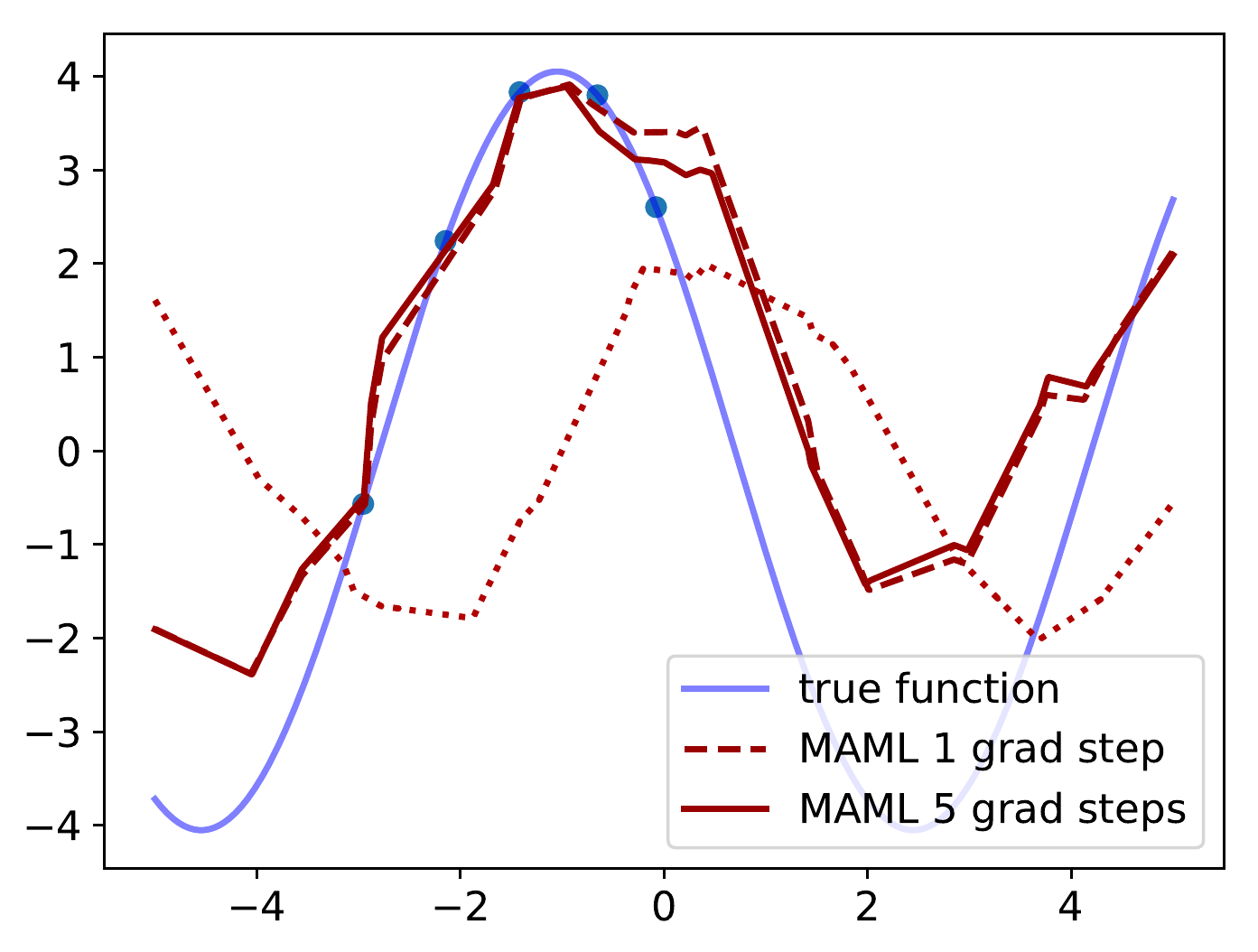}
\includegraphics[width=0.48\linewidth]{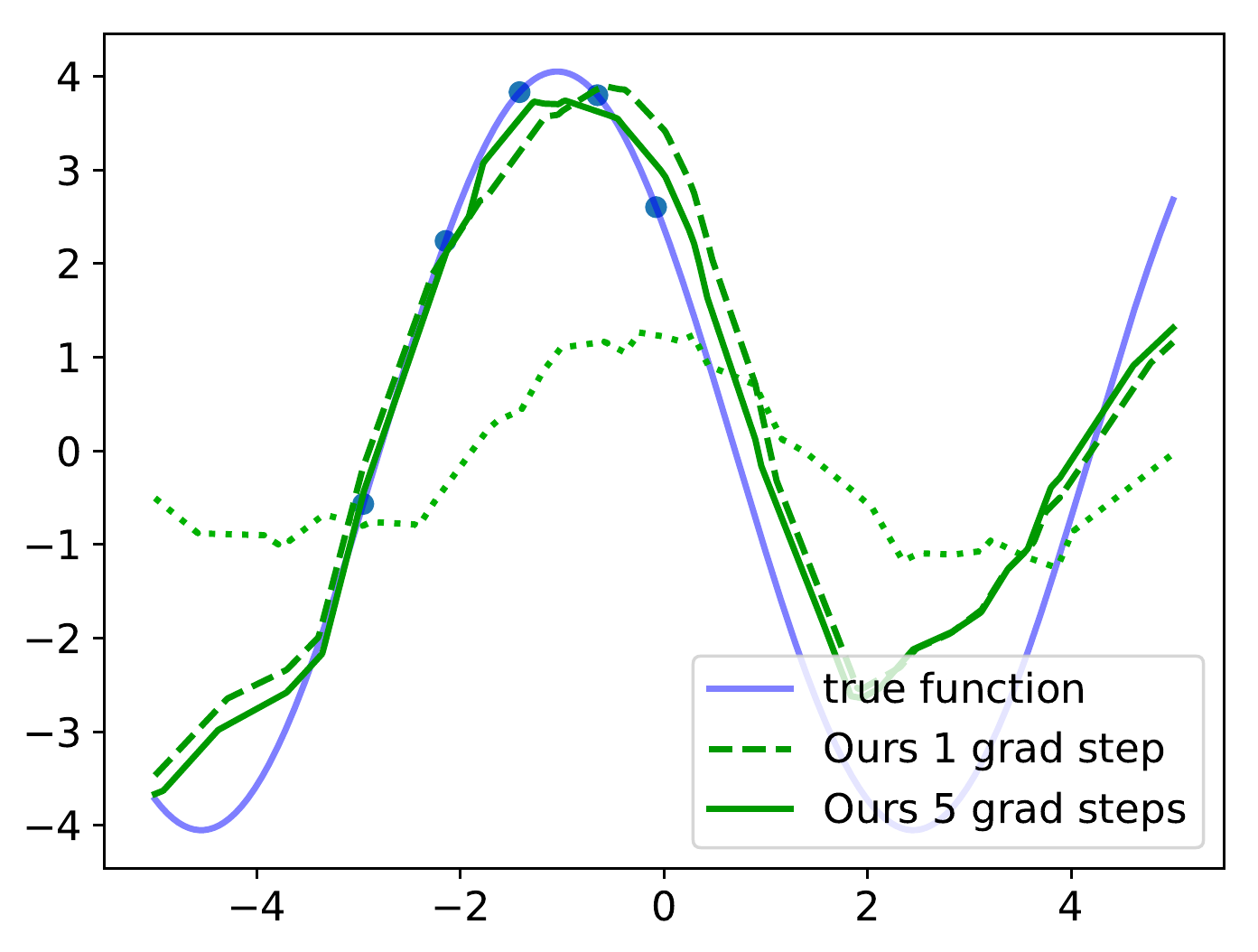}\\
(a) Non-overlapped task distributions on 5-shot
\end{minipage}
\begin{minipage}[b]{.5\textwidth}
\centering
\includegraphics[width=0.48\linewidth]{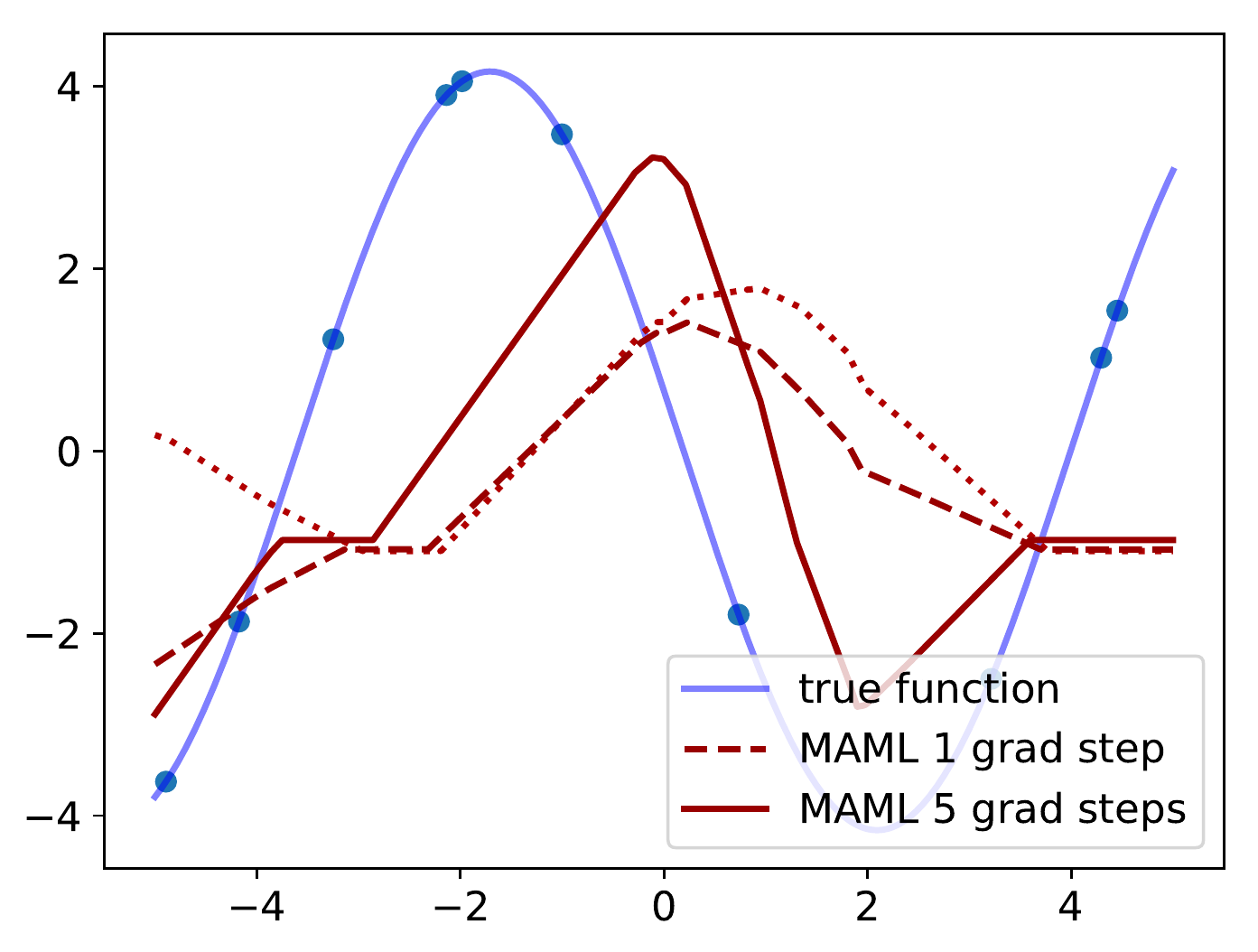}
\includegraphics[width=0.48\linewidth]{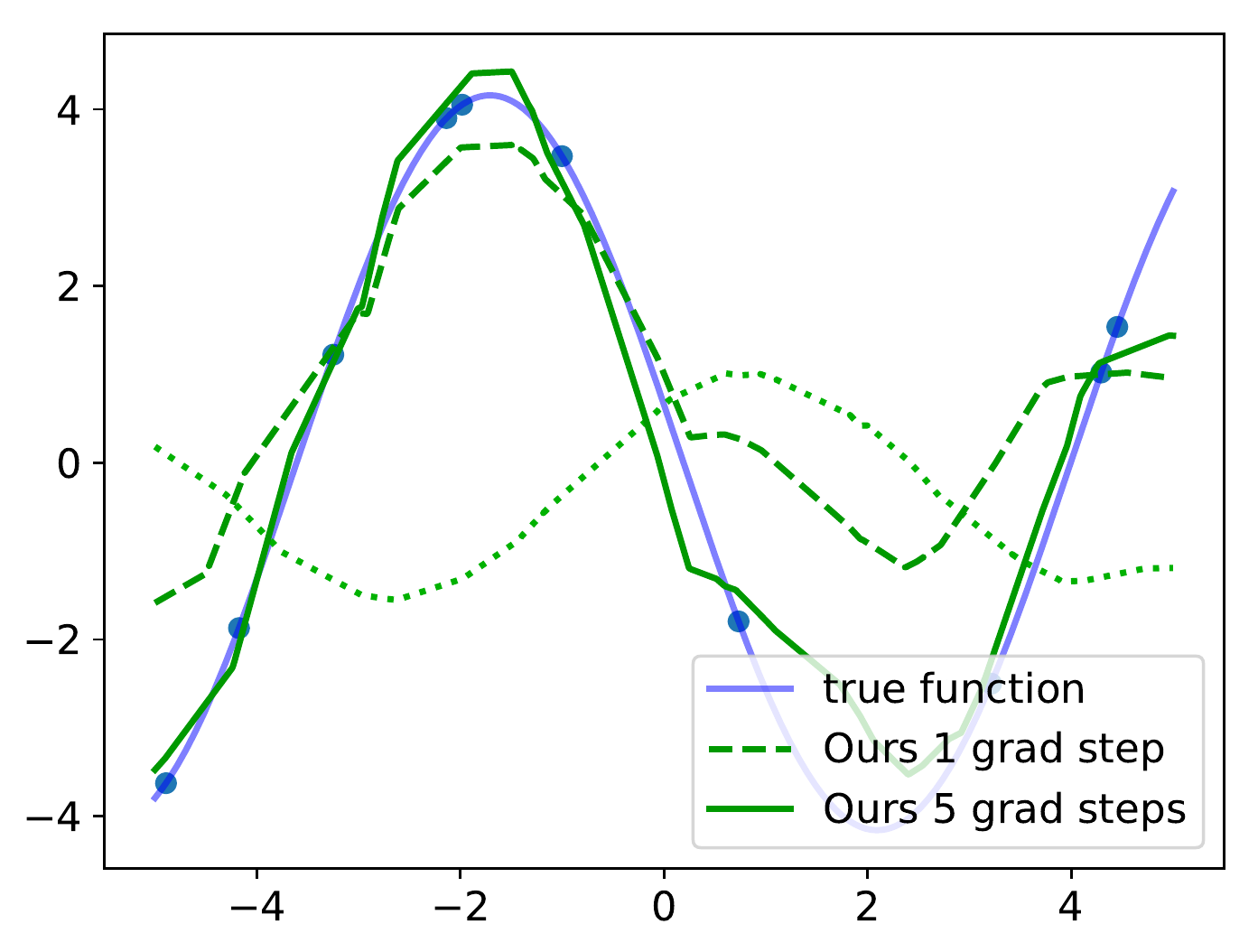}
\vfill
\includegraphics[width=0.48\linewidth]{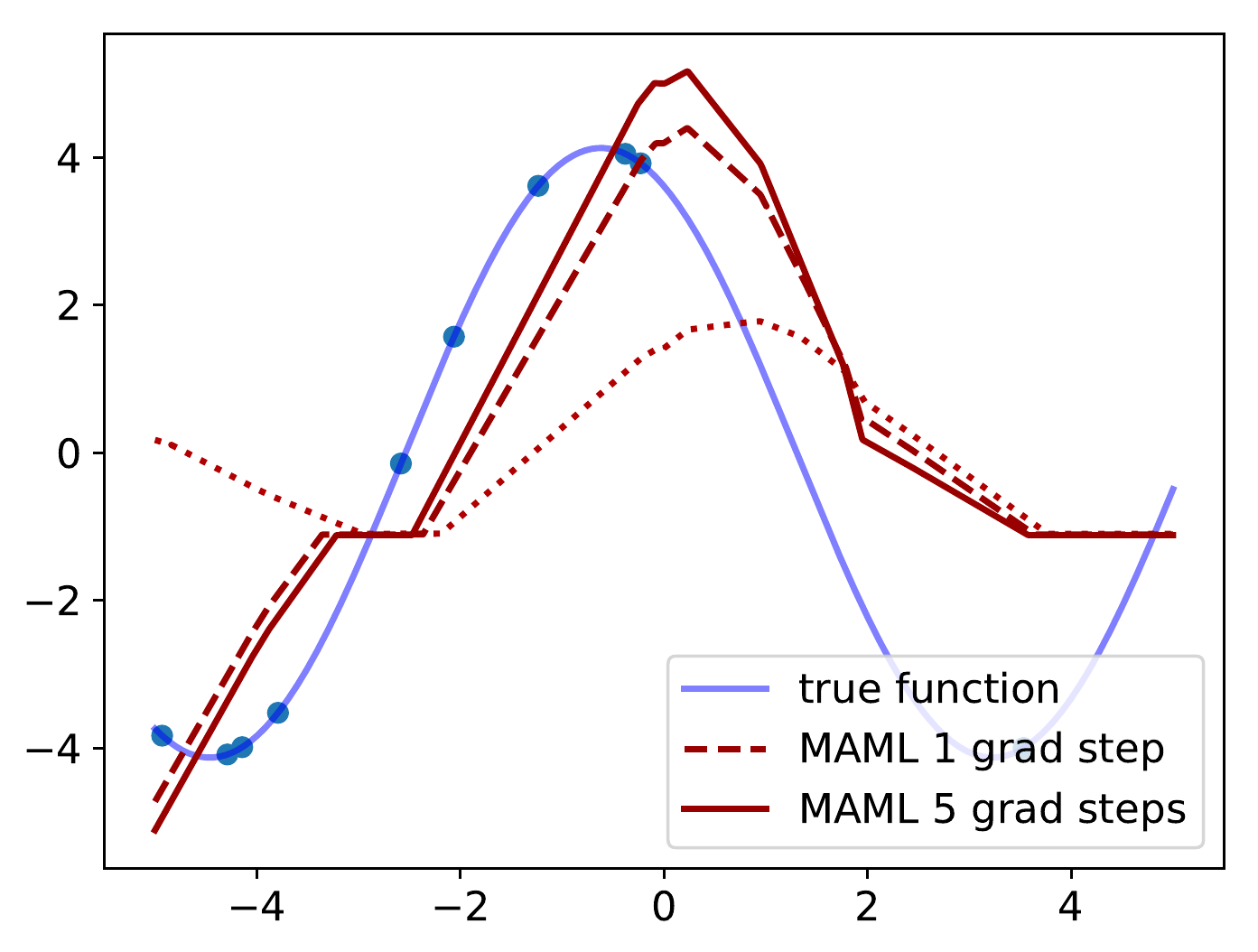}
\includegraphics[width=0.48\linewidth]{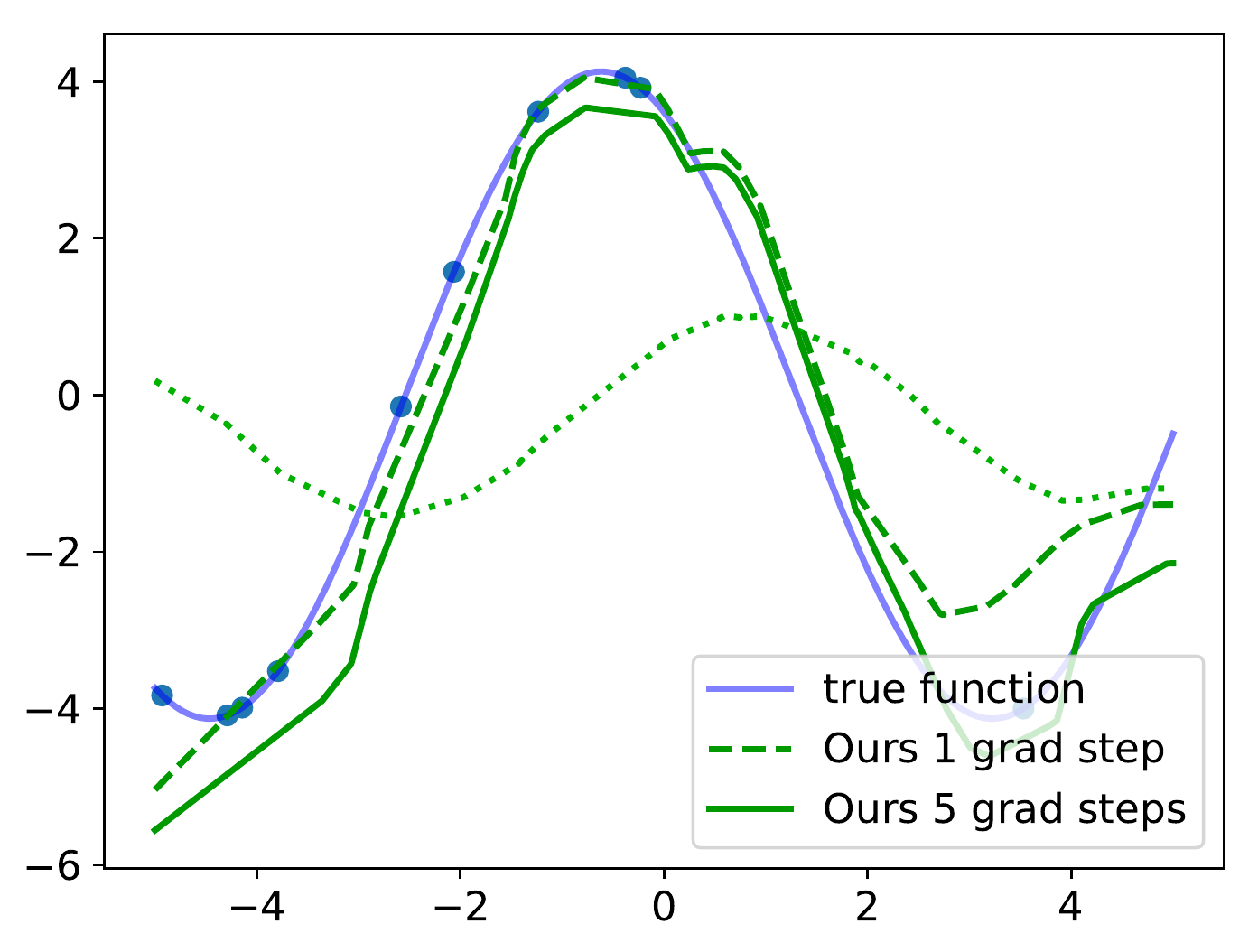}
\vfill
\includegraphics[width=0.48\linewidth]{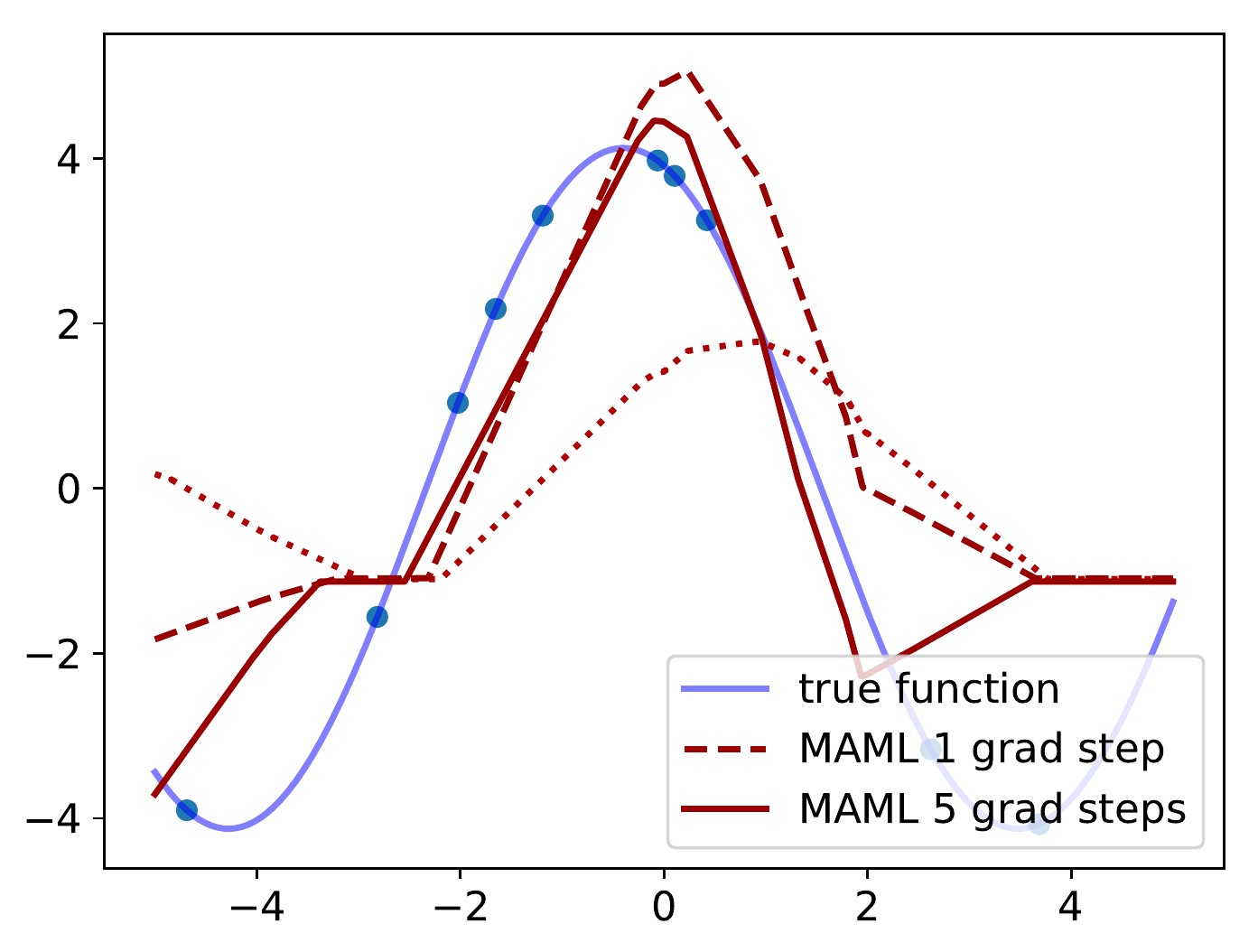}
\includegraphics[width=0.48\linewidth]{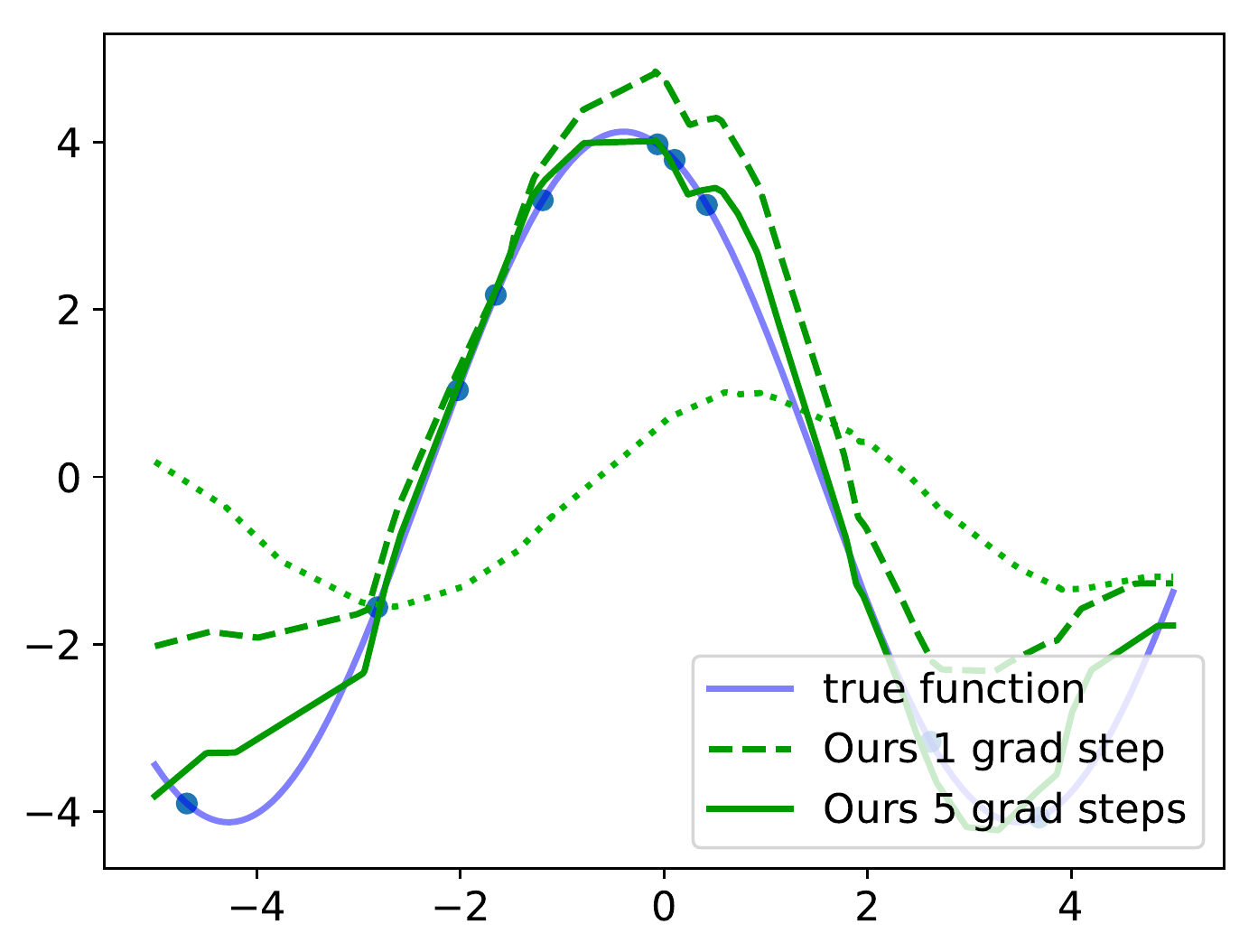}
\vfill
\includegraphics[width=0.48\linewidth]{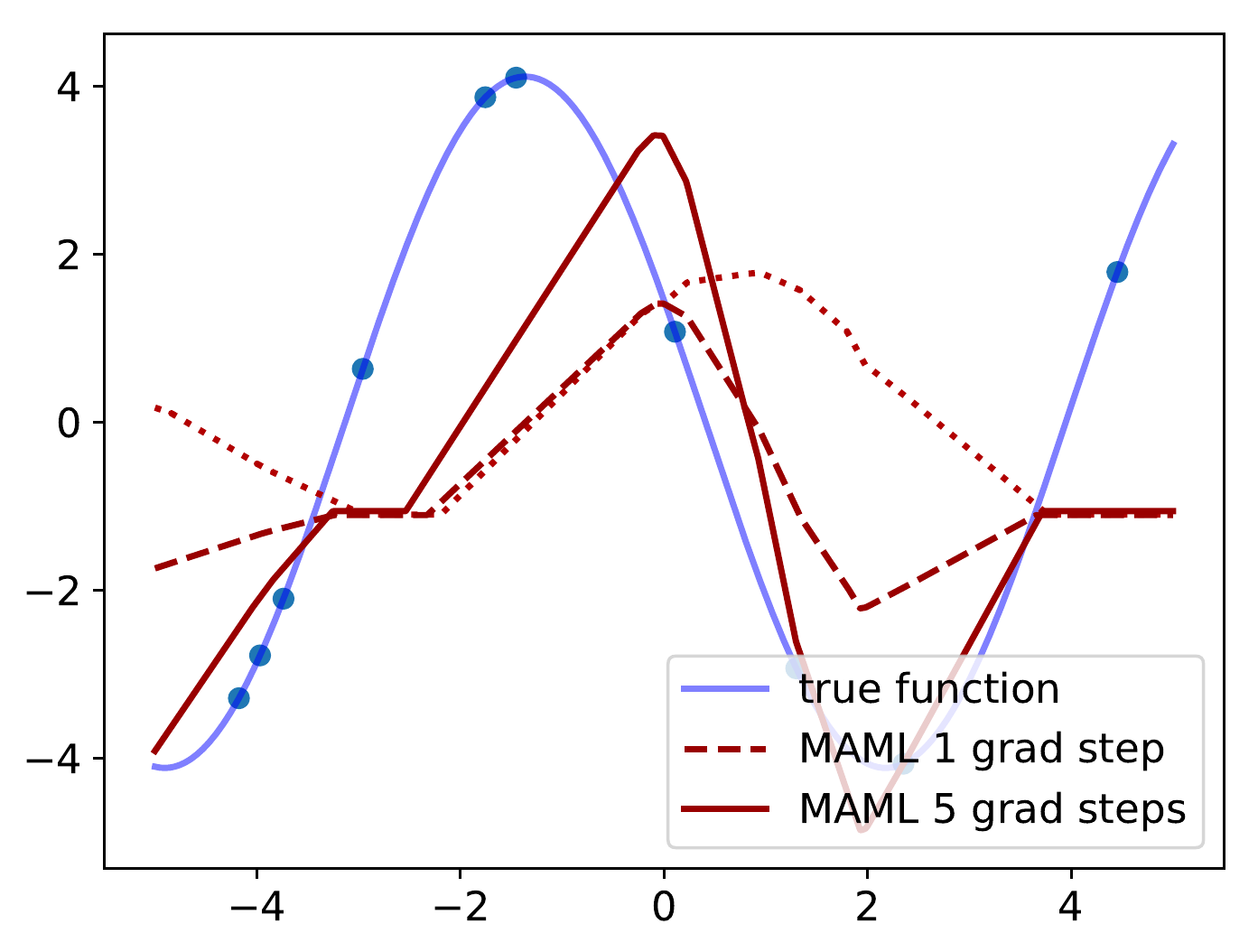}
\includegraphics[width=0.48\linewidth]{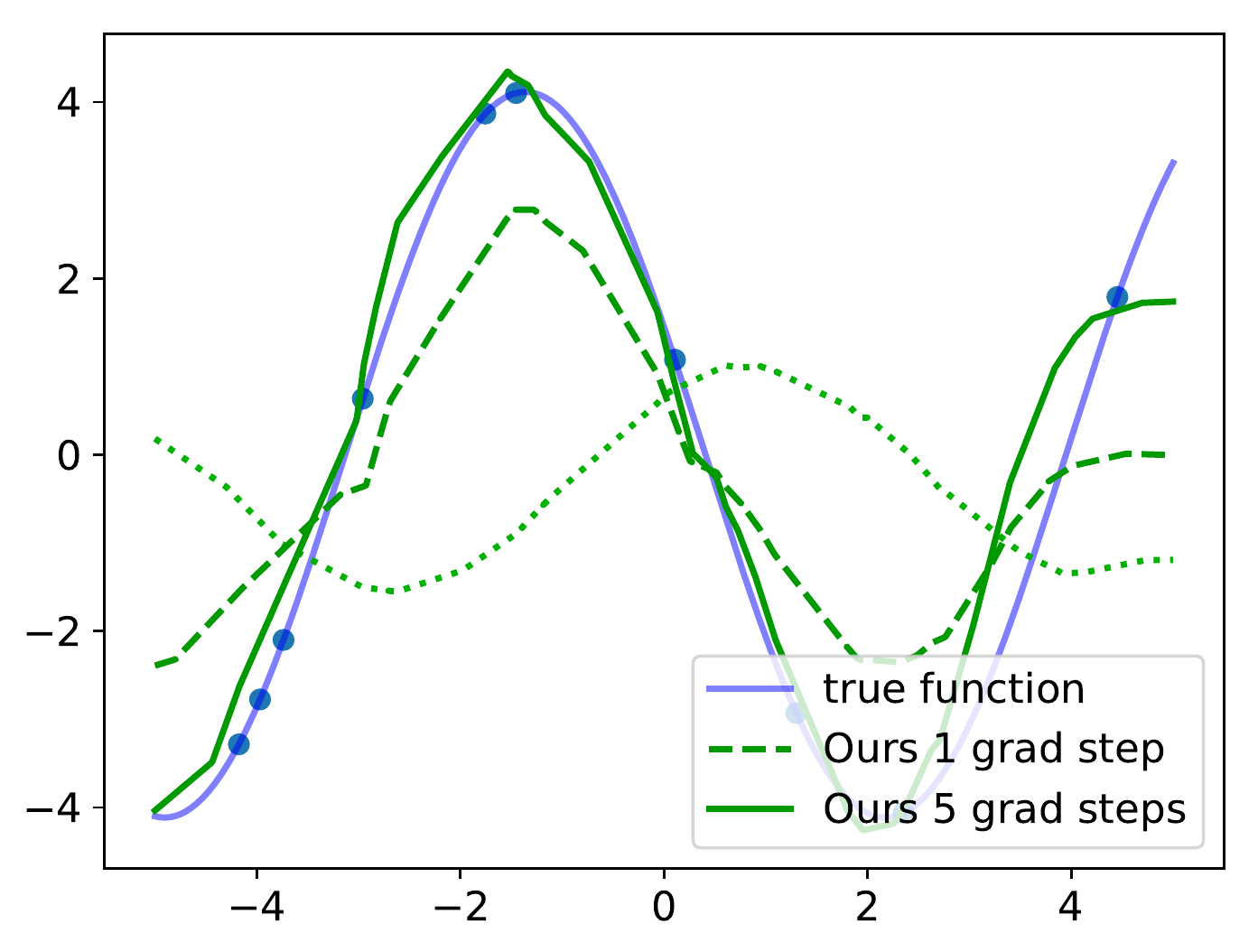}\\
(b) Non-overlapped task distributions on 10-shot
\end{minipage}
\caption{Qualitative results for the K$\in [5,10]$-shot sinusoid regression$\left( y (x) = Asin (\omega x + b) \right)$. Parameters are sampled from non-overlapped ranges for training and evaluation.}
\vspace{-1em}
\label{fig:regression figure 2}
\end{figure*}

In Figure \ref{fig:regression figure 1} and \ref{fig:regression figure 2}, we show a random sample of qualitative results from the $k$-shot sinusoid regression, where $k\in[5,10]$. The target function (or true function) is a sine curve $y(x) =Asin(\omega x+b)$ with the amplitude $A$, frequency $\omega$, phase $b$, and the input range $[-5.0, 5.0]$. The sampling range of amplitude, frequency, and phase defines a task distribution.
In Figure \ref{fig:regression figure 1}, we follow the general settings in \cite{finn2017model,Li2017meta}, where amplitude $A$, frequency $\omega$, and phase $b$ are sampled from the uniform distribution on intervals $[0.1,5.0]$, $[0.8, 1.2]$, and $[0, \pi]$, respectively. MAML+L2F demonstrates more accurate regression for both 5 and 10-shot cases, compared to the baseline, MAML. 
To further stress the generalization of the MAML+L2F initialization, we extensively increase the degree of conflicts between new tasks and the prior knowledge. To that end, we modify the setting such that amplitude, frequency, and phase are sampled from the non-overlapped ranges for training and evaluation. In training, amplitude $A$, frequency $\omega$, and phase $b$ are sampled from the uniform distribution on intervals $[0.1,3.0]$, $[0.8, 1.0]$, and $[0, \pi/2]$, respectively. In evaluation, amplitude $A$, frequency $\omega$, and phase $b$ are sampled from the uniform distribution on intervals $[3.0,5.0]$, $[1.0, 1.2]$, and $[\pi/2, \pi]$, respectively. In Figure \ref{fig:regression figure 2}, our method(MAML+L2F) exhibits better fitting and thus claims the better generalization than MAML for both 5 and 10-shot regression.
\subsection{Additional Quantitative results}
In Table \ref{tab:additional}, we compare the proposed method against other advanced MAML-based methods, which are generalizable across domains, specifically MuMoMAML and MAML++. As with results on classification, our method consistently outperforms in regression task.
\section{Reinforcement Learning}
\subsection{Additional Qualitative results}
\begin{figure*}
\begin{minipage}[b]{.5\textwidth}
\centering
\includegraphics[width=0.49\linewidth]{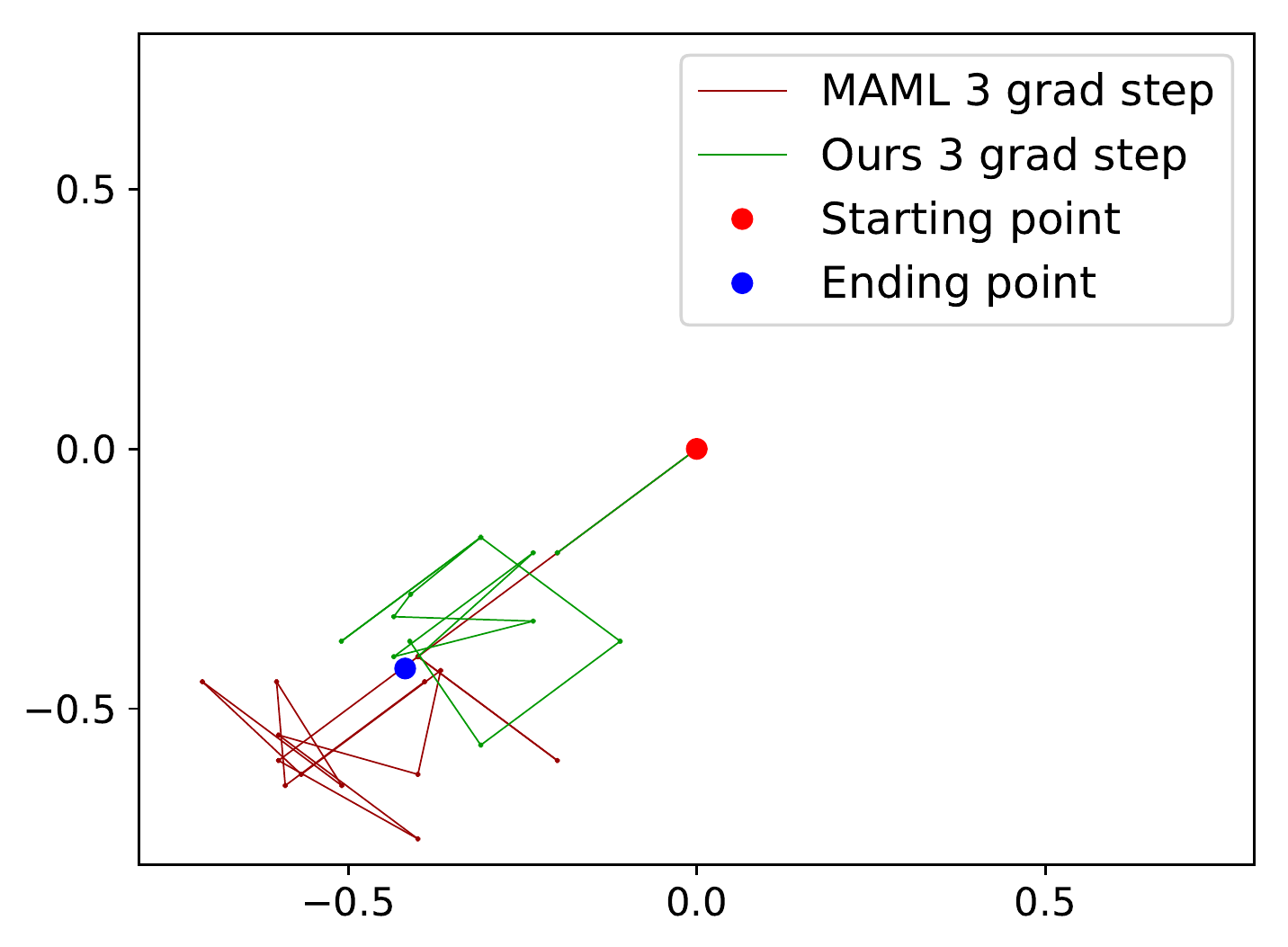}
\includegraphics[width=0.49\linewidth]{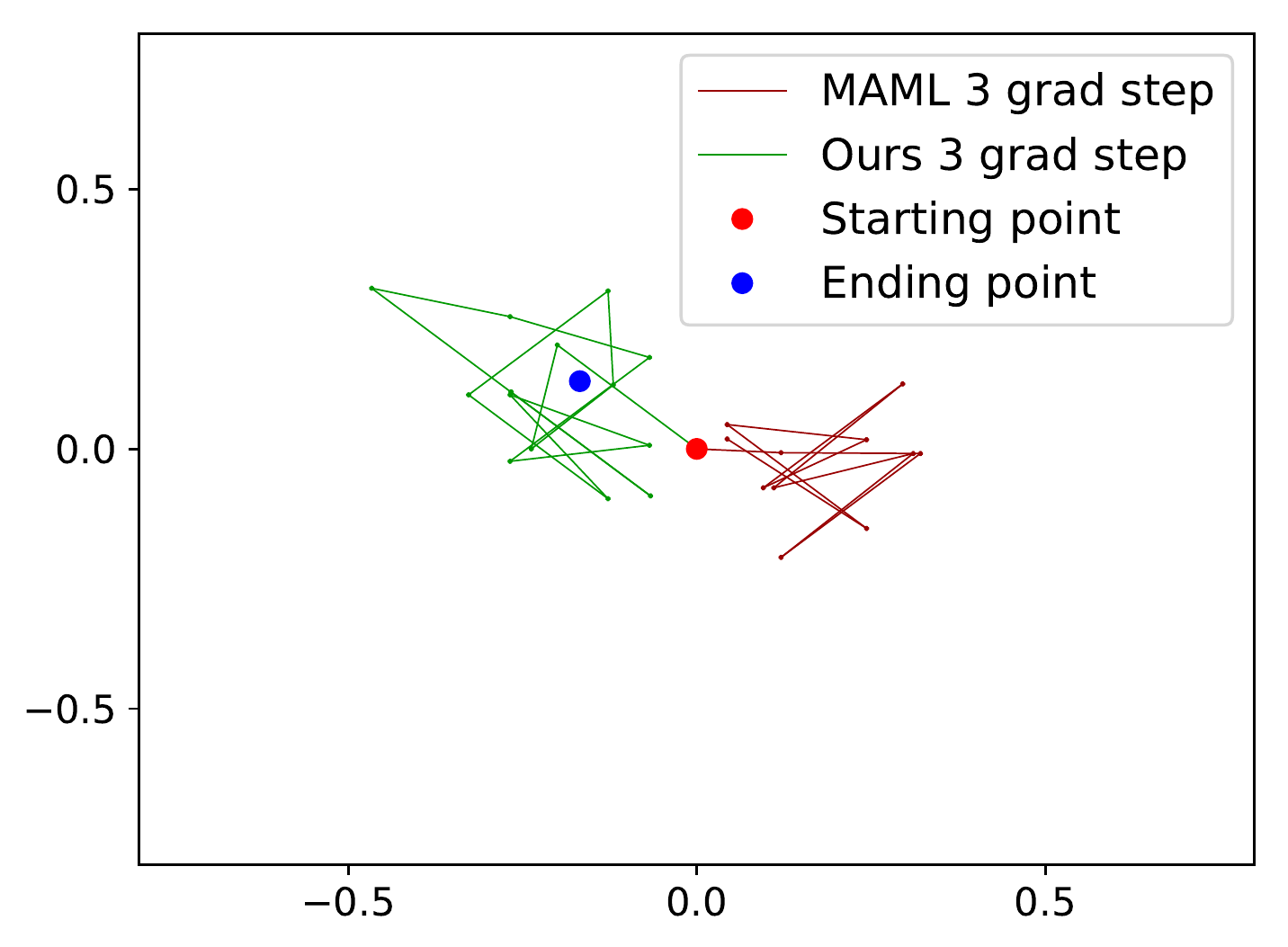}
\vfill
\includegraphics[width=0.49\linewidth]{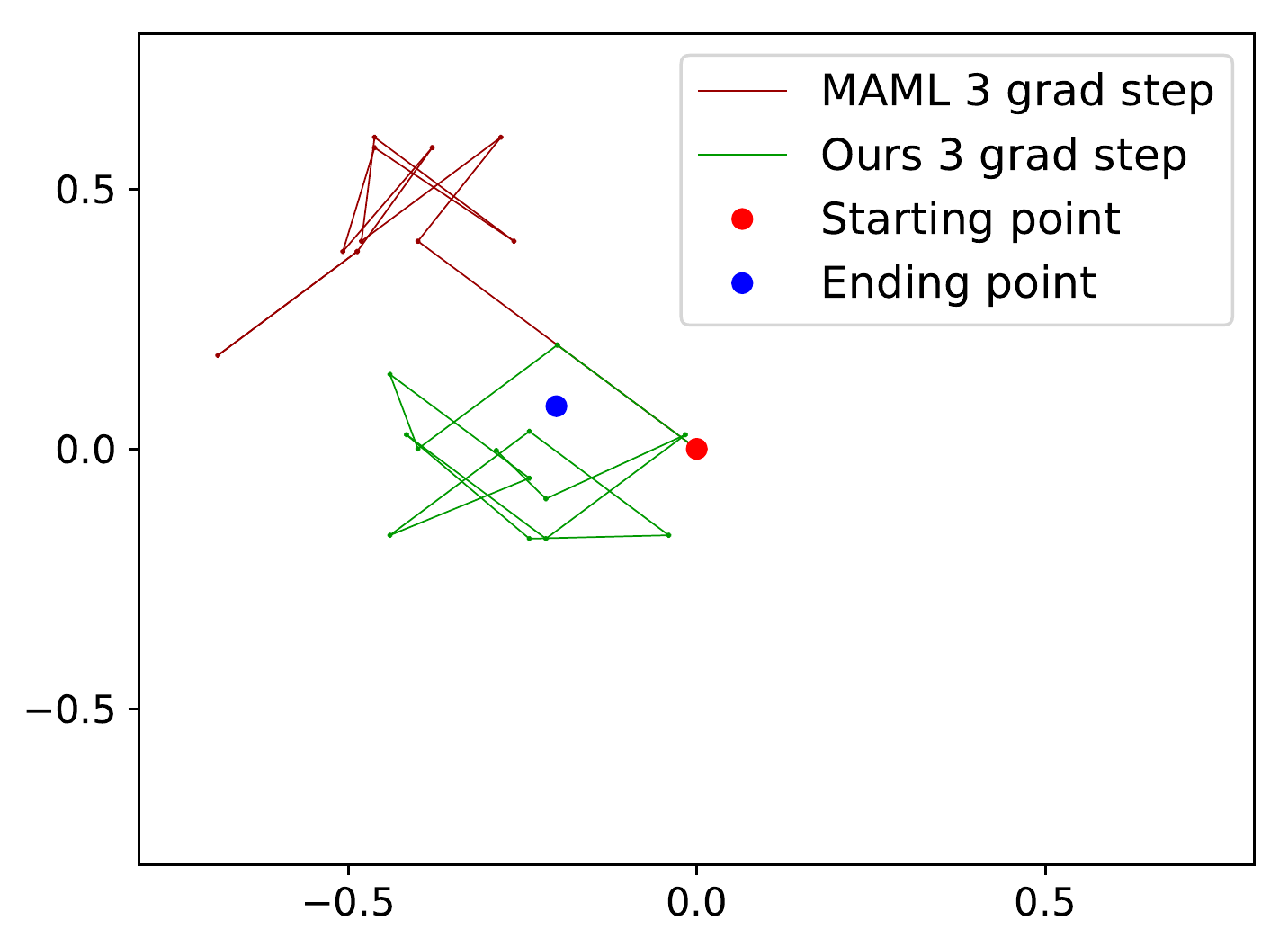}
\includegraphics[width=0.49\linewidth]{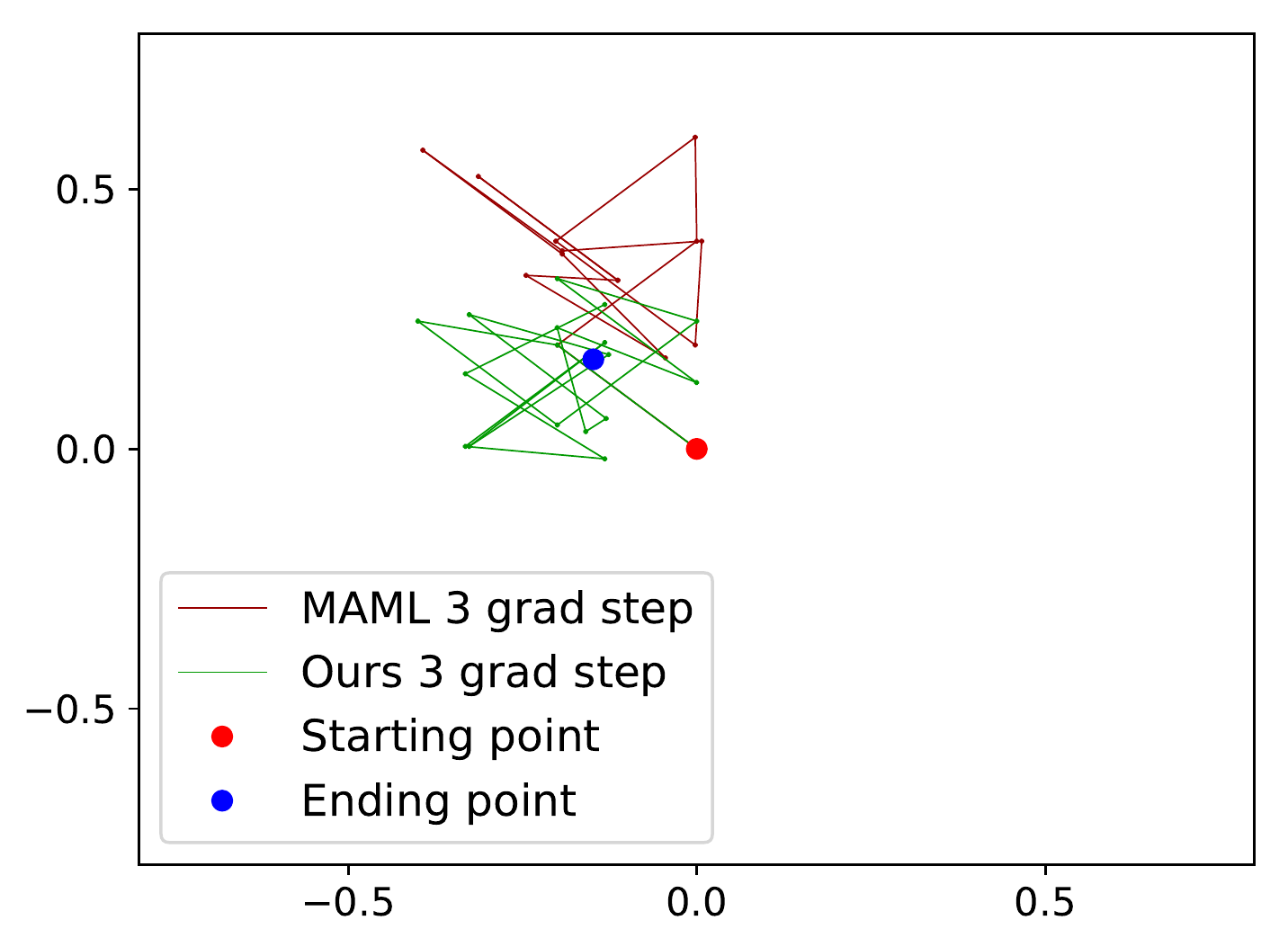}\\
\label{fig:rl1}
\small{(a)} \scriptsize{Start} \tiny{$ = [0.0]$,} \scriptsize{End} \tiny{$\in [-0.5, 0.5] \times [-0.5, 0.5]$}
\end{minipage}
\begin{minipage}[b]{.5\textwidth}
\centering
\includegraphics[width=0.49\linewidth]{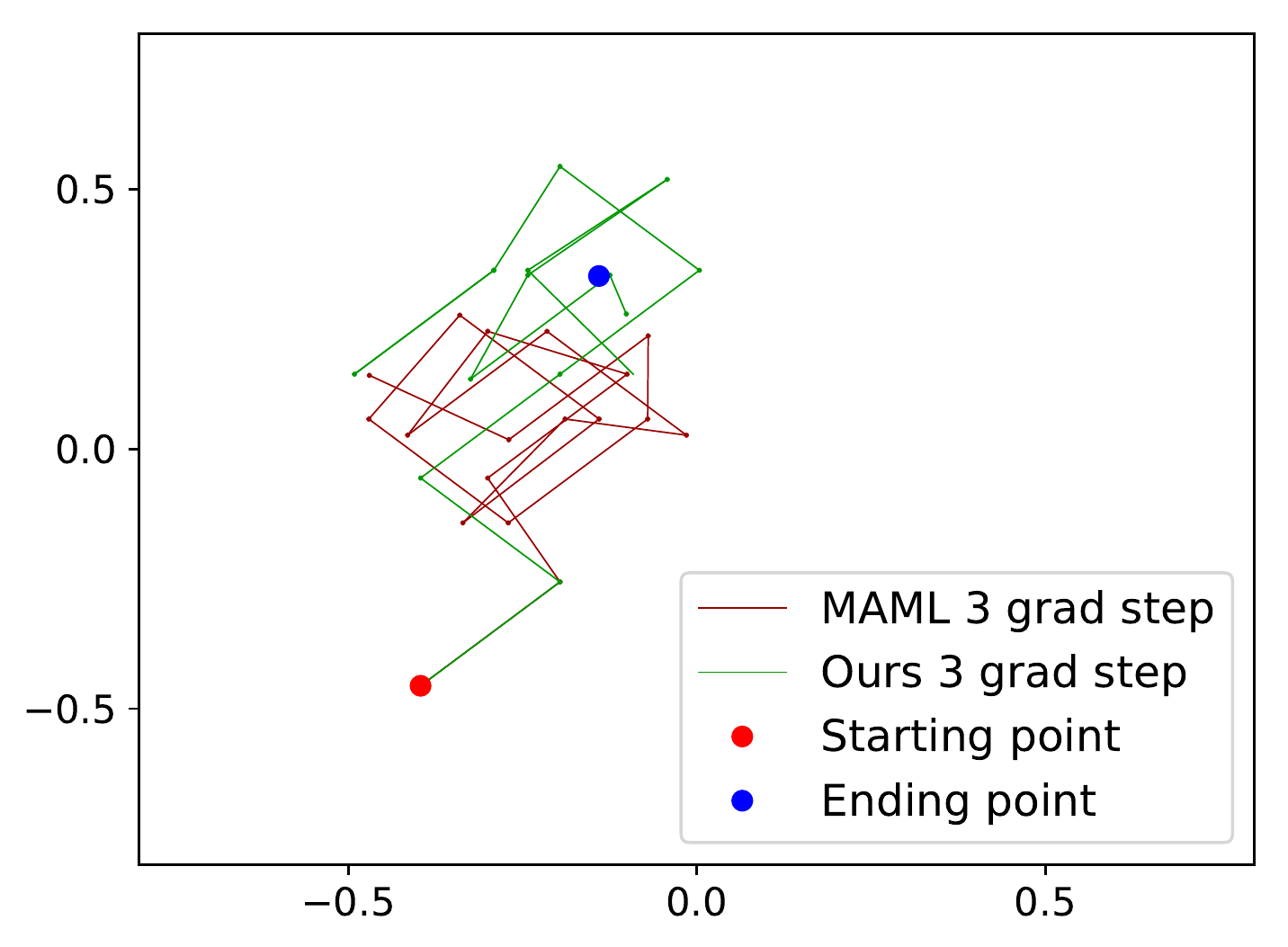}
\includegraphics[width=0.49\linewidth]{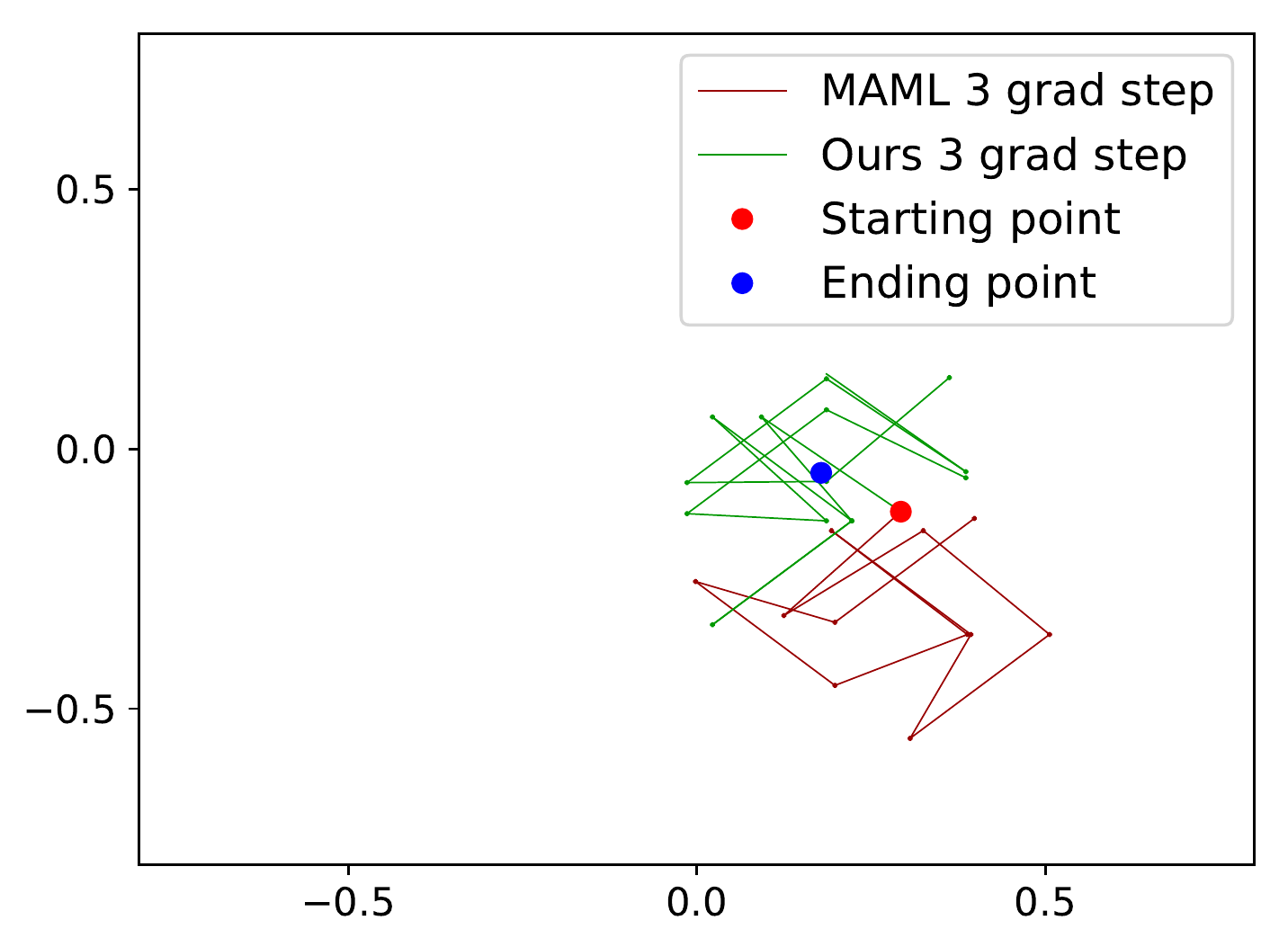}
\vfill
\includegraphics[width=0.49\linewidth]{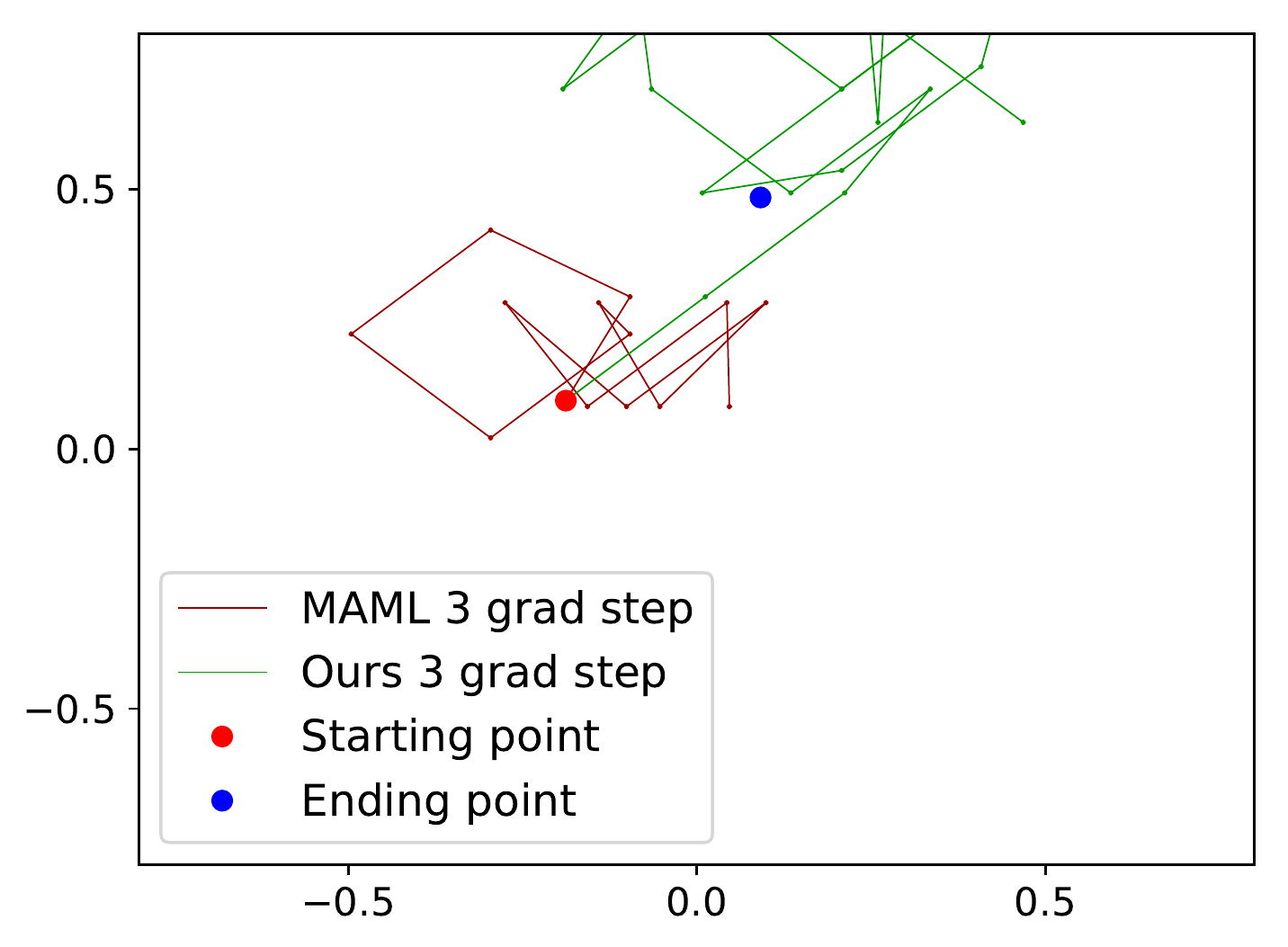}
\includegraphics[width=0.49\linewidth]{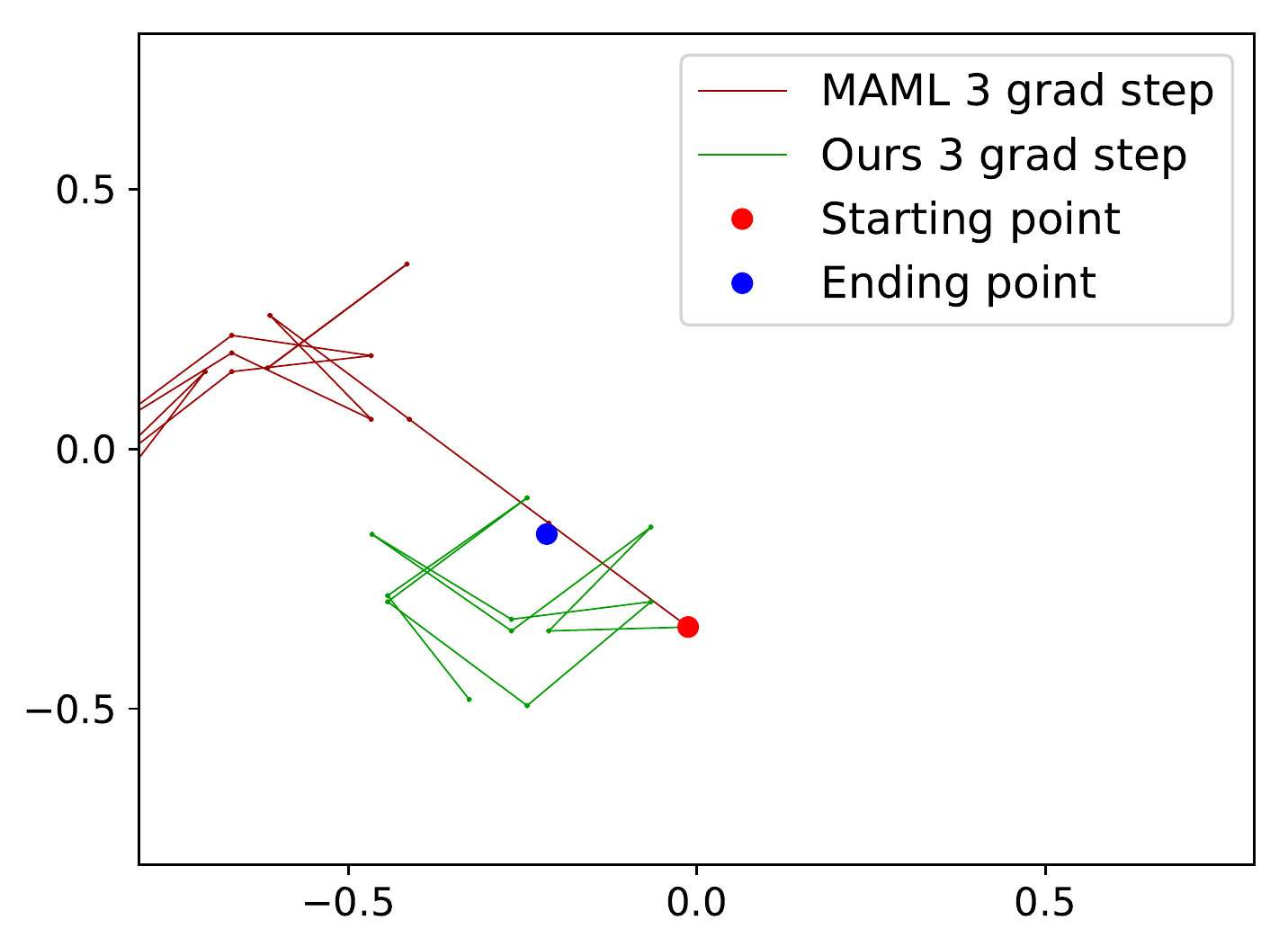}\\
\small{(b)\label{fig:rl2}} \scriptsize{Start, End} \tiny{$ \in [-0.5, 0.5] \times [-0.5, 0.5]$} 
\end{minipage}
\vfill
\vspace{5mm}
\begin{minipage}[b]{.5\textwidth}
\centering
\includegraphics[width=0.49\linewidth]{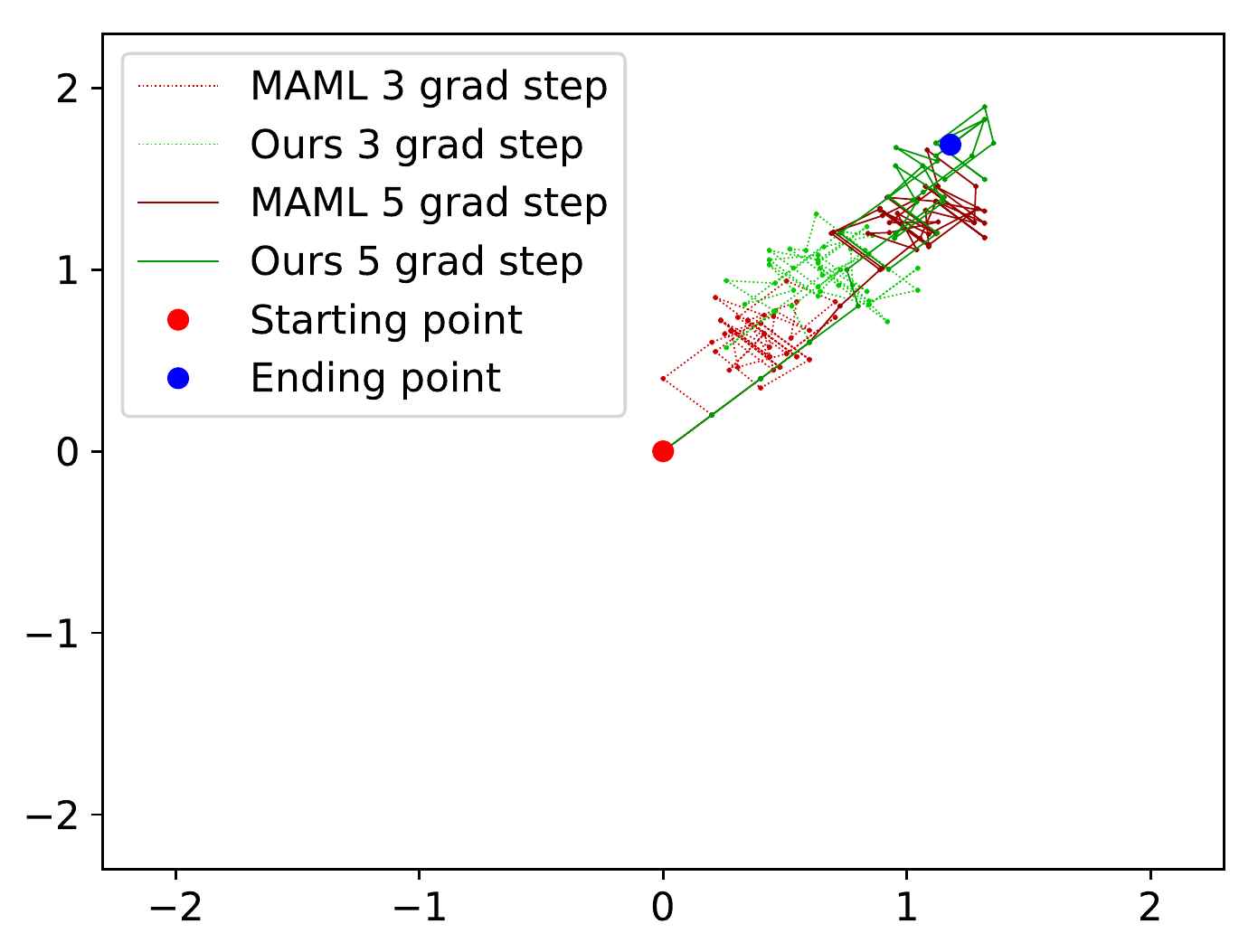}
\includegraphics[width=0.49\linewidth]{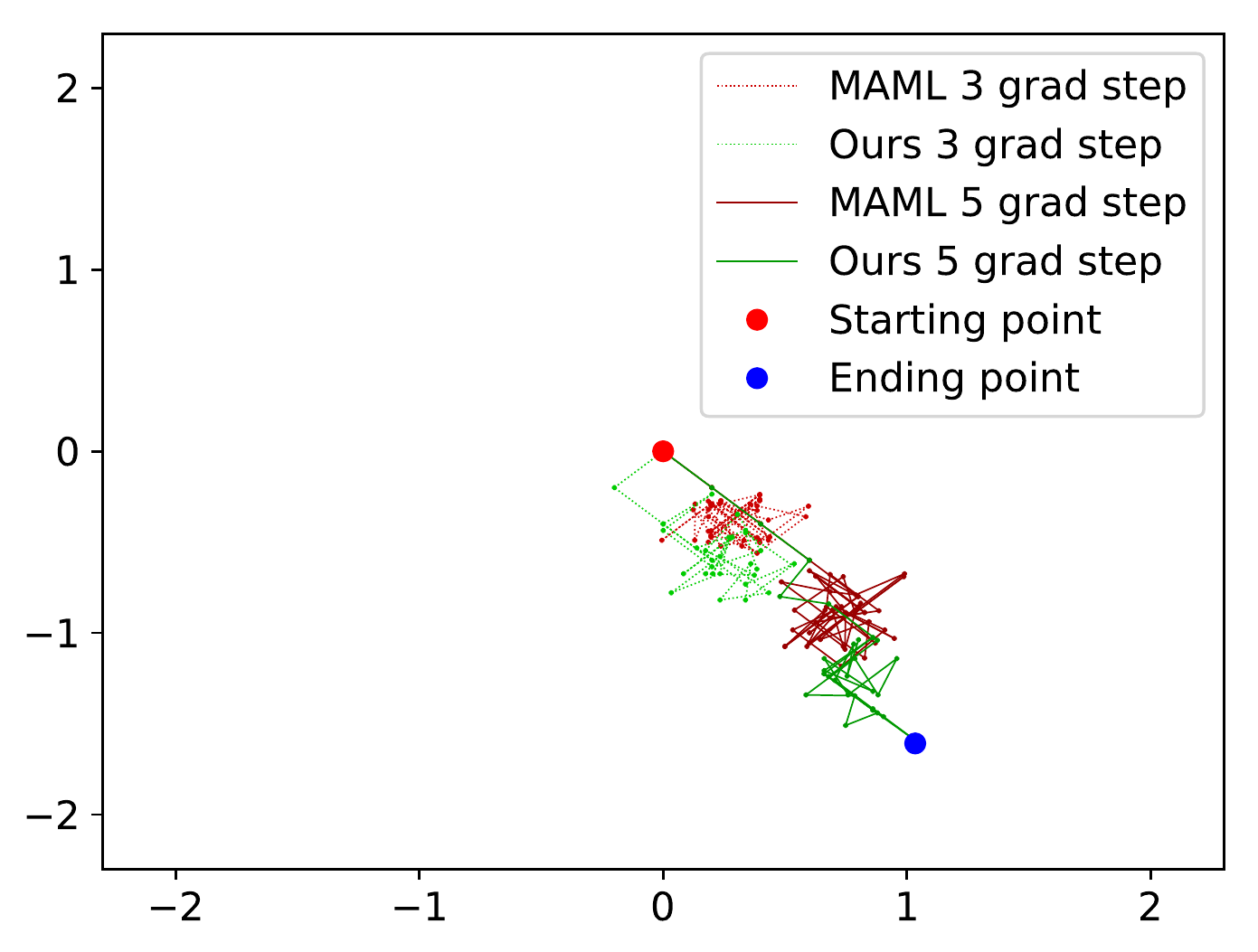}
\vfill
\includegraphics[width=0.49\linewidth]{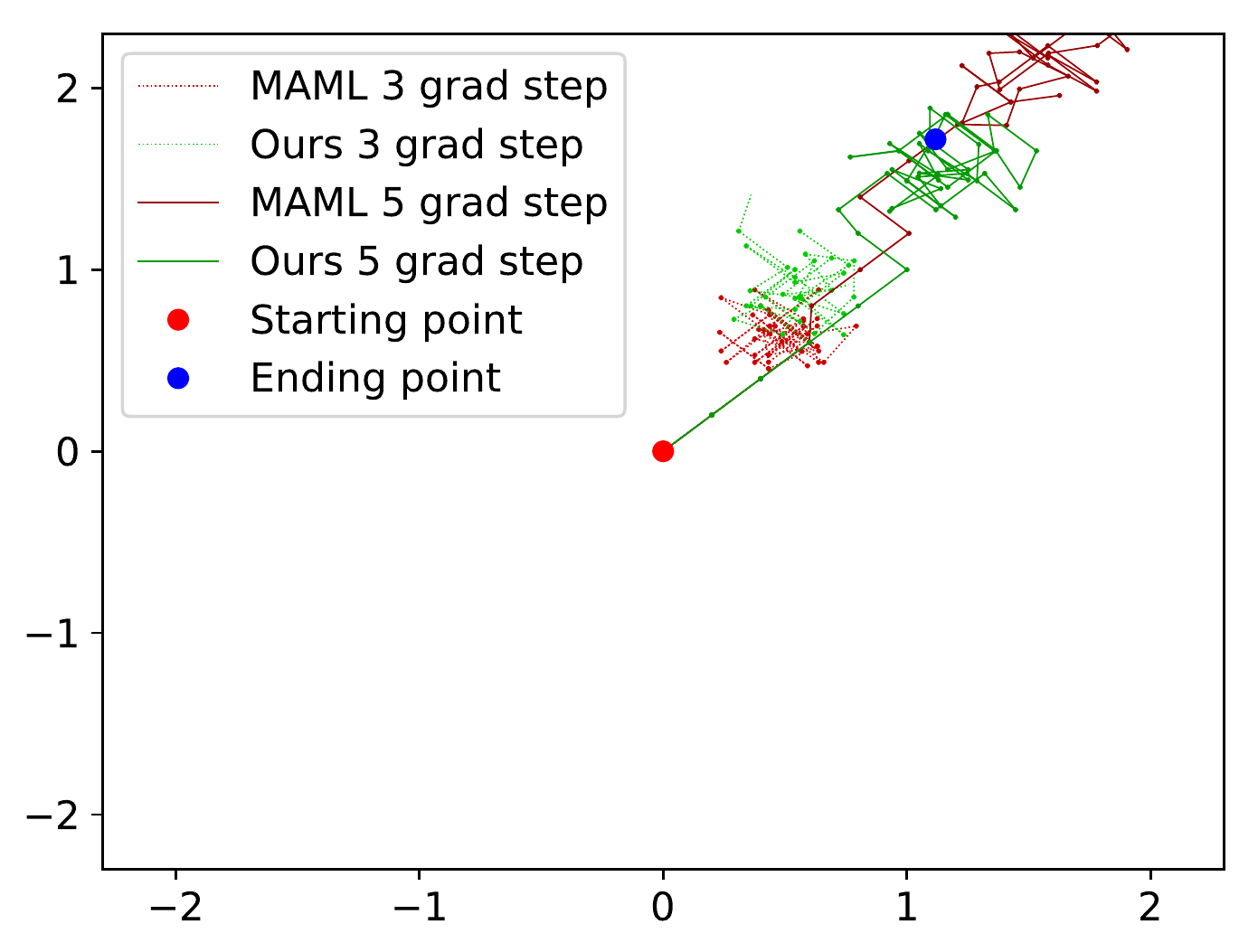}
\includegraphics[width=0.49\linewidth]{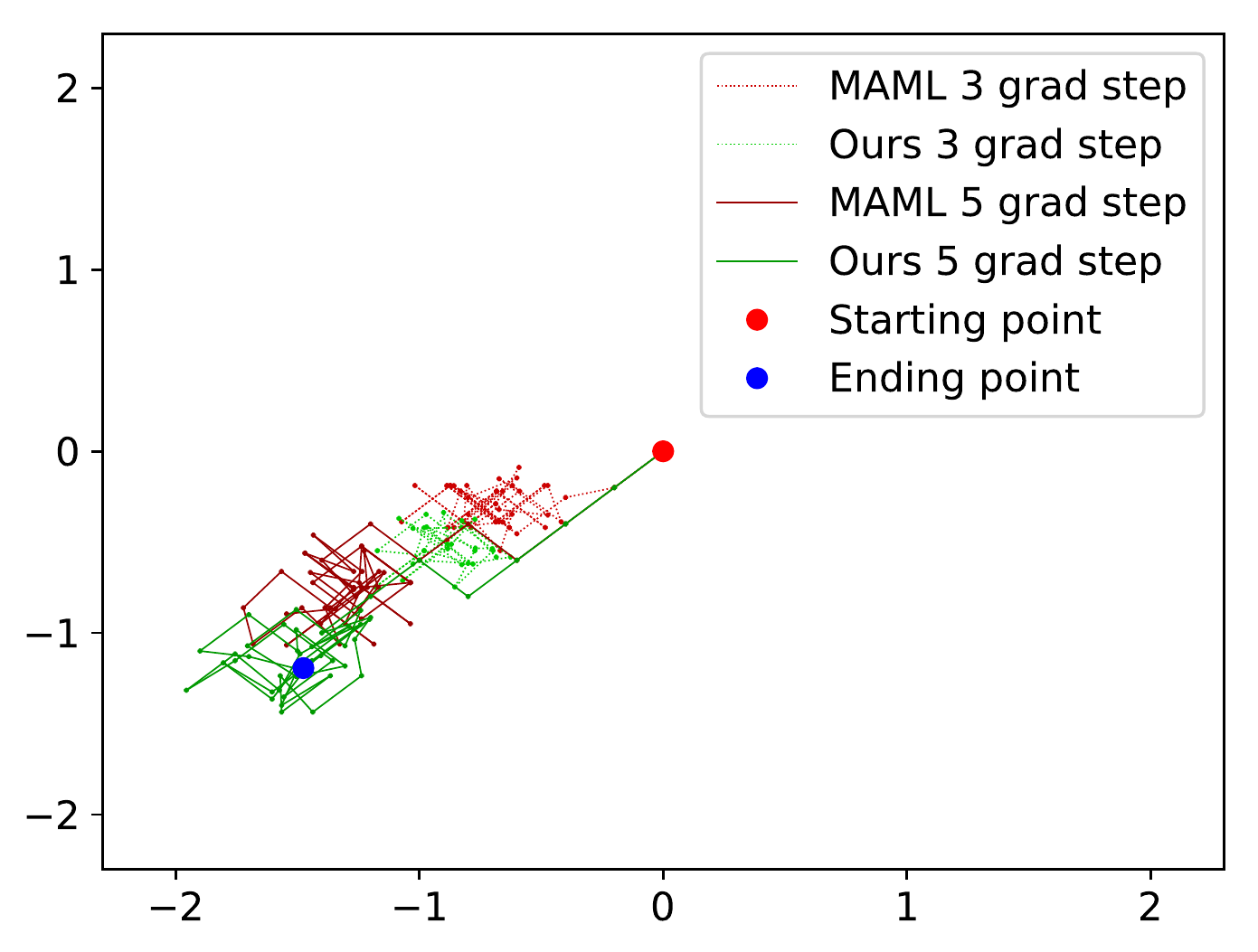}
\vfill
\includegraphics[width=0.49\linewidth]{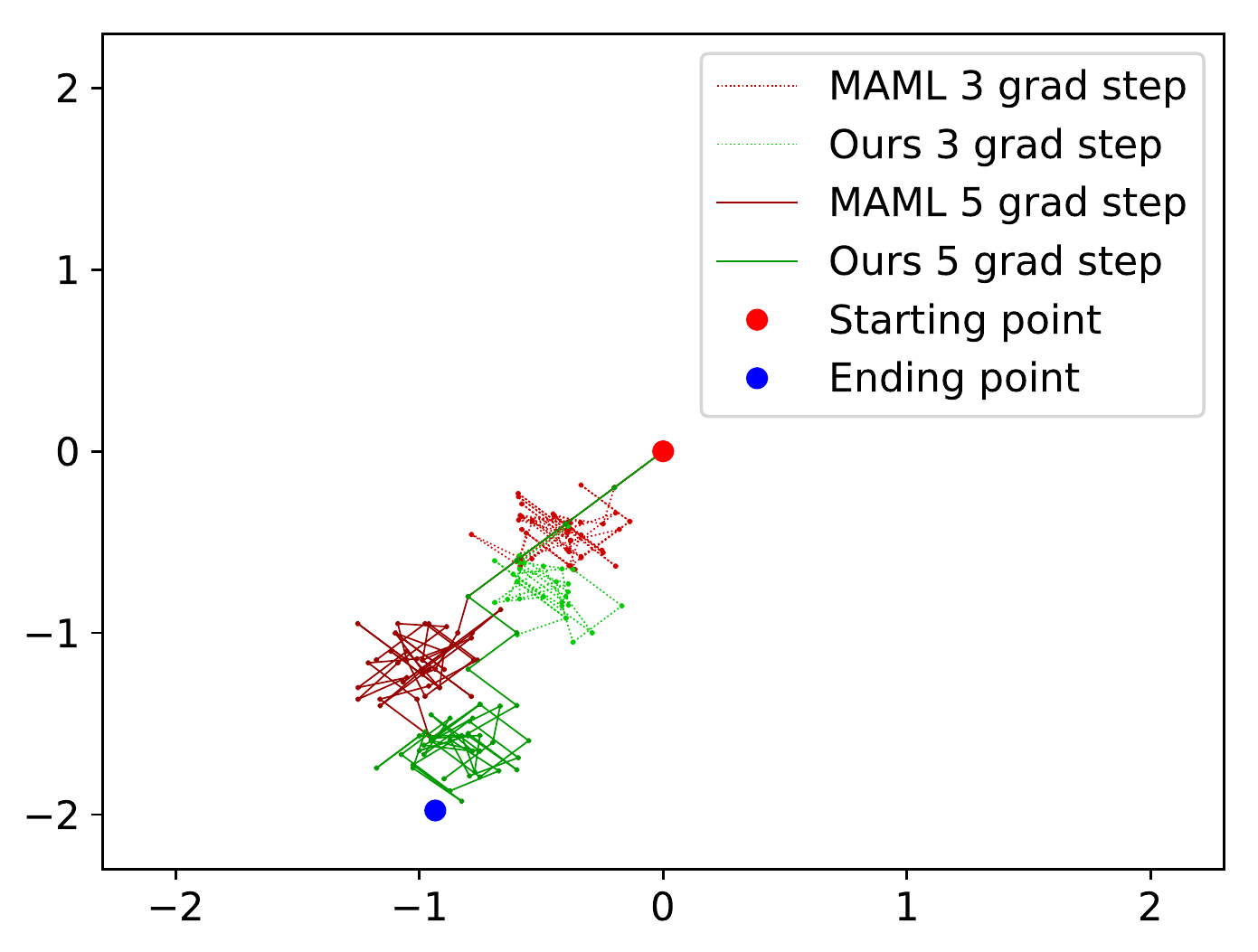}
\includegraphics[width=0.49\linewidth]{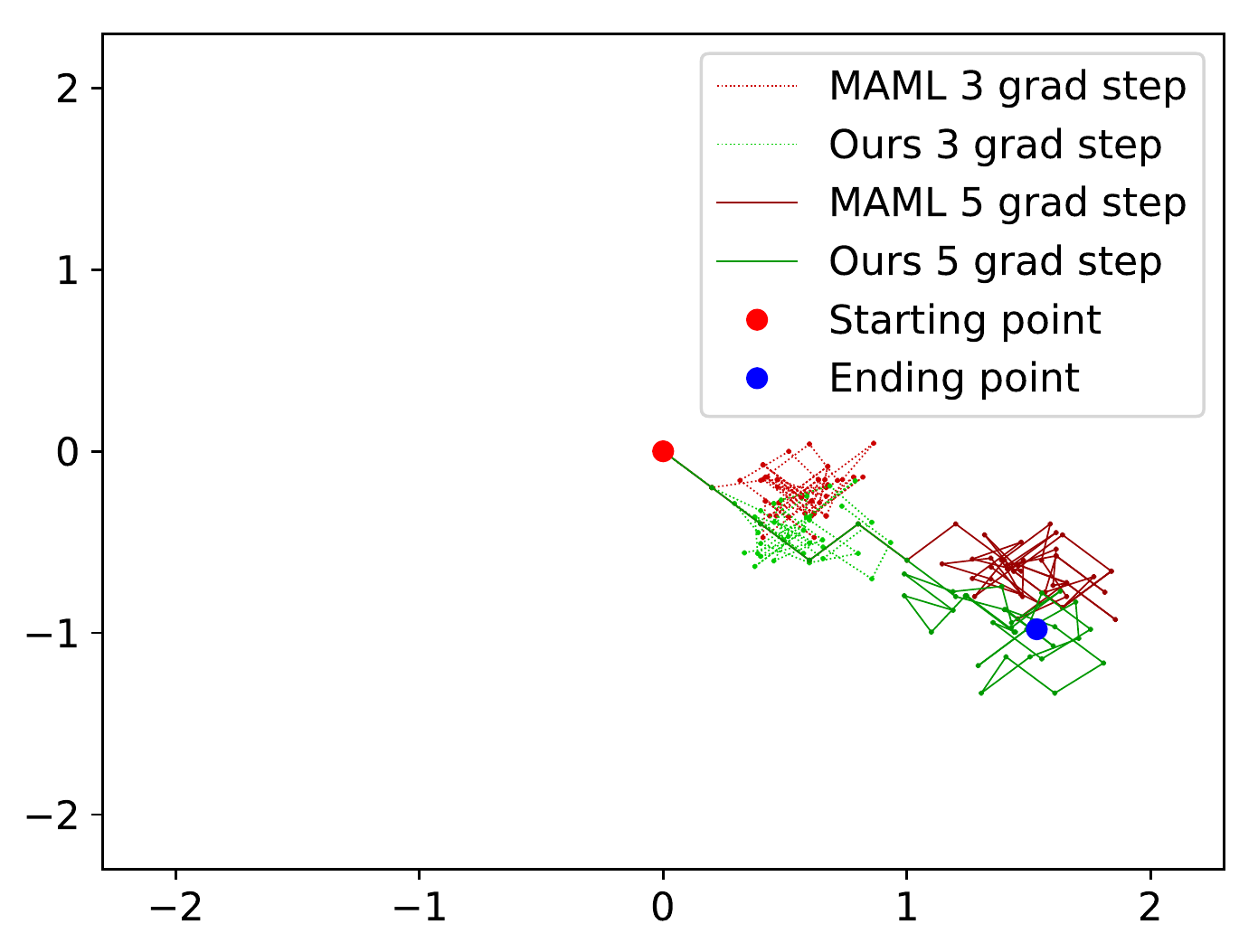}\\
\small{(c)} \scriptsize{Start} \tiny{$ = [0.0]$,} \scriptsize{End} \tiny{$\in [-2.0, 2.0] \times [-2.0, 2.0]$}
\label{fig:rl3}
\end{minipage}
\begin{minipage}[b]{.5\textwidth}
\centering
\includegraphics[width=0.49\linewidth]{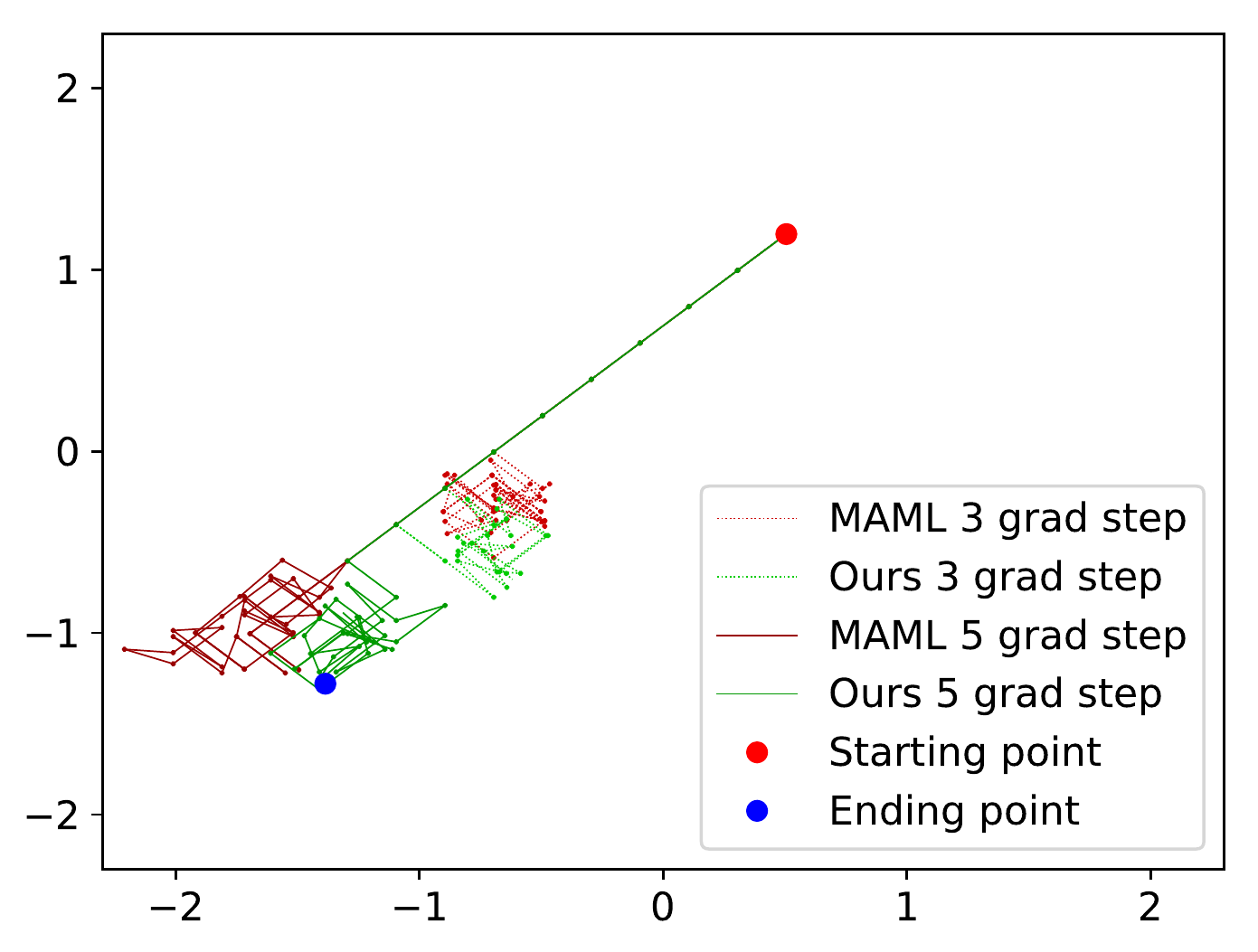}
\includegraphics[width=0.49\linewidth]{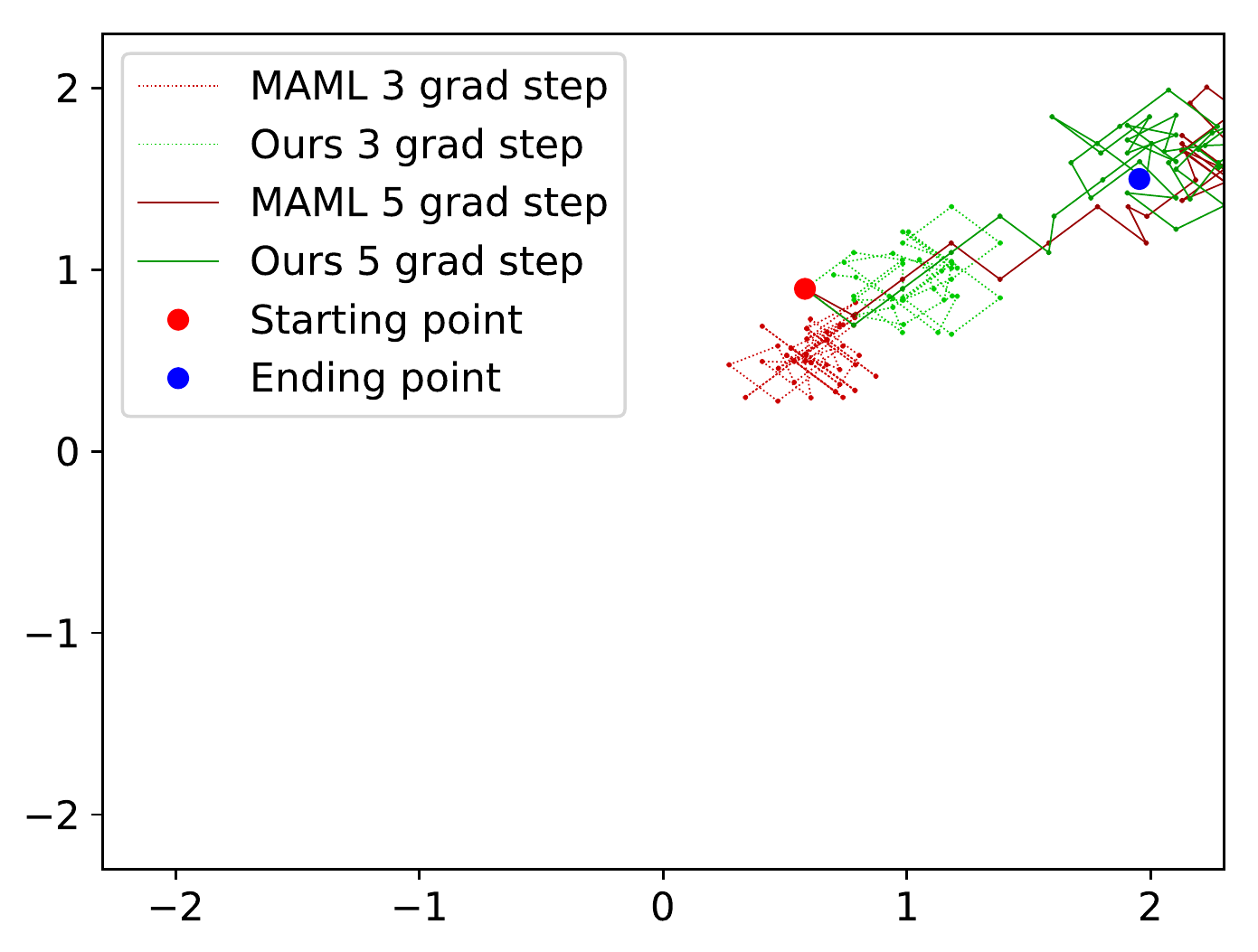}
\vfill
\includegraphics[width=0.49\linewidth]{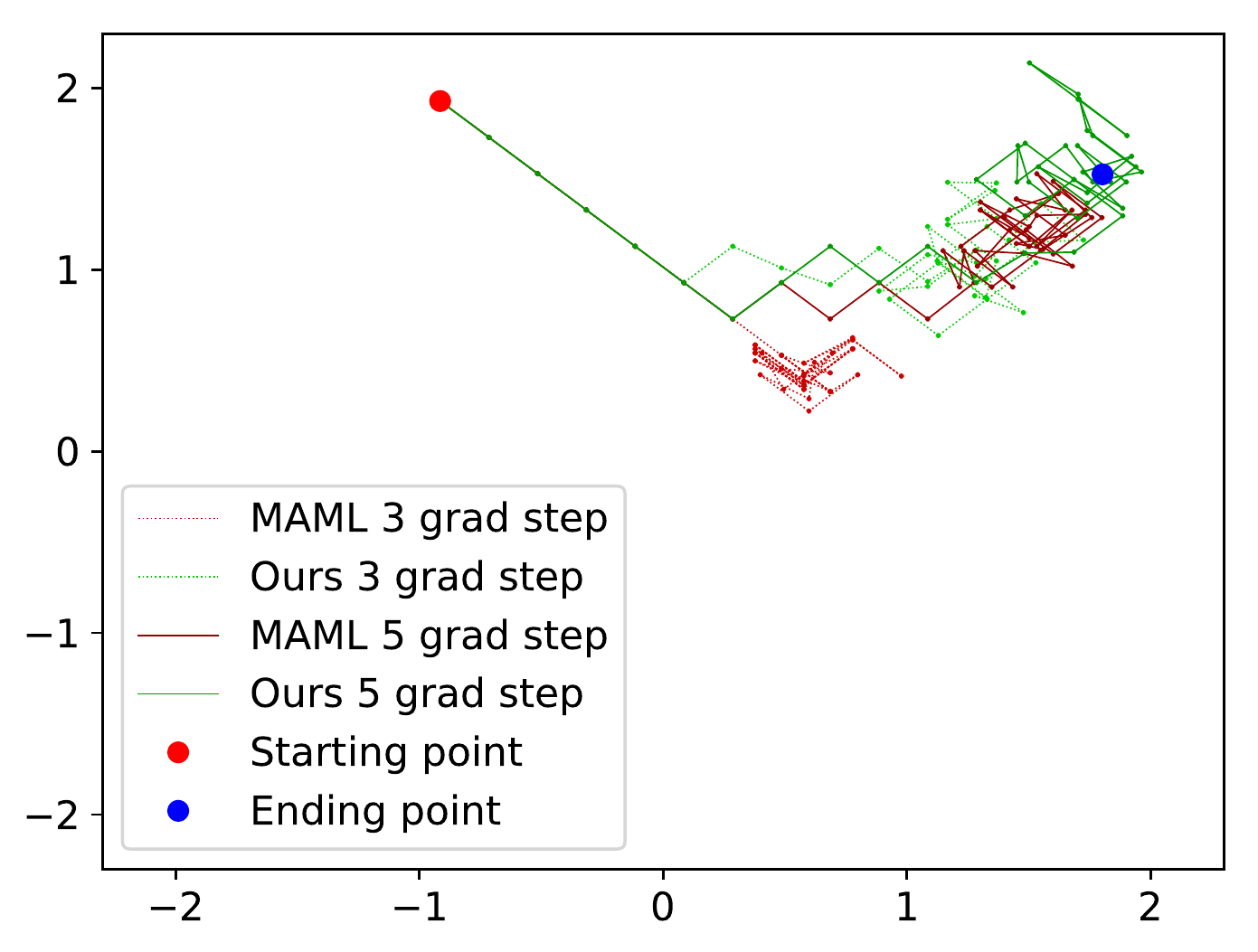}
\includegraphics[width=0.49\linewidth]{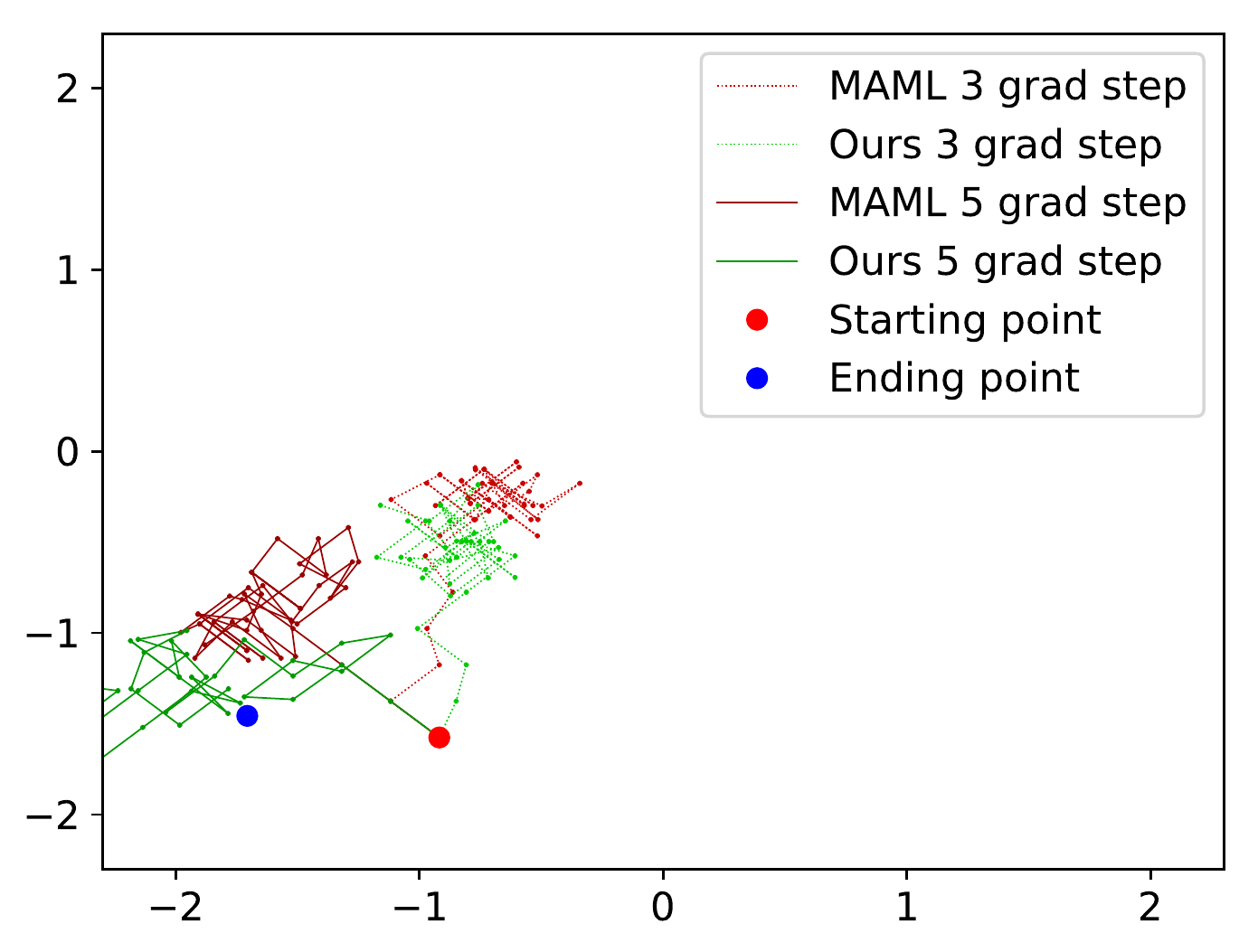}
\vfill
\includegraphics[width=0.49\linewidth]{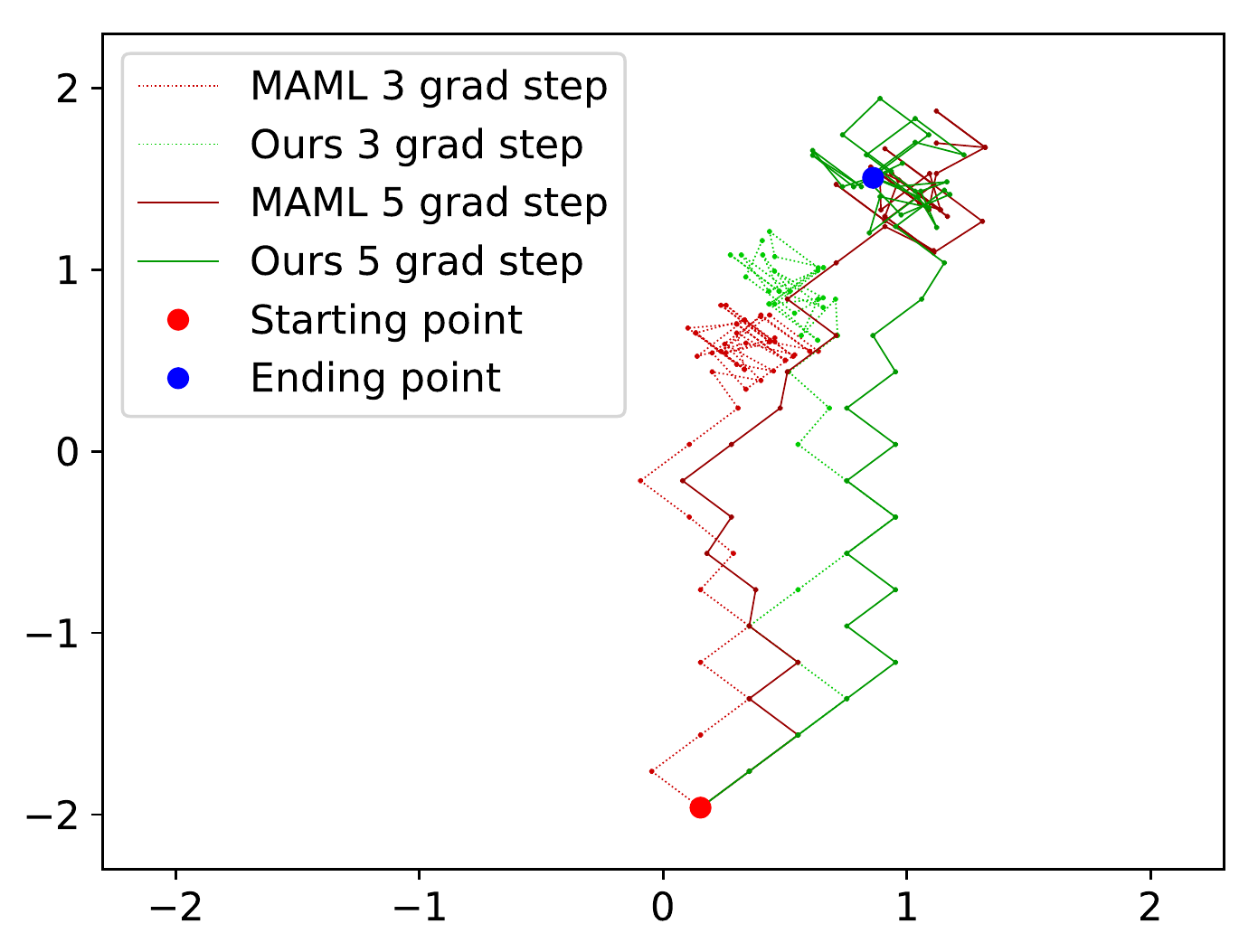}
\includegraphics[width=0.49\linewidth]{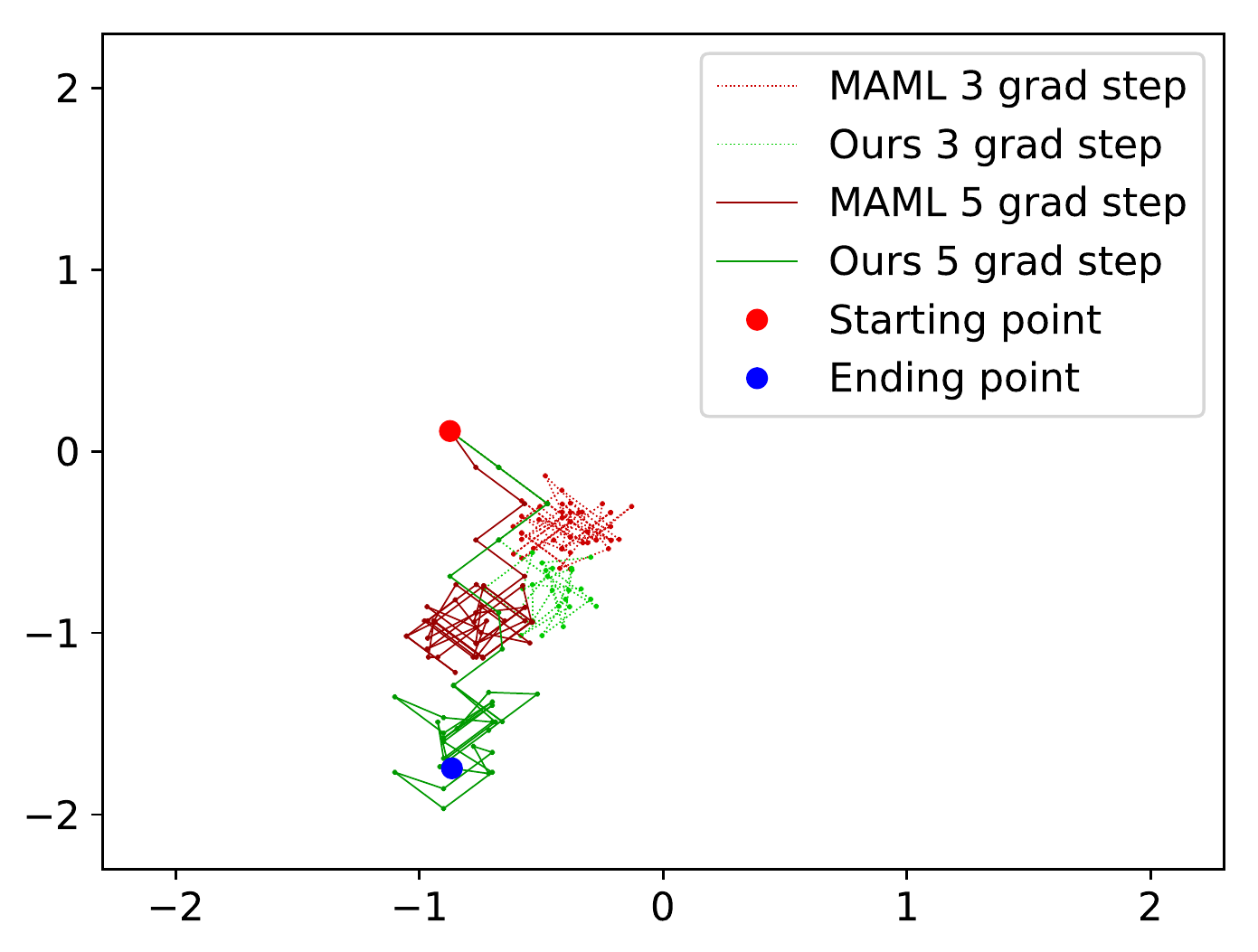}\\
\small{(d)} \scriptsize{Start, End} \tiny{$ \in [-2.0, 2.0] \times [-2.0, 2.0]$} \label{fig:rl4}
\end{minipage}
\caption{
Qualitative results for the 2D Navigation task with MAML vs MAML+L2F (Ours) comparison. Only (a) experiment, tasks are sampled from the same distribution for training and evalution, and (b), (c), and (d) experiments, tasks are sampled from the non-overlapped ranges for training and evaluation}
\label{fig:rl_final}
\end{figure*}

The qualitative results for the 2D navigation experiments are shown in Figure \ref{fig:rl_final}.
In training, the position of starting point is fixed at $[0,0]$ and the position of destination is randomly sampled from space $[-0.5\times 0.5,-0.5\times 0.5]$, which is the same experiment procedure from \cite{finn2017model}. Velocity is clipped to be in the range $[-0.2, 0.2]$.
In evaluation, we performed experiments with four different task distributions. In Figure \ref{fig:rl_final}\textcolor{red}{(a)}, the task distribution for evaluation is the same as for training. 
On the other hand, as in regression experiment \ref{fig:regression figure 2}, we perform additional 3 experiments (Figure \ref{fig:rl_final}\textcolor{red}{(b), (c), (d)} that evaluate models under extreme conditions, where the task distribution for evaluation is chosen to be different from the task distribution for training. In Figure 
\ref{fig:rl_final}\textcolor{red}{(b)}, the staring point is no longer fixed but rather sampled from space $[-0.5 \times 0.5, -0.5 \times 0.5]$. In Figure \ref{fig:rl_final}\textcolor{red}{(c)}, the position of starting point is fixed at $[0,0]$. However, the position of the ending point is sampled from a larger space $[-2.0 \times 2.0,-2.0 \times 2.0]$. In Figure \ref{fig:rl_final}(d), both the starting and destination positions are sampled from space $[-2.0 \times 2.0, -2.0 \times 2.0]$. Overall, our proposed method demonstrates more accurate and robust navigation, compared to the baseline MAML.

\section{Implementation Details}
\subsection{classification}
\subsubsection{Experiment Setup}
We use the standard settings \cite{finn2017model} for $N$-way $k$-shot classification in both miniImageNet\cite{ravi2017optimization} and tieredImageNet\cite{ren2018meta}. When calculating gradients for fast adaptation to each task, the number of examples $\mathcal{D}$ used is either $N$. The fast adaptation is done via 5 gradient steps with the fixed step size, $\alpha = 0.01$ for all models, except LEO and LEO+L2F during both training and evaluation. Gradients for meta-updating the networks $f_\theta$ and $g_\phi$ are calculated with 15 number of examples $\mathcal{D}'$ at each iteration. The MAML and its variants were trained for 50000 iterations in miniImageNet and 125000 in tieredImageNet to account for the larger number of examples as in \cite{learning2019liu}. The meta batch size of tasks is set to be 2 for 5-shot and 4 for 1-shot, with the exception that the batch size is 1 for ResNet12 in miniImageNet and tieredImageNet. This is due to the limited memory and the heavy computation load from the combination of second-order gradient computation, large image size, and a larger network. As for experiments with LEO, we follow the exact setup from LEO \cite{andrei2019meta}. We only add attenuation process before adaptation in latent space and fine-tuning in parameter space for LEO+L2F.
\subsubsection{Network Architecture for $f_\theta$}
\noindent{\textbf{4 conv}}
As with most algorithms \cite{vinyals2016matching, ravi2017optimization, NIPS2017_6996, Sung_2018_CVPR} that use 4-layer CNN as a backbone, we use 4 layers each of which contains 64-filter 3 $\times$ 3 convolution filters, a batch normalization \cite{ioffe2015batch}, a Leaky ReLU nonlinearity, and a 2$\times$2 max pooling. Lastly, the classification linear layer and softmax are placed at the end of the network.

\noindent{\textbf{ResNet12}} As for the ResNet12 architecture, the network consists of 4 residual blocks. each residual block consists of three $3 \times 3$ convolution layers. The first two convolution layers are followed by a batch normalization and a ReLU nonlinearity. The last convolution layer is followed by a batch normalization and a skip connection that contains a $1 \times 1$ convolution layer and a batch normalization. After a skip connection, a ReLU nonlinearity and a max $2 \times 2$ are placed at the end of each residual block.

The number of convolution filters for 4 residual blocks is set to be 64, 128, 256, 512 for 4 residual blocks in the increasing order of depth. 
\subsubsection{Network Architecture for $g_\phi$}
$g_\phi$ is a 3-layer MLP, with each layer of $l$ hidden units, where $l$ is the number of layers of the main network, $f_\theta$.
Activation functions are ReLU in between and a sigmoid at the end.
The input is layer-wise mean of gradients.
As for LEO, the case is a bit different because they perform one adaptation in latent codes and one on decoded classifier weights. Thus, we introduce two 3-layer MLPs, one for each. Again, for each MLP, the numbers of hidden units for each layer is $n$, where $n$ is the dimension of latent codes or the number of classifier weights. 
\subsection{Regression and RL}

The details of the few shot regression and reinforcement learning experiments are listed in Table \ref{tab:impl_details}.

\begin{table*}
\centering
\subfloat[Regression]{\label{tab:reg_imp}\scalebox{.8}{
\begin{tabular}{@{\extracolsep{5pt}} ll} 
\vspace{-0.3cm}
\\[-1.8ex]\hline 
\hline \\[-1.8ex] 
\multicolumn{1}{c}{} & \multicolumn{1}{c}{Hyperparameters}\\
\hline \\[-1.8ex] 
policy network & 2 hidden layers of size 40 with ReLU\\
training iterations & 50000 epochs\\
   inner update $\alpha$ & 0.01\\
   meta update optimizer & Adam\\
   meta batch size & 4\\
   k shot & [5,10,20]\\
   loss function & MSE loss\\
    \multirow{2}{*}{eval}& randomly sample 100 sine curves, \\
  & sample 100 examples (repeated 100 times)\\
\hline  \\[-1.8ex] 
\end{tabular}
}}
\quad
\subfloat[RL]{\label{tab:rl_imp}\scalebox{.8}{
\begin{tabular}{@{\extracolsep{5pt}} llrrrr} 
\vspace{-0.3cm}
\\[-1.8ex]\hline 
\hline \\[-1.8ex] 
\multicolumn{1}{c}{} & \multicolumn{1}{c}{Hyperparameters}\\
\hline \\[-1.8ex] 
  policy network & 2 hidden layers of size 100 with ReLU\\
   inner update & vanilla policy gradient\\
   inner update $\alpha$ & 0.01\\
   meta-optimizer & TRPO\\
   training iterations & 500 epochs, choose best model\\\hline
   MuJoCo horizon & 200\\
   MuJoCo batch size & 40\\
   \multirow{2}{*}{MUJoCo evals} & update 4 gradient updates, \\
   &each with 40 samples for a task\\\hline
   2D navigation horizon & 100\\
   2D navigation batch size & 20\\
   \multirow{2}{*}{2D navigation evals} & update 4 gradient updates, \\
   & each with 20 samples for a task\\
\hline  \\[-1.8ex]
\end{tabular}
}}
\caption{Implementation Details for Regression and RL}
\label{tab:impl_details}
\vspace{-0.5cm}
\end{table*}
\subsection{System}
All experiments were performed on a single NVIDIA GeForce GTX 1080Ti.

\clearpage
\clearpage
\begin{figure}[H]
\begin{center}
\includegraphics[width=\columnwidth]{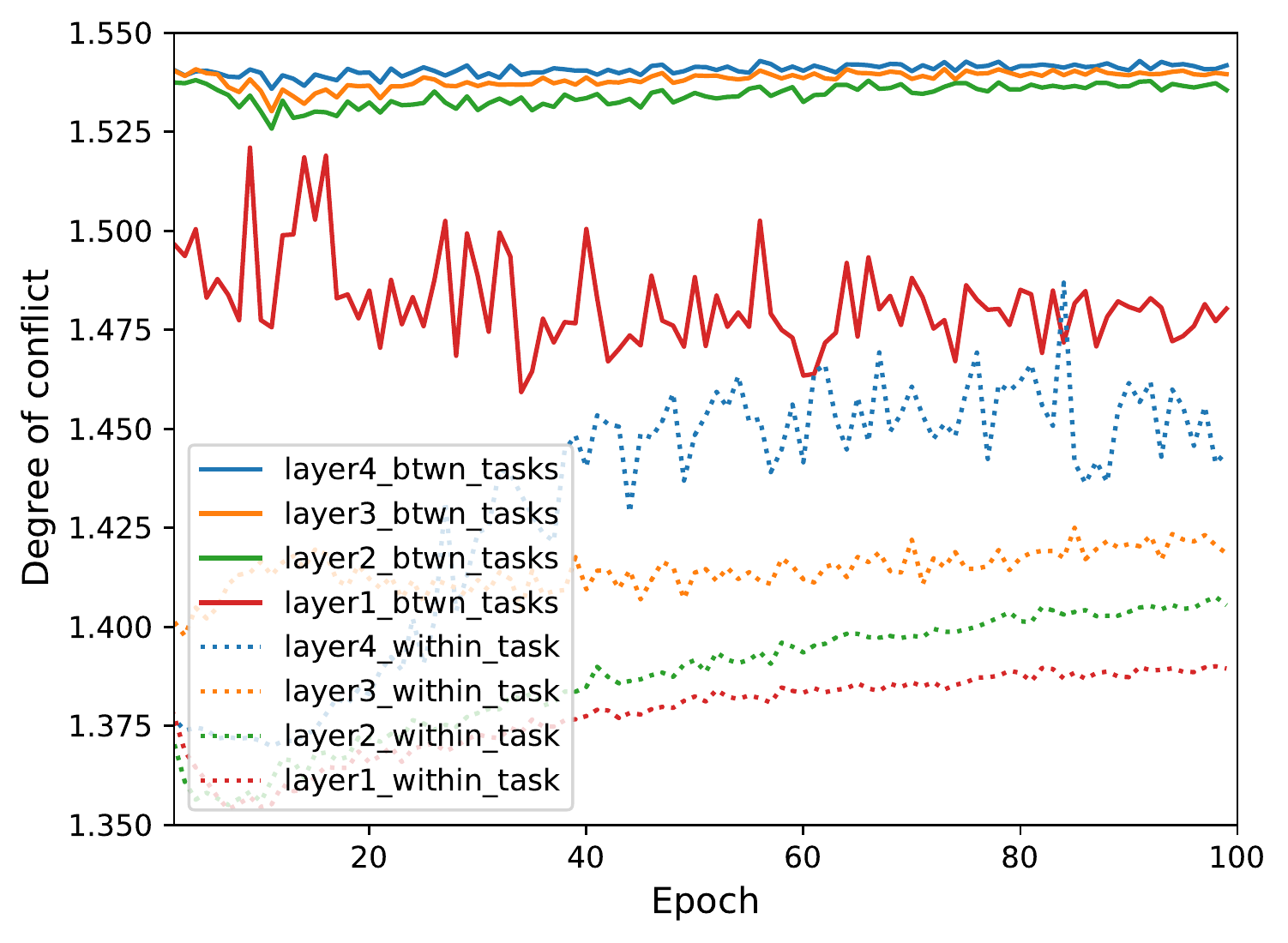}
\end{center}
\caption{Degree of conflict within a task: The degree of conflict within each task is averaged across tasks per each epoch. The degree of conflict within each task is observed to be lower than the degree of conflict between tasks. This is expected as the examples within a task are more similar than examples from different tasks. Thus, the gradients are more aligned within tasks.}
\label{fig:conflict_within_task}
\end{figure}
\section{Conflict Within a Task}
In this section, we also measure the degree of conflict that exist within task (conflict between examples in the same task), as illustrated in Figure~\ref{fig:conflict_within_task}. As expected, the degree of conflict within task is observed to be lower than the degree of conflict between tasks. This is because the examples within a task are more similar to each other than examples from different tasks. This leads to gradients being more aligned within a task, while gradients between tasks have a larger amount of disagreement.
\paragraph{Acknowledgements}

\noindent This work was supported by IITP grant funded by the Ministry of Science and ICT of Korea (No. 2017-0-01780), and Hyundai Motor Group through HMG-SNU AI Consortium fund (No. 5264-20190101).
\clearpage
\clearpage
{\small
\bibliographystyle{ieee_fullname}
\bibliography{camera_ready}
}

\end{document}